\documentclass[review]{elsarticle}

\usepackage{lineno,hyperref}

\usepackage{mathrsfs}
\usepackage{enumitem}
\usepackage{algorithm}
\usepackage{algpseudocode}
\usepackage{color}

\usepackage{amsmath}
\usepackage{amssymb}
\usepackage{bbding}
\usepackage{subfigure}

\usepackage{booktabs} 
\usepackage{multirow}
\usepackage{multicol}

\usepackage{amsfonts}%
\usepackage{bm}
\modulolinenumbers[5]
\biboptions{numbers,sort&compress}

\begin{document}
	
	\begin{frontmatter}
		
		%
		%
		%
		%
		
		\title{IA-FaceS: A Bidirectional Method for Semantic Face Editing}
		
		
		\author{Wenjing Huang}
		\ead{huangwenjing@sjtu.edu.cn}
		\author{Shikui Tu\corref{mycorrespondingauthor}}
		\ead{tushikui@sjtu.edu.cn}
		\author{Lei Xu\corref{mycorrespondingauthor}}
		\ead{leixu@sjtu.edu.cn}
		
		\address{Department of Computer Science and Engineering,\\Shanghai Jiao Tong University, Shanghai, China}
		\cortext[mycorrespondingauthor]{Corresponding author}
		\begin{abstract}
			Semantic face editing has achieved substantial progress in recent years. Known as a growingly popular method, latent space manipulation performs face editing by changing the latent code of an input face to liberate users from painting skills. However, previous latent space manipulation methods usually encode an entire face into a single low-dimensional embedding, which constrains the reconstruction capacity and the control flexibility of facial components, such as eyes and nose. This paper proposes IA-FaceS as a bidirectional method for disentangled face attribute manipulation as well as flexible, controllable component editing without the need for segmentation masks or sketches in the original image.  To strike a balance between the reconstruction capacity and the control flexibility, the encoder is designed as a multi-head structure to yield embeddings for reconstruction and control, respectively: a high-dimensional tensor with spatial properties for consistent reconstruction and four low-dimensional facial component embeddings for semantic face editing. A simple, fixed mask is devised over an intermediate latent feature
			tensor to segment component-specific regions and extract an embedding for each component separately through component-specific sub-embedding layers. Manipulating the separate component embeddings can help achieve disentangled attribute manipulation and flexible control of facial components. To further disentangle the highly-correlated components, a component adaptive modulation (CAM) module is proposed for the decoder. The semantic single-eye editing is developed for the first time without any input visual guidance, such as segmentation masks or sketches. According to the experimental results, IA-FaceS establishes a good balance between maintaining image details and performing flexible face manipulation. Both quantitative and qualitative results indicate that the proposed method outperforms the other techniques in reconstruction, face attribute manipulation, and component transfer.
		\end{abstract}
		
		
	\end{frontmatter}
	

	\section{Introduction}   \label{sec:introduction}
	
	Semantic face editing has grown in popularity in computer vision due to its various practical applications including face attribute manipulation and semantic facial component editing. Recently, latent space manipulation \cite{interfacegan,stylegan,indomain,stylegan2,homomorphic-latent-space,ganspace,gan-parameter,closed-form} has become an increasingly popular approach to face editing. Latent space manipulation methods are implemented by changing the latent codes of images, thereby liberating users from complex modifications in visual spaces. They generally follow the pipeline of encoding a face into the latent space, manipulating the latent vector, and synthesizing the face accordingly. By utilizing the semantics within the latent space, latent space manipulation can continuously edit arbitrary attributes of facial images. 
	
	Real image manipulation usually requires accurate reconstruction and high-quality interpolation results. In most latent space manipulation methods \cite{stylegan,alae,interfacegan,homomorphic-latent-space,indomain,voynov2020unsupervised}, an entire face image is often embedded into a low-dimensional vector for better interpolation quality. However, a low-dimensional vector cannot maintain image details and thus will constrain the reconstruction capacity of the whole network. Some methods \cite{stylemapgan} utilize a high-dimensional tensor with spatial properties as the representation of an input face and enhance the reconstruction quality of the input image. Unfortunately, the direct manipulation of the high-dimensional tensor would introduce artifacts in face editing, according to Figure \ref{fig:exp-com-stylemapgan}. Other methods \cite{image2stylegan, image2stylegan++} were also proposed for image reconstruction by optimizing the intermediate representation for every layer of StyleGAN \cite{stylegan}; however, they suffer from high computational cost and poor interpolation quality \cite{indomain, stylemapgan}, limiting their applications in practice.  Hence, it is a challenging task to utilize the same embedding  for reconstruction and face manipulation simultaneously.
	
	As another requirement for face manipulation models, disentangled attribute manipulation requires the model to separate attribute-irrelevant characteristics from attribute information. It is a challenging task because there often exist strong correlations between different attributes. For example, as shown in Table \ref{tab:intro-sample-bias}, \emph{eyebrows} in CelebA-HQ \cite{pg-gan} have a significant correlation with \emph{gender}  (p-value=$5.8e^{-219}$, $\chi^2$ test). To achieve attribute correctness whilst avoiding unintended altering, many recent works \cite{interfacegan,ganspace,l2mgan,li2020latent,abdal2020styleflow,closed-form,gan-parameter,voynov2020unsupervised} focus on discovering disentangled factors or directions for attributes in GAN's latent space. According to a fixed pre-trained GAN, recent works either map the latent vector to each attribute via subspace projection using attribute labels \cite{interfacegan,abdal2020styleflow} or identify interpretable directions in the latent space of GAN in an unsupervised manner \cite{voynov2020unsupervised,ganspace,spingarn2021gan,closed-form,gan-parameter}. Although some attributes can be edited individually \cite{interfacegan,ganspace,closed-form,l2mgan,li2020latent}, the whole-face single-embedding manner limits the flexibility of editing facial components. Another recent line of approach resorts to geometry-guided disentanglement \cite{ganimation,make-a-face,r-face,mask-guided,deng2020disentangled}, which requires additional geometric guidance in visual space as input, such as facial landmarks \cite{ganimation, make-a-face}, segmentation mask \cite{mask-guided} or 3D information \cite{deng2020disentangled}, etc. Although these works can achieve flexible component editing \cite{ganimation,r-face,mask-guided,deng2020disentangled}, the visual geometric guidance may be difficult or inconvenient to obtain. 
	
	To address the above issues, this paper proposes IA-FaceS, which is a deep bidirectional network for high-quality image reconstruction, disentangled face attribute manipulation, and flexible, controllable component editing. Following the deep yIng-yAng (IA)  bidirectional intelligence framework \cite{xu-bidirectional}, IA-FaceS is characterized by an encoder network (A-mapping for abstract) to efficiently compute latent representation for the input face, a synthesis reasoning process over the latent space for face manipulation, and a decoder network (I-mapping for inference) for high-quality synthesis outputs. To strike a balance between the reconstruction capacity and the interpolation quality, a multi-head encoder was designed to separate embeddings for reconstruction and manipulation. In particular, after encoding the face image into an intermediate feature tensor, the encoder splits into two branches. In one branch, we compute a high-dimensional tensor with spatial properties called face icon for capturing the face outline information and reconstructing image details. In the other branch, a simple, fixed mask is devised over the intermediate feature tensor to extract separate one-dimensional component embeddings for flexible component editing. In the inference stage, a synthesis reasoning process is performed on component embeddings for various applications such as attribute manipulation and component transfer. To further disentangle highly correlated components (i.e., the left eye and the right eye), a component adaptive modulation (CAM) module is proposed for the decoder in order to modulate the variances of component-specific regions in each layer of the decoder. It is possible to employ CAM to achieve semantic single-eye editing without any input visual guidance for the first time. Moreover, semantic single-eye editing is still a challenging task due to the high correlations between the left and right eyes. Most existing face datasets lack faces with two different eyes. Therefore, it is difficult for neural networks \cite{stylespace,ganlocal} to learn the features of a single eye from these data sets.
	
	Experiments show that the high-dimensional face icon enhances the consistency between the reconstruction and the original image, while the component embeddings effectively capture the facial component's features and allow the users to independently edit the component. There is no need for an input mask or sketch from the user in the visual space, which liberates users from painting skills. Moreover, IA-FaceS can find interesting directions at the component level by implementing PCA in the latent space of a specific component, providing users more accurate editing of components and creating more facial expressions.

	\begin{table}
		\centering
		
		{\tabcolsep0.04in\begin{tabular}{c|c|c|c}
				\multicolumn{4}{c}{Mouth (row) vs Eyes (column)} \\ \midrule
				&Narrow  &Wide & Total\\ \midrule
				Open  &  494        & 337              & 831    \\ \midrule
				Closed &  262        & 419              & 681    \\ \midrule
				Total    &  756        & 756              & 1512   \\ \bottomrule
			\end{tabular}
			\hspace{4mm}
			\begin{tabular}{c|c|c|c}
				\multicolumn{4}{c}{Gender (row) vs Eyebrows (column)} \\ \midrule
				&Bushy & Arched &Total   \\ \midrule
				Male    &  895         &    107          & 1002   \\ \midrule
				Female  &  422         &   1210          & 1632   \\ \midrule
				Total   &  1317        &   1317          & 2634   \\ \bottomrule
			\end{tabular}			
		}
		\caption{The number of collected faces among attribute labels on $10,000$ faces randomly chosen from CelebA-HQ \cite{pg-gan}. Left: narrow or wide (no-narrow) eyes versus open or closed mouth. Right: bushy or arched eyebrows versus gender. From the left sub-table, \emph{narrow eyes} is significantly correlated with \emph{open mouth} (p-value =$7.41e^{-16}$, $\chi^2$ test). From the right sub-table, \emph{eyebrows} is significantly correlated with \emph{gender} (p-value=$5.8e^{-219}$, $\chi^2$ test). }
		\label{tab:intro-sample-bias}
	\end{table}
	
	The research contributions of this paper are summarized as below:
	\begin{itemize}
		\item A bidirectional method is proposed for continuous, disentangled face attribute manipulation and flexible component editing without the need for any additional guidance (e.g., masks or sketches) in the visual space. 
		\item  To maintain rich details in reconstruction and simultaneously perform interpolation in the latent space, the coding for reconstruction is separated from the coding for face manipulation. A high-dimensional tensor is employed for consistent reconstruction, and four separate low-dimensional vectors are utilized for attribute manipulation and component transfer. 
		\item The proposed model computes a separate embedding for each component and implements independent component editing by manipulating separate embeddings. Crucially, it neither relies on intervention experiments to find the corresponding representation for a specific component nor involves visual geometric guidance. 
		\item To further disentangle the highly correlated components (i.e., the left eye and the right eye), a component adaptive modulation (CAM) module is proposed for the decoder. To the best of our knowledge, the proposed model is the first latent manipulation method that can flexibly edit a sing eye without affecting the other one.
		\item Both qualitative and quantitative results demonstrate the effectiveness of the proposed framework and improvements in comparison with the state-of-the-art methods.
	\end{itemize}

	\section{Related Works}   \label{sec:related-work}
	
	\subsection{Generative Adversarial Networks}
	GAN \cite{gan} has achieved impressive results in a wide range of computer vision applications. It consists of two networks, the generator and the discriminator, where the two networks are trained in an adversarial way, forcing the generator to generate samples indistinguishable from the target distribution. Due to the high diversity and fidelity of the generated images, GAN has attracted extensive attention and research. The development of GANs has significantly improved the quality of the synthesized images \cite{big-gan,pg-gan, stylegan}. BigGAN \cite{big-gan} and PGGAN \cite{pg-gan} can generate high-quality and diverse images from a simple distribution, which forms the structure called the latent $Z$ space. StyleGAN \cite{stylegan} further converts the latent code $z\in \mathcal{Z}$ to the mapped style code $w$ in another intermediate latent space $\mathcal{W}$, which contains more disentangle features \cite{stylegan} and performs better in separability \cite{interfacegan}. 
	
	\subsection{Face Attribute Manipulation}
	Face attribute manipulation is one of the most popular research topics in image processing due to its various practical applications. Depending on the different pipelines, face manipulation methods can be roughly divided into two main streams:  GAN's methods and the ones that fall into the encoder-decoder-discriminator paradigm. The first stream usually leverages the captured semantics in GAN's latent space after the generator is trained \cite{radford2015unsupervised,interfacegan,ganspace}. Although GAN's methods produce some impressive results, they require another computational inversion module to project an image back into the latent space \cite{stylegan,indomain,image2stylegan,image2stylegan++}. In contrast, the second stream adopts an encoder for efficient image projection \cite{resgan,diat,fader,sagan,icgan,stargan,starganv2,alae,stylemapgan,cafe-gan,relgan,stgan}, but due to the low-dimensional bottleneck layer (i.e., the innermost feature map with minimal spatial size), the encoder-decoder structure often faces the problem of inaccurate reconstruction \cite{stargan,starganv2,alae}. Recent works often treat attribute manipulation as an image-to-image translation task \cite{resgan,sagan,stargan,starganv2,l2mgan,cafe-gan,relgan,stgan} and learn to synthesize an output image directly according to user input. Despite promising results, most image-to-image translations \cite{attgan,stargan,starganv2,l2mgan,cafe-gan,relgan,stgan} can only edit the predefined attributes before training, limiting their flexibility in the inference stage.
	
	A key issue to consider is the entanglement between different attributes. Much research has been performed on the design of loss functions \cite{chen2016infogan,odena2017conditional,tran2017disentangled} or architectures \cite{shen2018faceid,donahue2017semantically}, locating attribute-specific regions \cite{resgan,diat,sagan,semantic-decomposition}, using additional 3D priors \cite{deng2020disentangled} or other additional annotation labels \cite{qian2019make,freenet,ganimation}. However, these methods either are not flexible enough or require additional visual guidance. There are also investigations on finding disentangled directions in GAN's latent space, including supervised methods \cite{goetschalckx2019ganalyze,interfacegan,abdal2020styleflow,Jahanian*2020On} and unsupervised methods \cite{voynov2020unsupervised,ganspace,spingarn2021gan,closed-form}. Although some of them can independently manipulate the light, pose, identity \cite{interfacegan,deng2020disentangled}  for a face image and successively disentangled some attributes \cite{interfacegan,closed-form,gan-parameter}, they still cannot achieve disentanglement between different components. This is due to the intrinsic correlation between components and the way these methods encode a face, i.e., embedding a whole face into a single vector in the latent space. Compared to existing methods, the proposed method in this paper allows flexible manipulation over each component by computing embeddings separately for every facial component.

	\subsection{Local Face Editing}
	To facility flexible local edits, a line of recent works utilize geometric guidance provided by users, e.g., mask-guided editing \cite{mask-guided,celebahq-mask, sean} or sketch-guided editing \cite{sangkloy2017scribbler,portenier2018faceshop, deepfacedrawing}.  The mask-guided portrait editing (MGPE) method \cite{mask-guided} computes feature embedding for each component in the face by multiple encoders and requires users to segment the components in the original image. The masking requirement is relaxed by a local face component mask in the r-FACE model \cite{r-face}, which uses an example-guided attention module to concentrate on the target component. Studies have also focused on a specific component in faces \cite{eyeopener, eyeinpainting}, but cannot generalize to entire areas of the face. 
	Instead of masking in the original image, the proposed method employs a simple, fixed mask in the latent feature space for separate component embeddings and does not require an attention module to focus on specific component regions. The proposed method also allows the users to edit abstract attributes, e.g., gender, while geometry-guided methods mainly focus on attributes related to geometric changes \cite{ganimation,make-a-face}.
	
	Another line for spatial disentanglement resorts to explore representation for a specific semantic in the latent space \cite{gan-dissection}, feature space \cite{ganlocal}, style space \cite{style-intervention,stylespace}, or parameter's space \cite{gan-parameter} of a pre-trained GAN. For example, GAN dissection \cite{gan-dissection} tries to interpret the internal representations of GANs and find a cause-effect relationship between the top units and objects, guided by supervision from an independent semantic segmentation model. However, considerable intervention experiments are required to find the cause-effect relationship. Besides, a neuron can control multiple components, which may lead to unexpected object entanglement. Another unsupervised approach \cite{ganlocal} identifies meaningful activations through simple k-means clustering on feature space. Recently, as an increasingly popular generative model,  StyleGAN \cite{stylegan} designs channel-wise \emph{style codes} to control the statistics at each of the generator’s convolution layers. The \emph{style codes} compose a \emph{style space} and attract researchers \cite{style-intervention, stylespace} to  explore the \emph{style space}   of StyleGAN \cite{stylegan} for highly disentangled and localized visual attribute. However, the disentanglement is restricted to what semantics \cite{stylespace,ganlocal} they can find. For areas  where the corresponding representation cannot be found in \emph{style space} \cite{stylespace} or feature space \cite{ganlocal} (e.g., a single-eye), they cannot be edited.
	
	Distinguished  from all these works, the proposed approach implements independent component manipulation by controlling their separate embeddings. Crucially, it does not rely on intervention experiments to find the corresponding representation for a specific component in a large search space, nor does it involve geometric guidance or a complicated attention module. Last but not least, the proposed method is the first latent space manipulation method that can flexibly edit a single eye without affecting the other one.
	
	The method most relevant to us is StyleMapGAN \cite{stylemapgan}, a recent encoder-decoder method for local face editing. It embeds the face into a tensor with explicit spatial dimensions, called \emph{stylemap}, which enables the model to reconstruct the details of an image and manipulate the relevant regions in the \emph{stylemap} to achieve local edits.  Although that we also have a high-dimensional tensor with spatial properties for reconstruction, the face editing is controlled by the low-dimensional component embeddings instead of the high-dimensional tensor. We show that direct editing of the high-dimensional tensor with spatial properties will introduce artifacts due to the lack of semantic understanding of the whole face.

	\section{Preliminaries}
	\subsection{Deep yIng-yAng Bidirectional Intelligence}
	Deep yIng-yAng Bidirectional Intelligence (IA-BI) \cite{xu-bidirectional} was proposed to learn the synthesis space from data and derive reasoning by data or cause, or both. Under the framework of IA-BI, two domains are defined: A-domain, denoting the real bodies or patterns in the actual world, and I-domain, indicating the inner coding domain. The IA-BI is featured by an alternation of A-mapping (from A-domain into I-domain) and I-mapping (from I-domain into A-domain). A-mapping abstracts the input data into a compressed code in I-domain, while I-mapping implementing inference and synthesis. Readers are referred to Figure 14 in \cite{xu-bidirectional} for an illustration of the IA-BI system, as well as a systematic review and more details. 
	
	\subsection{Deep Synthesis Reasoning}
	The synthesis reasoning model was proposed early in \cite{pan}, which recombines different sources in the synthesis space and integrates the characteristics of each source to obtain some creative designs and imaginations. For completeness, we restated the formulations defined in Section 3 in \cite{pan} as follows.
	A \emph{source} $T$ can be expressed as $T=(P,m,F)$, where $P$ is the set of the components $P=\{P_1,P_2,P_3,P_n\}$ compose $T$. $m$ describes how the $n$ components construct the whole object. $F$ is the influence field of each component to the whole object  $T$. 
	The synthesis space is denoted as $SS (x, y, z)$, where each position is a potential synthesis result.
	Specifically, the space composed of M sources is defined as
	$S S(x, y, z)=\sum_{j=1}^{m} \sum_{i=1}^{n}\left(F P_{i j}(x, y, z) \cdot P_{i j}, F m_{j} \cdot m_{j}\right).$	
	The initial synthesis reasoning mainly occurred in the visual space, requiring artificial construction of individuals, source fields, etc., which hindered its application and development. Based on the Deep IA-BI, Xu \cite{xu-reasoning} proposed a technique to extend the synthesis reasoning from visual space to the latent space. We can establish the synthesis space in the top-level coding space, using the code as the source, to achieve deep synthesis reasoning. In such a way, we can avoid building the synthesis space by man-crafted sources. 
	
	\section{Methods}  \label{sec:method}
	This section describes the proposed method for disentangled face attribute manipulation and component editing. To begin, we describe the motivations and overall framework of the proposed method (Section \ref{sec:me-overview}). Then, we introduce the proposed method in detail, including the multi-head structure for the encoder (Section \ref{sec:me-encoder}) and the component adaptive modulation module for the decoder (Section \ref{sec:me-decoder}). Finally, we describe how we apply synthesis reasoning over the latent space to conduct multiple applications, e.g., attribute editing, component transfer (Section \ref{sec:me-synthesis}). 
	\subsection{Overview} \label{sec:me-overview}
	Existing methods usually project the input image to a single low-dimensional embedding ($1\times 512$ vector in ALAE \cite{alae} and StyleGAN2 \cite{stylegan2}), which is easy to perform linear interpolation, but has two limitations: 1) embedding low dimensions loses the original image details and limits the reconstruction capacity of the entire network.
	2) The whole face single-embedding manner limits the flexibility of editing facial components, e.g., mouth, eyes, and nose.
	
	To address the above issues, we propose IA-FaceS, a bidirectional method for disentangled attribute manipulation and flexible component editing. Following the Deep IA-BI framework (See Figure 14 in \cite{xu-reasoning}), IA-FaceS is featured by an encoder network (A-mapping) to compute latent representation for input face, a synthesis reasoning process over the latent space for face manipulation, and a decoder network (I-mapping) for high-quality synthesis output. By separating codes for reconstruction and face manipulation, IA-FaceS balances well between reconstruction capacity and control flexibility. 
	
	
	\begin{figure}[!ht]
		\begin{center}
			\includegraphics[width=\textwidth]{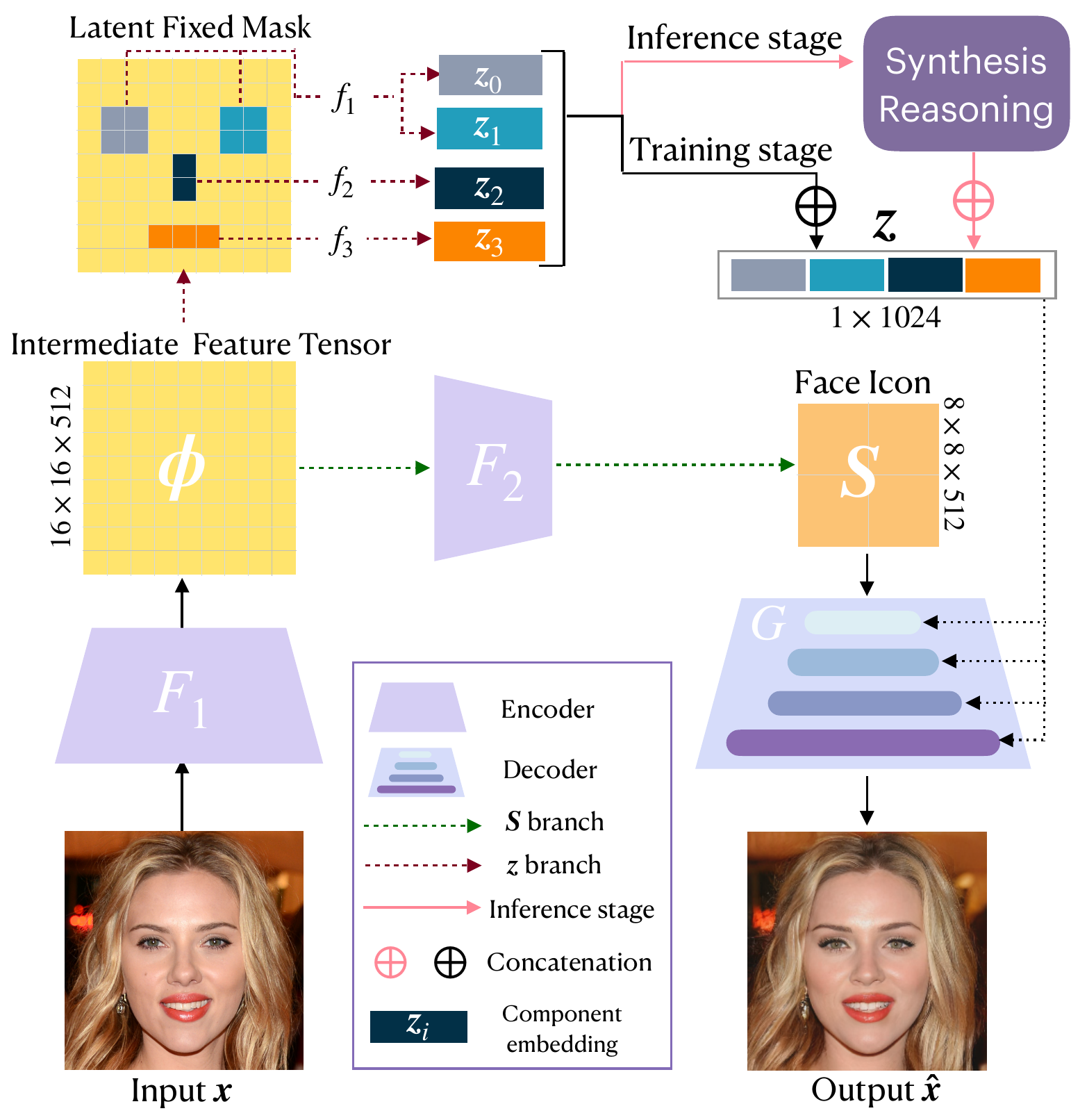}
			\caption{An overview of IA-FaceS, which is composed of a multi-head encoder, a synthesis reasoning process, and a decoder.}
			\label{fig:me-framework}
		\end{center}
	\end{figure}
	
	\subsection{Multi-head Encoder}  \label{sec:me-encoder}
	There exists a contradiction between reconstruction and face manipulation. Specifically, maintaining rich details in reconstruction requires the latent code to be a high-dimensional tensor with spatial properties. However, face manipulation is often performed by interpolating the latent code, while interpolations are difficult on high-dimensional tensors. To balance reconstruction with manipulation, we design a multi-level and multi-head encoder to obtain the embeddings for reconstruction and manipulation respectively.
	
	Let $\mathcal{X}$ be the visual space of observed face images. We define the input face $\boldsymbol{x}\in\mathcal{X}$ as a combination of the face outline and four semantic components $c_i$, $i \in \{0,1,2,3\}$, i.e., left eye, right eye, nose, and mouth. As illustrated in Figure \ref{fig:me-framework}, the mapping network $F_1$ encodes the input image $\boldsymbol{x}$ into an intermediate feature tensor $\boldsymbol{\phi}$ of size $c\times h\times w$.
	Subsequently, we develop a multi-head structure, namely Facial Component Embedding Branch and Face Icon Branch. The former is used for extracting separate component embeddings (red dashed line), and the latter computes a high-dimensional tensor $\boldsymbol{S}$ with a smaller spatial size than $\boldsymbol{\phi}$ (green dashed line). 
	\paragraph{Facial Component Embedding Branch}
	Each facial component has a fixed position in a face, and the convolutional layers preserve the relative locations of the features. This nature enables us to extract component embeddings $\boldsymbol{z_i}$ from their corresponding values of the intermediate feature map $\boldsymbol{\phi}$. As shown in Figure \ref{fig:me-framework}, a simple and fixed mask is used to segment the intermediate feature tensor  $\boldsymbol{\phi}$, and component-specific layers are then introduced to compute the component embeddings. Specifically, we calculate the embedding for the $i$-th component as follows, 
	\begin{equation}
		\boldsymbol{z_i} = f_i(\boldsymbol{\phi}[b_i]),
	\end{equation}
	where $[\cdot]$ is to take the values inside the box $b_i$ across the channels in $\boldsymbol{\phi}$, $f_i$ is a component-specific fully-connected layer with non-linear activation. The latent box $b_i$ is resized from the predefined component region in visual space. Let $B_i$  defines the area occupied by the component $c_i$ on the face, which is fixed for all the input face images. Then, we re-scale $B_i$ to fit the size of $\boldsymbol{\phi}$ as follows,
	\begin{equation}
		b_i = \lceil \frac{B_i - \frac{r}{2}}{r} \rceil,\, r=\frac{H}{h}
		\label{eq:rescale}
	\end{equation}
	where $\lceil \cdot \rceil$ indicates rounding up, $r$ is the ratio of the image resolution $H$ to the feature resolution $h$. The spatial distribution of $B_i$ and $b_i$ are visualized in Figure \ref{fig:sup-inputs} of the supplementary material.
	
	In a recent method for component editing, i.e., MGPE \cite{mask-guided}, the components are segmented by a segmentation mask in the visual space and the component features are calculated by five separate local encoders. Distinct from MGPE, we segment the components by a simple, fixed mask in the feature map and enhance the representation learning by sharing the bottom layers of the encoder.
	As a segmentation-based method, the component features in MGPE can only control the textures of components, while the shape and location of the components are decided by the target mask. In contrast, both the shape and textures of a component are controlled by component embeddings in our method. Besides, the proposed model allows the users to edit global abstract attributes (e.g., gender), which is difficult to be described by a segmentation mask.
	\paragraph{Face Icon Branch}
	In the other branch (green dashed line, in Figure \ref{fig:me-framework}), the encoding network $F_2$ preserves the face topology by computing the face icon $\boldsymbol{S}=F_2(\boldsymbol{\phi})$, i.e., a high-dimensional feature tensor with spatial properties that captures the face outlines and helps connect the facial components. According to the experiments, the high-dimensional face icon is better than a low-dimensional vector for assisting image reconstruction. Note that the resolution of the face icon $\boldsymbol{S}$ will significantly influence the performance of the model. A larger face icon can improve reconstruction accuracy, but also competes with the facial component branch for capturing component-specific features. As a result, the component embeddings would be ignored in face synthesis. A detailed analysis will be given in Section \ref{sec:exp-ablation}.
	
	\subsection{Component Adaptive Modulation} \label{sec:me-decoder}
	\begin{figure}[t]
		\subfigure[StyleGAN2]{
			\centering
			\includegraphics[height=0.52\columnwidth]{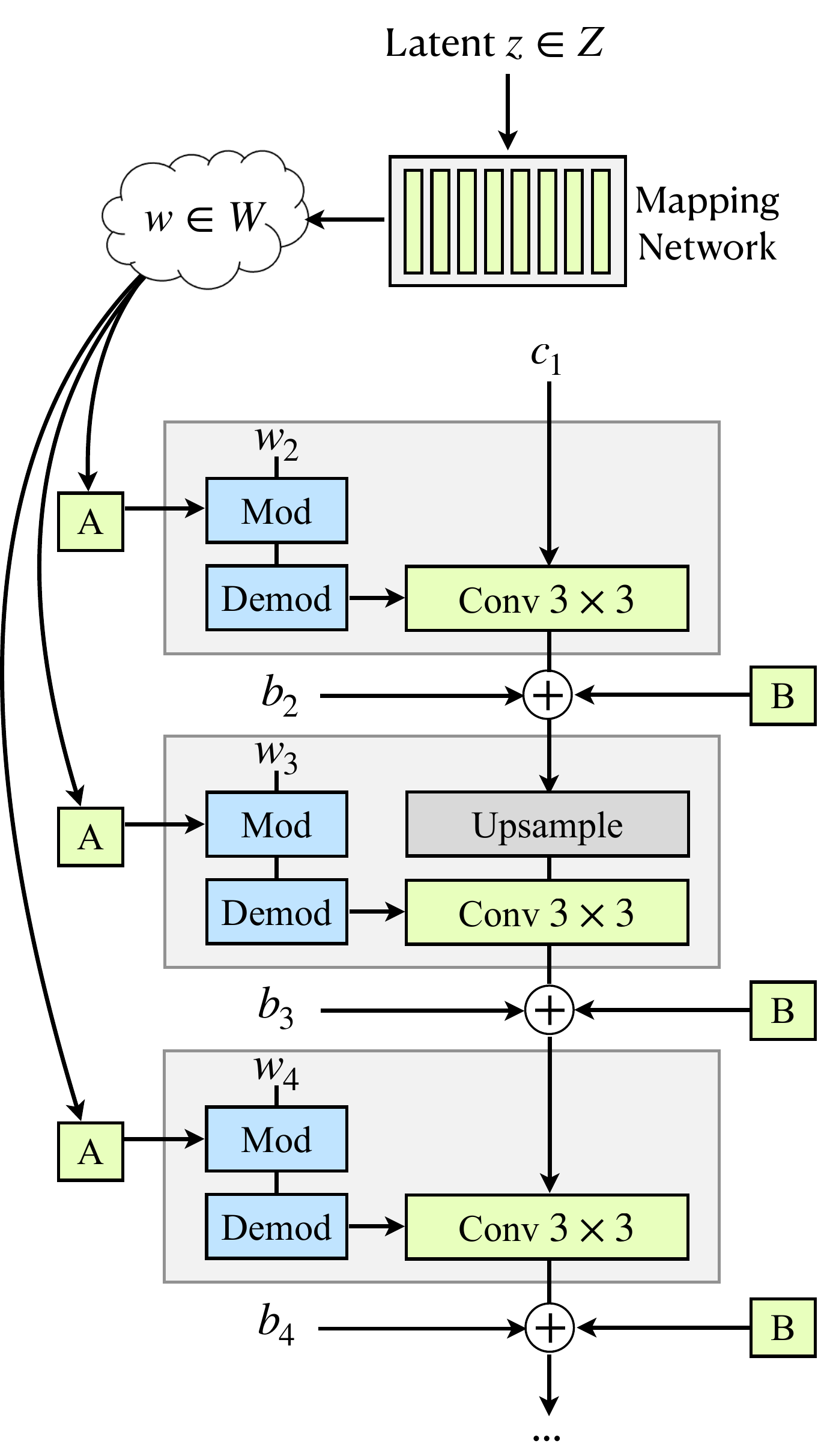}
			\label{fig:me-stylegan2}
		}
		\subfigure[IA-FaceS]{
			\centering
			\includegraphics[height=0.52\columnwidth]{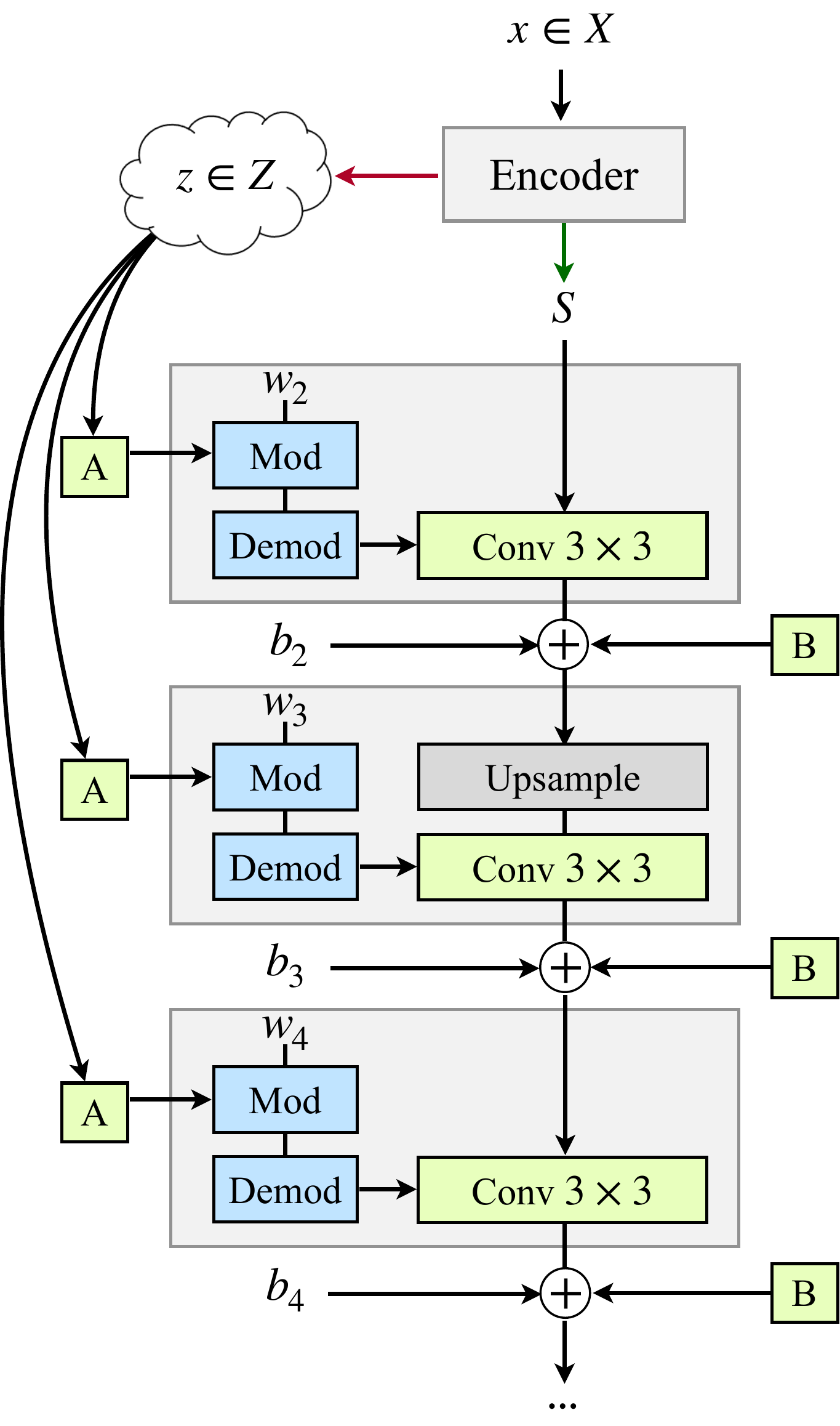}
			\label{fig:me-iafaces}
		}
		\subfigure[IA-FaceS+CAM]{
			\centering
			\includegraphics[height=0.52\columnwidth]{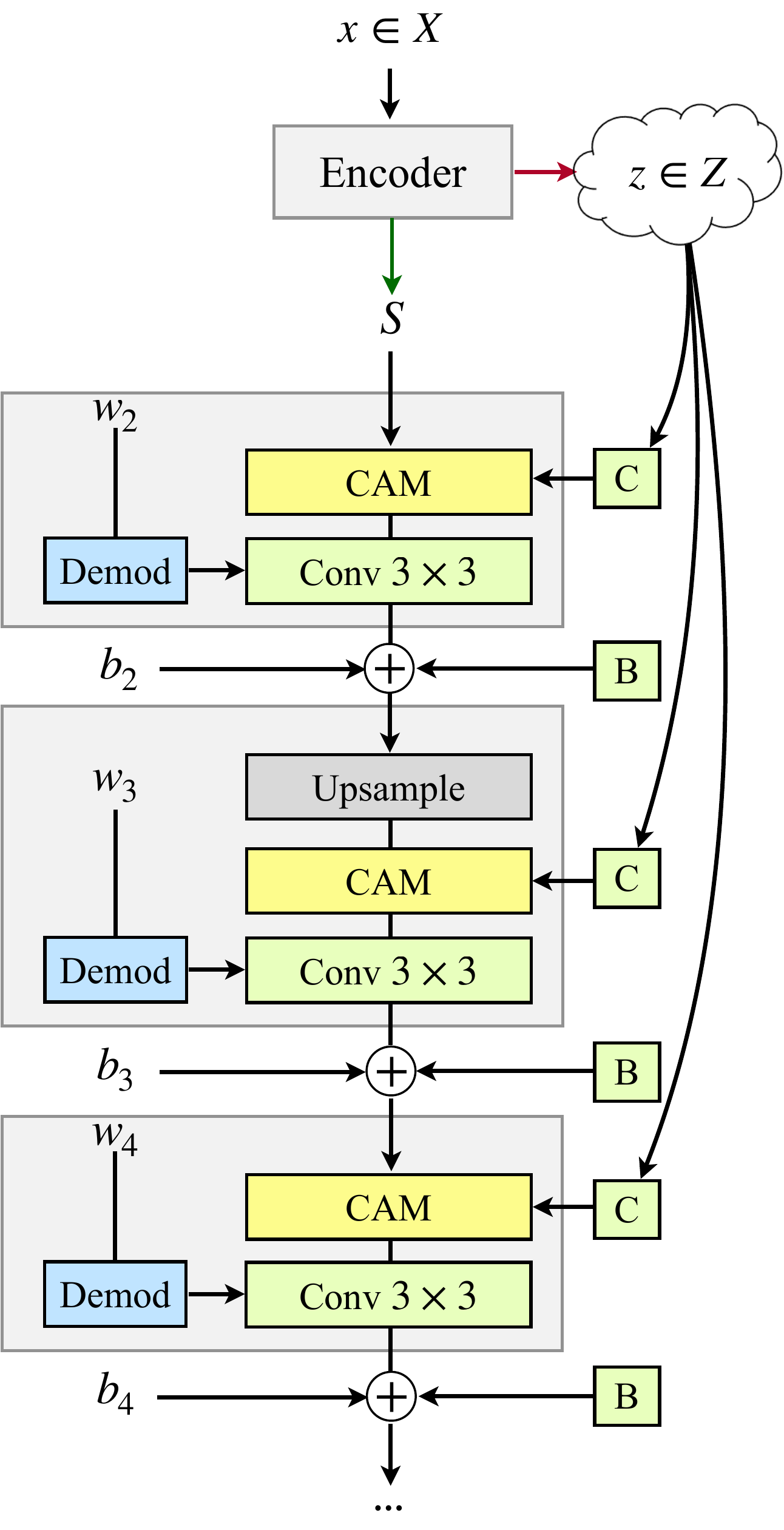}
			\label{fig:me-cam}
		}
		\caption{The architectures of the decoders. (a) Style-based generator in StyleGAN2 \cite{stylegan2}, \boxed{A} denotes a learned affine transform from $z$ that produces a style and \boxed{B} is a noise broadcast operation. (b) The decoder in IA-FaceS. (c) The decoder of IA-FaceS+CAM, where the global modulation is replaced with CAM, \boxed{C} denotes learned component-specific affine transforms. }
		\label{fig:me-generator}
	\end{figure}
	
	For the decoder $G$ in IA-FaceS, the backbone of StyleGAN2 \cite{stylegan2} is adopted. As illustrated in Figure \ref{fig:me-iafaces}, the decoder takes $\boldsymbol{S}$ as the input and uses the concatenation of component embeddings $\boldsymbol{z}$ to predict style parameters for each layer through a learned affine transform \boxed{A}. The affine transform specializes $\boldsymbol{z}$ to per-layer \emph{style} that controls the channel-wise variances by modulating the convolutional kernel weights. Note that the decoder in IA-FaceS differs from  StyleGAN2 (see in Figure \ref{fig:me-stylegan2}) in the following aspects: 1) the decoder in IA-FaceS starts from the face icon $\boldsymbol{S}$, while StyleGAN2 starts from a learned constant tensor $c_1$. 2) IA-FaceS uses the component embeddings $\boldsymbol{z}$ to predict \emph{styles} for each layer, while StylaGAN2 first maps the latent code to an intermediate latent space $\mathcal{W}$, then uses $w \in \mathcal{W}$ to predict the \emph{styles}. 
	
	In the inference stage, when editing a specific component, one can manipulate its corresponding embedding $\boldsymbol{z_i}$ while keeping $\boldsymbol{S}$ unchanged and then synthesize the output with the renewed $\boldsymbol{z}$. This scheme works well for independently editing eyes, nose, and mouth. However, due to the limitation of the backbone adopted from StyleGAN2, IA-FaceS cannot well disentangle the highly correlated components, e.g., changing the left eye and the right eye respectively. Specifically, the \emph{style} in StyleGAN2 is globally applied on the channel-wise feature maps. Therefore,
	once no feature map in the generator corresponds to the given component, the component cannot be edited.	We follow the steps in \cite{style-intervention} to visualize the top-5\% activations of feature maps in the decoder of IA-FaceS, and find that no feature map corresponds to a single eye. Based on the above analysis, the global modulation in StyleGAN2 and IA-FaceS limits the flexibility of controlling a single eye without affecting the other one.
	
	
	

	To further disentangle the highly-correlated components, we propose a component adaptive modulation (CAM) module to predict component-specific per-layer style parameters for each component, and then apply the component-specific \emph{style} to modulate the variances of component regions on each feature map. Considering a batch consisting of a single image, let $H^k\in \mathcal{R}^{(C\times H \times W)}$ be the feature maps of layer $k$. Then, the original box $B_i$ is re-scaled to $b^k_i$ to fit the size of $H^k$ following Eq. (\ref{eq:rescale}). Here, $b^k_i$ acts as a fixed latent mask for layer $k$ to segment the facial component $c_i$ in $H^k$. CAM works as follows:
	\begin{equation}
		H^k[b^k_i] = H^k[b^k_i] \odot \sigma_i,\quad \sigma_i =   f_i(\boldsymbol{z_i}),
	\end{equation}
	where $\odot$ denotes channel-wise multiplication,  $[\cdot]$ is to take the values inside the box $b^k_i$ across the channels in $H^k$, $f_i$ is a component-specific affine transformation which transforms component embedding into a style scaling coefficient $\sigma_i \in \mathcal{R}^{(C\times 1)}$ vector.
	Utilizing the CAM module, IA-FaceS+CAM can disentangle highly-correlated components and achieve flexible controls over different eyes (for example, one eye open, and the other close).
	
	\subsection{Deep Synthesis Reasoning} \label{sec:me-synthesis}
	\begin{figure}[!ht]
		\subfigure[Component transfer.]{
			\centering
			\includegraphics[width=\columnwidth]{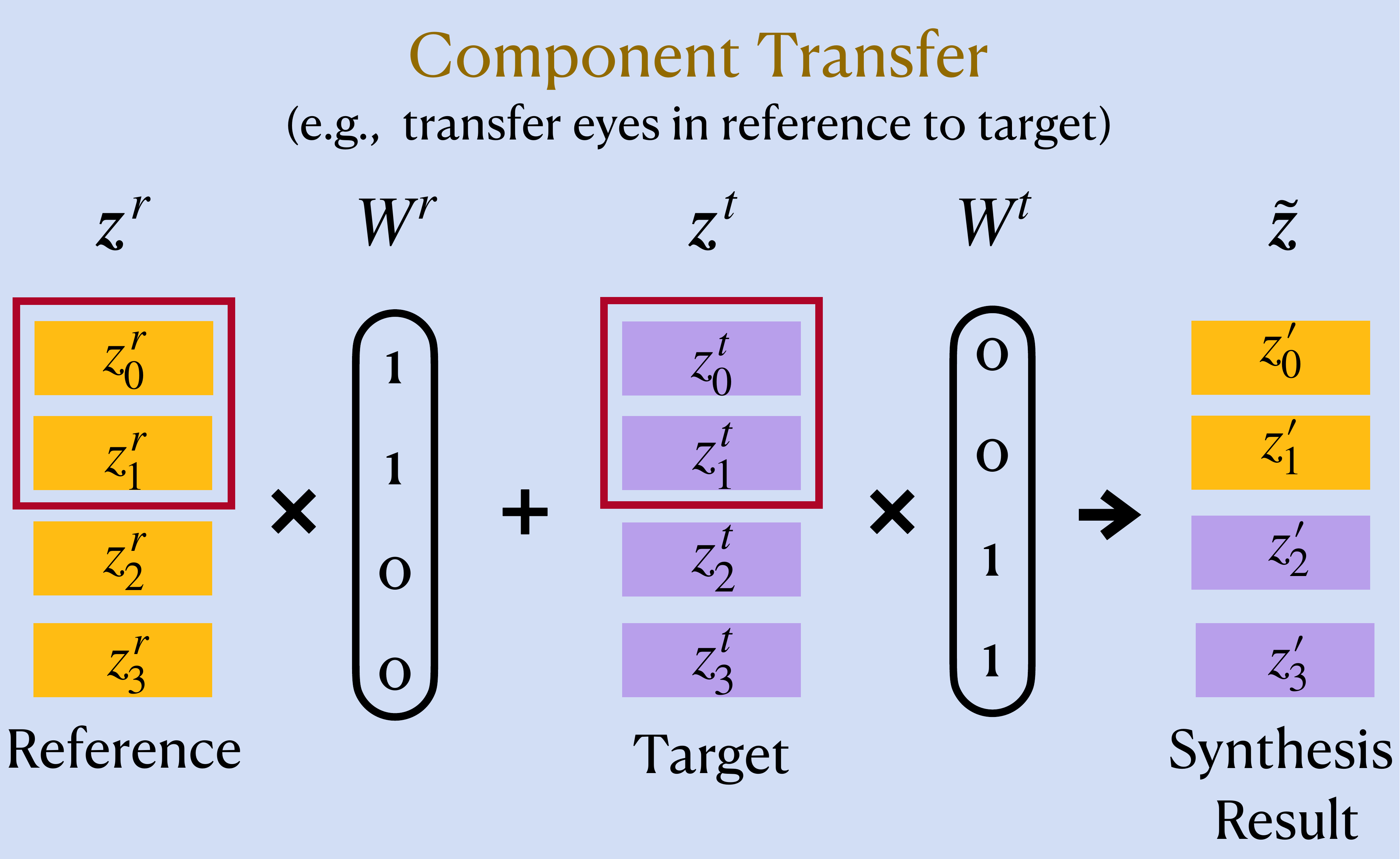}
			\label{fig:me-synthesis-comtrsf}
		}
		\subfigure[Attribute manipulation.]{
			\centering
			\includegraphics[width=\columnwidth]{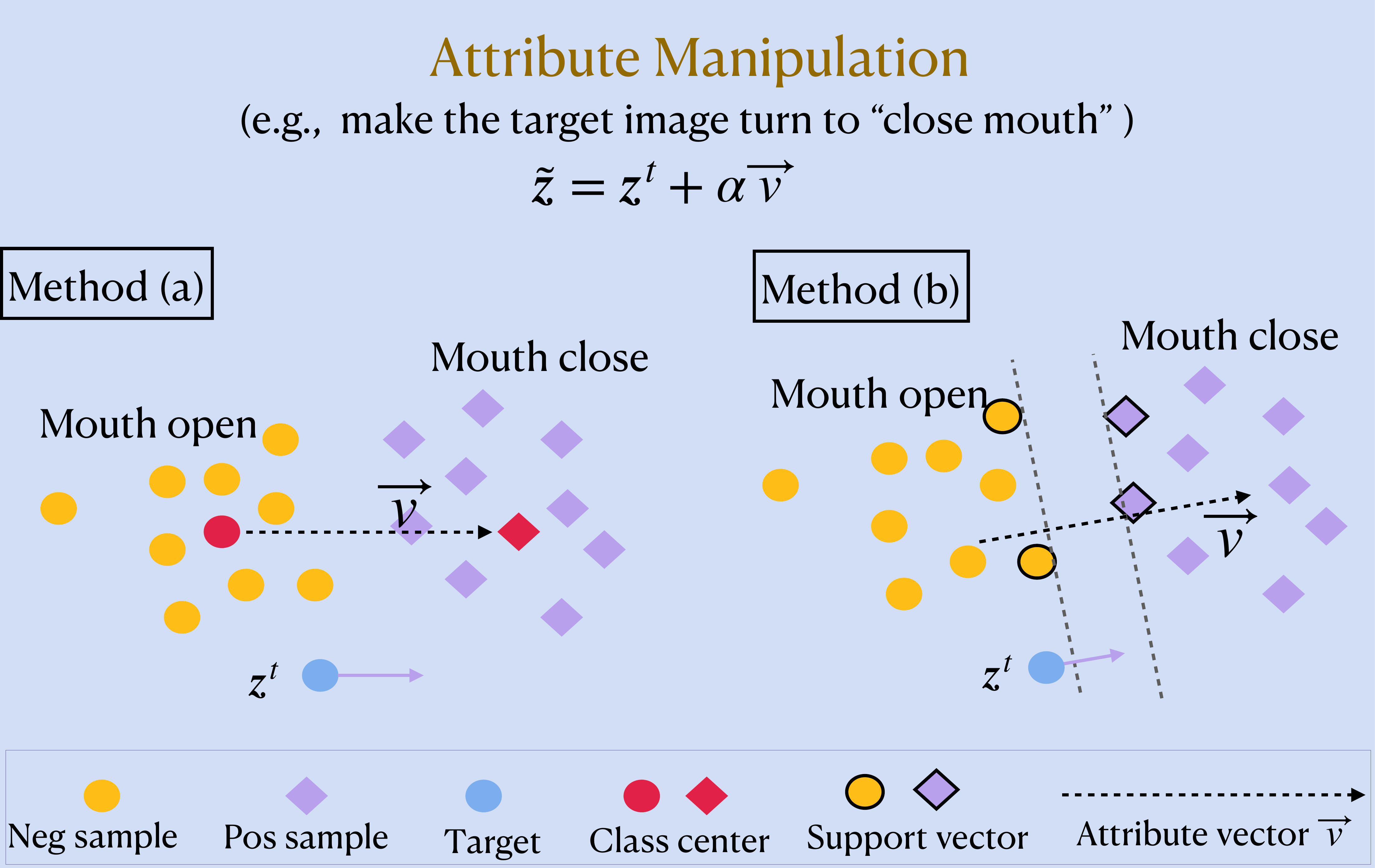}
			\label{fig:me-synthesis-attr}
		}
		\caption{Examples of synthesis reasoning.}
		\label{fig:me-synthesis}
	\end{figure}
	
	This section introduces how to implement deep synthesis reasoning \cite{xu-reasoning} based on the component embeddings $\{\boldsymbol{z_i}\}|_{i=0}^3$  
	for component transfer and face attribute manipulation.
	We simplify the source definition in Pan's work \cite{pan} and formulate synthesis results of $n$ faces as: 
	\begin{equation}
		\label{eq:syn-infer}
		\Tilde{\boldsymbol{z}} = [\Tilde{\boldsymbol{z}}_j],\; \Tilde{\boldsymbol{z}}_j = \sum_{i=0}^n{W^i_j\, \boldsymbol{z}^i_j},
	\end{equation}
	where $\boldsymbol{z}^i_j$ denotes the $j$-th component embedding in $i$-th source, $W^i_j$ denotes the influence of $\boldsymbol{z}^i_j$ to the final synthesis results $\Tilde{\boldsymbol{z}}$. Here, Eq. (\ref{eq:syn-infer}) defines possible linear combinations of different sources, while the general combination function could be non-linear. Given different settings for $W^i_j$, we can achieve different synthesis results and produce diverse creations. As demonstrated in Figure \ref{fig:me-synthesis}, synthesis reasoning includes some typical face manipulation methods as special cases, such as component transfer, attribute manipulation, and interpolation between two or multiple images. 
	\begin{itemize}
		\item Component transfer is a special case of Eq.~(\ref{eq:syn-infer}). As given in Figure \ref{fig:me-synthesis-comtrsf}, when transferring the eyes in a reference image $\boldsymbol{x}^r$ to a target image $\boldsymbol{x}^t$, we can set $W^r= \{1,1,0,0\}$, $W^t=\{0,0,1,1\}$ in Eq. (\ref{eq:syn-infer}). 
		\item Some supervised attribute manipulation methods can be viewed as special cases of Eq (\ref{eq:syn-infer}). For attribute $v$,  $\{\boldsymbol{z}^i \text{,} y^i\}|_{i=1}^n$ are the collected positive and negative samples, where $\boldsymbol{z}^i$ is the latent embedding for a source image $\boldsymbol{x}^i$, and $y^i \in\{+1\text{,}-1\}$ is the label indicating $\boldsymbol{x}^i$ own attribute $v$ (+1) or not (-1). The attribute vector $\vec{v}$ can be calculated in two ways.\\
		(a) $\vec{v} = \frac{1}{n}\sum_{i=1}^{n}{y^i\boldsymbol{z^i}}$. In this case, Eq. (\ref{eq:syn-infer}) is expressed as
		\begin{equation}
			\label{eq:syn-attmeth-a}
			\Tilde{\boldsymbol{z}} =\boldsymbol{z}^t+\alpha\vec{v}=\boldsymbol{z}^t+\sum_{i=1}^{n}\alpha y_i \boldsymbol{z}^i.
		\end{equation}
		(b) $\vec{v} = \sum_{i=1}^{n}{\eta_i y_i \boldsymbol{z^i}}$, where $\eta_i$ is the optimum solution of a support vector machine (SVM) classification problem with details described in \ref{sec:sup-svm}. In this case, Eq. (\ref{eq:syn-infer}) is expressed as:
		\begin{equation}
			\label{eq:syn-attmeth-b}
			\Tilde{\boldsymbol{z}} = \boldsymbol{z}^t+\alpha \vec{v}=\boldsymbol{z}^t+\sum_{i=1}^{n}{\alpha\eta_i y_i\boldsymbol{z}^i}.
		\end{equation}
	\end{itemize}
	
	The above attribution manipulation by Eq.~(\ref{eq:syn-attmeth-b}) is equivalent to the one in InterFaceGAN \cite{interfacegan} which only focuses on a single latent embedding. Instead, in IA-FaceS, $\boldsymbol{z}$ is usually composed of multiple components, and thus we can calculate attribute vector in a subset of $\boldsymbol{z}$.
	Given attribute $v$, for each sample $\boldsymbol{z^i}$, we can split it into $Z_a$ and $Z_{-a}$, where $Z_a=\{\boldsymbol{z}_i|i \in C_v\}$, $Z_{-a}=\{\boldsymbol{z}_i|i \notin C_v\}$, where $C_v$ denotes the components relevant to the attribute $v$. Under this setting, $	\Tilde{\boldsymbol{z}}= \boldsymbol{z}^t+\alpha \vec{v}$, where $\vec{v} = \{\vec{v}_a, \vec{0}_{-a}\}$.
	
	\subsection{Training Procedure and Losses} \label{sec:me-objective}
	We use $F$, $G$,  and $D$ to indicate the encoder network, decoder network, and discriminator network respectively. $D$ is the same as StyleGAN2 \cite{stylegan2}. $G$ follows the settings in StyleGAN2 \cite{stylegan2} except for the number of channels. $F$ is stacked by residual blocks \cite{he2016deep} and the detailed structure is described in \ref{sec:sup-network}. All networks are jointly trained from scratch using reconstruction loss and adversarial loss.
	
	\paragraph{Reconstruction Loss}
	$G$ and $F$ are trained to reconstruct real images in terms of both pixel-level ($L_1$ loss) and perceptual-level \cite{lpips}. To improve the smoothness of the underlying latent space, we perturb the input image $\boldsymbol{x}$ by randomly masking some points, as described below:
	\begin{equation}
		\label{eq:mask}
		\boldsymbol{x^{in}} =\left\{
		\begin{aligned}
			&  \boldsymbol{x}, \rm\delta=0; \\
			&  \boldsymbol{x}* \boldsymbol{M}, \rm\delta=1; \\
		\end{aligned}
		\right.
	\end{equation}
	where $\delta$ follows a Bernoulli distribution with the probability $Pr(\delta =1)=0.5$, the mask $\boldsymbol{M}$ is created by randomly drawing irregular shapes\footnote{\url{https://github.com/MathiasGruber/PConv-Keras}} (examples of masked images are given in Figure \ref{fig:sup-inputs} in \ref{sec:sup-perturb-input}). Then, the randomly perturbed images $\boldsymbol{x^{in}}$ are used as input to the encoder for the reconstruction of the original image $\boldsymbol{x}\in\mathcal{X}$. The Bernoulli-modify method acts like noise injection to regularize the model training and make the model robust to various inputs. 
	To sum up, the reconstruction loss is defined as:
	\begin{equation}
		\label{eq:recon-loss}
		L_{recon}(F,G) = \mathbb{E}_{x\sim P_{data}}(\left\|\boldsymbol{x}-\boldsymbol{\hat{x}}\right\|_1+\left\|V(\boldsymbol{x}),V(\boldsymbol{\hat{x}})\right\|_1),
	\end{equation}
	where $\mathbb{E}[\cdot]$ denotes the expectation, $\boldsymbol{\hat{x}}=G(F(\boldsymbol{x^{in}}))$ is the reconstruction results, $V(\cdot)$ denotes the feature vector outputted by a pre-trained VGG-16 \cite{Simonyan15} model.
	
	\paragraph{Adversarial Loss}
	For generating sharper images and textures, adversarial loss is introduced: 
	\begin{equation}
		\label{eq:adv-loss}
		L_{adv}(G,D) =  \mathbb{E}_{\boldsymbol{x}\sim  p_{data}}[f(D(\boldsymbol{x}))+ 
		f(1-D(\boldsymbol{\hat{x}}))].
	\end{equation} 
	
	The overall objective of IA-FaceS is to minimize the following min-max loss:
	\begin{equation}
		\label{eq:total-loss}
		\begin{aligned}
			\min_{F,G}\max _{D}L(D,F,G)=  L_{recon}(F,G) + L_{adv}(G,D).
		\end{aligned}
	\end{equation}

	\section{Experiments} \label{sec:experiments}
	In this section, we first describe the experimental setup (Section \ref{sec:exp-settings}) as well as the evaluation metrics used in quantitative comparisons (Section \ref{sec:exp-metrics}). Then, we compare the proposed model with leading methods in image projection (Section \ref{sec:exp-reconstruction}), face attribute manipulation  (Section \ref{sec:exp-attr}), and component editing (Section \ref{sec:exp-component}). What's more, we show the results of single-eye editing by the proposed CAM (Section \ref{sec:exp-single-eye}) and evaluate the performance under challenging conditions (Section \ref{sec:exp-generalization}).  Finally, we perform a complete analysis of our model (Section \ref{sec:exp-ablation}). Based on the analysis, we explain the limitations of the proposed method and show some failure cases (Section \ref{sec:exp-limitation}).
	\subsection{Settings} \label{sec:exp-settings}
	\paragraph{Datasets}
	We mainly evaluate the proposed model on CelebA-HQ \cite{pg-gan} and FFHQ \cite{stylegan} datasets, where most of the qualitative results on FFHQ dataset are shown in the appendix. Additional qualitative evaluations are performed on Cartoon Faces used in Toonify \cite{toonify}.
	\begin{itemize}
		\item CelebA-HQ \cite{pg-gan} contains 30,000 high quality facial images, which are picked from CelebA \cite{celeba} dataset and processed to $1024\times 1024$ resolution. Inherited from the original CelebA, each face in CelebA-HQ has 40 attributes annotations, where some annotations are used in the inference stage. We split the $30,000$ images into $27,000$ for training and $3,000$ for testing. All the images are resized to $256\times 256$, which is the resolution used by most previous works.
		\item FFHQ \cite{stylegan} is originally created as a benchmark for StyleGAN \cite{stylegan}. The dataset consists of $70,000$ high-quality human faces at $1024\times1024$ resolution. We take $65,000$ images for training, $2,000$ images for assessing the training process and the rest $3,000$ images for testing. 
		\item Cartoon Faces \cite{toonify} in Toonify includes 317 faces obtained from online images. All the images are aligned following the face alignment
		procedure used in the FFHQ \cite{stylegan} dataset.
	\end{itemize}
	
	\paragraph{Implementation details}
	Our end-to-end network is trained on PyTorch with two GeForce GTX TITAN V GPUs of 12 GB memory, which takes about 3 days to train ($256\times 256$ resolution). At training, we use Adam \cite{kingma2014adam} optimizer with a batch size of 16 and learning rate of 0.002, $\beta_1=0$,  $\beta_2=0.99$ for both discriminator and encoder-decoder. We use the same training procedure as in SwapAE \cite{swapae} and train the networks for 200 epochs (about 337k iterations). The encoder is composed of Residual Blocks and the decoder is adapted from StyleGAN2. Readers can refer to \ref{sec:sup-implementation-our} for detailed network structure and implementations. 
	
	\paragraph{Baselines}
	We conduct experiments in multiple scenarios to demonstrate the advantages of the proposed method.
	For image projection, we consider four recent encoder-decoder methods, ALAE  \cite{alae}, StarGANV2 \cite{starganv2}, StyleMapGAN \cite{stylemapgan}, PSP \cite{psp}, and one state-of-the-art GAN method, StyleGAN2 \cite{stylegan2} with 1000-steps optimization\footnote{\url{https://github.com/rosinality/stylegan2-pytorch}}.
	For attribute manipulation, we compare the proposed method with typical methods in two main streams: StyleGAN2 \cite{stylegan2}, AlAE \cite{alae}, InterFaceGAN \cite{interfacegan} (noted as IFGAN) of latent space manipulation methods and AttGAN \cite{attgan}, StarGANV2 \cite{starganv2} of image-to-image translation methods. For component transfer, we mainly compare with StyleMapGAN \cite{stylemapgan}, which is capable of editing individual components. We also perform qualitative comparisons with MGPE \cite{mask-guided} and SEAN \cite{sean} in \ref{sec:sup-comtrsf}, which are segmentation-based methods. Note that we mainly compare against recent methods with released codes. The detailed implementations of the baselines are described in \ref{sec:sup-implementation}.
	
	\subsection{Evaluation metrics} \label{sec:exp-metrics}
	\paragraph{Reconstruction}
	To evaluate the projection quality, we estimate four metrics, including pixel-level and perceptual-level differences between target images and reconstructed images: 1) mean square error
	(MSE), 2) learned perceptual image path similarity (LPIPS) \cite{lpips}, 3) peak signal-to-noise ratio
	(PSNR), and 4) structural similarity index (SSIM) \cite{wang2004image}.  We also leverage Fr{\'e}chet Inception Distance (FID) \cite{heusel2017gans} to evaluate the fidelity of the reconstructed images. 
	
	\paragraph{Attribute manipulation} Attribute manipulation performance can be evaluated from four aspects: attribute editing accuracy, irrelevance preservation, identity preservation, and image quality preservation.  
	\textbf{Attribute editing accuracy} is used to evaluate whether a specified attribute correctly appears on the edited image. We adopt the attribute classifiers\footnote{\url{https://github.com/NVlabs/stylegan/}} used in StyleGAN \cite{stylegan} to calculate it. Irrelevance preservation is measured by \textbf{MSE$_{\text{irr}}$}, which is calculated as the $L_2$ difference of the attribute-irrelevant regions between the edited image and the original projected image. MSE$_{\text{irr}}$ evaluates whether the attribute irrelevant details are kept after the editing and can also measure the disentanglement of latent space manipulation methods. Following PA-GAN\cite{pa-gan}, we define an irrelevant region that should not be altered when editing a given attribute. (Refer to Figure \ref{fig:sup-exp-attr-region} in \ref{sec:sup-exp-metrics} for the irrelevant region definitions for the considered attributes.) To evaluate identity preservation, we propose \textbf{Arc-dis}, which adopts a recent popular face recognition model, ArcFace \cite{arcface}, to calculate the distance between the edited face and the reconstructed one. A larger distance means that the two faces tend to have different identities. Arc-dis is a person-level distance, measuring the influence of local attribute editing on the face identity. To the best of our knowledge, there is still a lack of a comprehensive quantitative metric to measure whether the edited image maintains the same quality and fidelity as the original image. To solve this problem, we propose a new measurement \textbf{image-fidelity-gap (IFG)} to estimate the changes in image quality and fidelity before and after editing. Specifically, a pre-trained discriminator can output a score to indicate whether the image is real or fake, and image-fidelity-gap is defined as the decline of the score before and after editing. Here, we use the well-trained discriminator in StyleGAN2 \cite{stylegan2}.
	\paragraph{Component transfer}
	Similar to attribute manipulation, we use irrelevance preservation metrics (\textbf{MSE$_{\text{irr}}$}) and image quality preservation metrics (\textbf{image-fidelity-gap (IFG)}) to quantify the performance of component transfer. MSE$_{\text{irr}}$  is similar to the definition of MSE$_{\text{src}}$ in StyleMapGAN \cite{stylemapgan}. Note that we do not use  MSE$_{\text{ref}}$ defined  in StyleMapGAN \cite {stylemapgan} because the pixel-level difference between the reference component and the transferred results may run counter to the natural component transfer. Taking Figure \ref{fig:exp-com-stylemapgan} (second row, column 2-4) as an example, when the component in the target image and reference image have large posture or position differences, very unnatural fusion occurs in the results of StyleMapGAN. In this scenario, it still achieves a high  MSE$_{\text{ref}}$ while the results are not satisfactory.
	
	\subsection{Real Image Projection} \label{sec:exp-reconstruction}
	\begin{table}[t]
		\centering
		\begin{tabular}{lccccr}
			\toprule
			Methods                        & MSE(e-2) $\downarrow$ & LPIPS $\downarrow$ & PSNR$\uparrow$ & SSIM $\uparrow$ & FID$\downarrow$ \\ \toprule
			StyleGAN2 \cite{stylegan2}     &         7.873         &       0.218        &     17.17      &      0.584      &           17.41 \\
			ALAE \cite{alae}               &         10.65         &       0.470        &     15.78      &      0.458      &           26.86 \\ %
			StarGANV2 \cite{starganv2}     &         9.862         &       0.431        &     15.97      &      0.507      &           24.33 \\ %
			StyleMapGAN \cite{stylemapgan} &         2.331         &       0.237        &     22.14      &      0.662      &           4.167 \\ \hline
			IA-FaceS                       &         2.315         &       0.224        &     22.34      &      0.642      &           3.131 \\
			IA-FaceS+CAM                   &    \textbf{ 1.978}    &  \textbf{ 0.207}   & \textbf{22.98} & \textbf{0.666}  &  \textbf{2.976} \\ \bottomrule
		\end{tabular}
		\caption{Comparisons of image projection on 3,000 CelebA-HQ \cite{pg-gan} images at $256\times 256$. FID is calculated on 10,000 images.}
		\label{tab:exp-recon}                                                                                                           
	\end{table}
	\begin{figure}[t]
		\centering
		\includegraphics[width=\columnwidth]{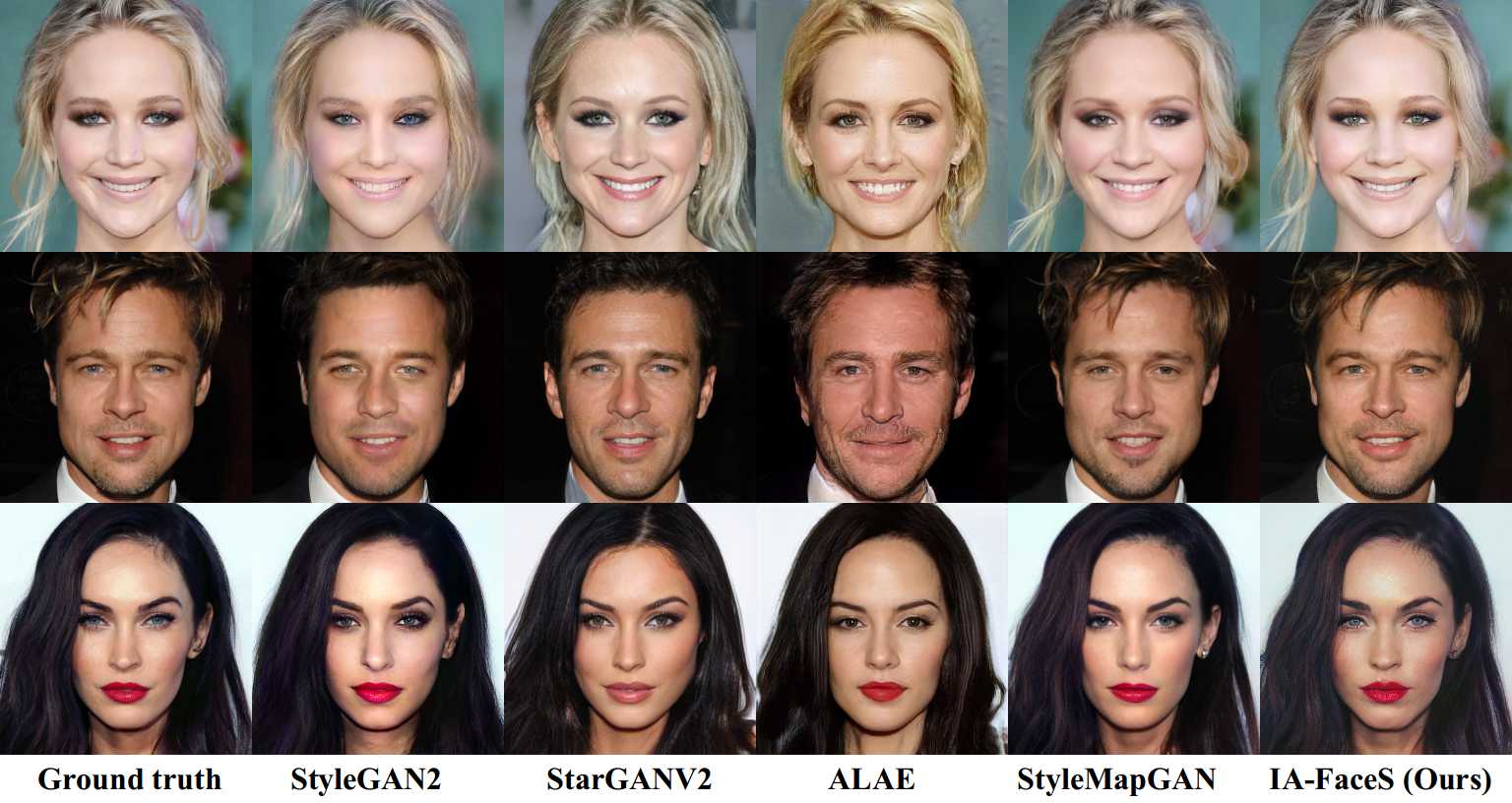}
		\caption{Reconstructions of real-world images with models trained on CelebA-HQ \cite{pg-gan} at $256\times256$.}
		\label{fig:exp-recon}
	\end{figure}
	\begin{table}[!ht]
		\centering
		\begin{tabular}{lcccr}
			\toprule
			Methods                      & MSE(e-2) $\downarrow$ & LPIPS $\downarrow$ & PSNR$\uparrow$ & SSIM $\uparrow$ \\ \toprule
			StyleGAN2 \cite{stylegan2}   &         10.42         &       0.471        &     16.40      &           0.619 \\
			ALAE \cite{alae}             &         18.65         &       0.603        &     13.52      &           0.480 \\
			StyleMapGAN \cite{stylemapgan} &         3.289         &       0.413        &     20.98      &           0.638 \\
			PSP \cite{psp}       &         4.426         &       0.486        &     19.33      &           0.647 \\ \hline
			IA-FaceS                     &        \textbf{ 2.397}         &   \textbf{0.384}   & \textbf{22.66} &          \textbf{ 0.659} \\
			IA-FaceS$^*$+CAM                     &        2.671         &   0.420   & 21.90 &          0.655 \\ \bottomrule
		\end{tabular}
		\caption{Comparisons of image projection on 500 FFHQ \cite{stylegan} images at $1024\times 1024$. Due to high GPU-memory, we remove the perceptual-level reconstruction loss when training IA-FaceS+CAM on FFHQ \cite{stylegan} dataset, which is denoted as IA-FaceS$^*$+CAM.}
		\label{tab:exp-recon-ffhq}
	\end{table}

	Efficient and accurate projection of the input image is a basic requirement for editing real-world images. According to the quantitative results in Table \ref{tab:exp-recon}, IA-FaceS+CAM consistently outperforms other benchmarks over most of the metrics, while IA-FaceS ranks second. We further find that StyleMapGAN \cite{stylemapgan} has a performance comparable to the proposed method. This is expected because StyleMapGAN also utilizes a  tensor with explicit spatial dimensions for reconstruction, which preserves more information. It can be seen from Figure \ref{fig:exp-recon} that ALAE \cite{alae} and StarGANV2 \cite{starganv2} suffer from obvious mismatches between reconstructed images and the target one since the low-dimensional bottleneck layer constrains the reconstruction capacities of the networks. In contrast, StyleMapGAN and the proposed models can produce a more realistic and consistent reconstruction of the real images with high-dimensional embeddings. Although StyleGAN2 has a lower LPIPS than IA-FaceS as shown in Table \ref{tab:exp-recon}, it can hardly recover the details in the original images in Figure \ref{fig:exp-recon} and takes a long time, about $1000\times$ to other benchmarks. 
	
\begin{figure}[t]
		\centering
		\includegraphics[width=\textwidth]{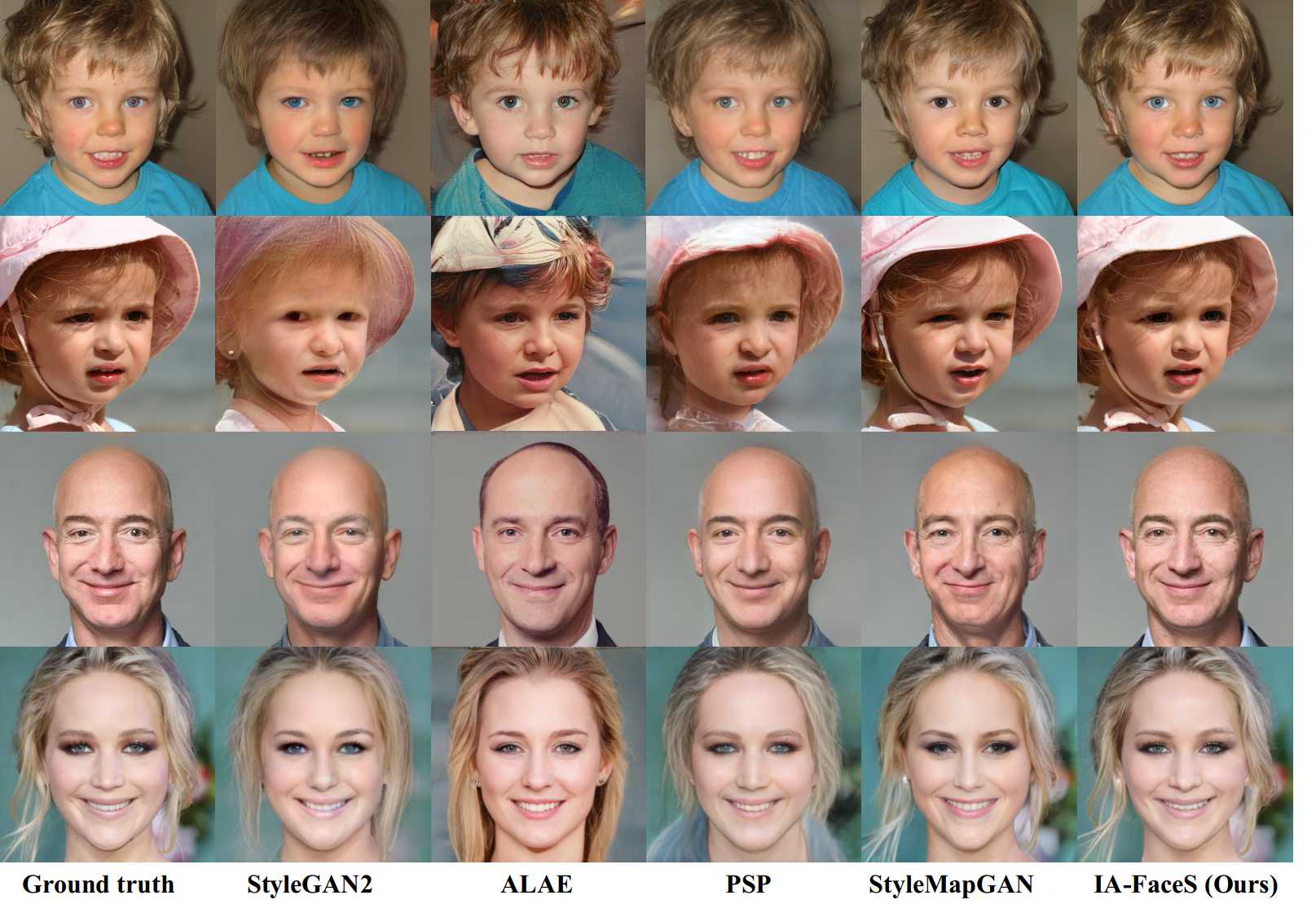}
		\caption{Comparisons of real image reconstructions at $1024\times1024$ resolution.}
		\label{fig:exp-recon-1024}
	\end{figure}

	Table \ref{tab:exp-recon-ffhq} summarizes the high resolution ($1024\times1024$) reconstruction performances, in which the proposed method maintains the best performance at all metrics. The qualitative comparison in Figure \ref{fig:exp-recon-1024} demonstrates the effectiveness of the proposed method in restoring high-resolution image details. Both qualitative and quantitative results demonstrate that IA-FaceS can preserve the original images in terms of pixel accuracy, perceptual similarity, and image quality. For more examples of reconstructions on real-world images, pleases refer to \ref{sec:sup-exp-recon}.
	
	\subsection{Face Attribute Manipulation} \label{sec:exp-attr}
	\begin{figure}[t]
		\centering
		\includegraphics[width=\columnwidth]{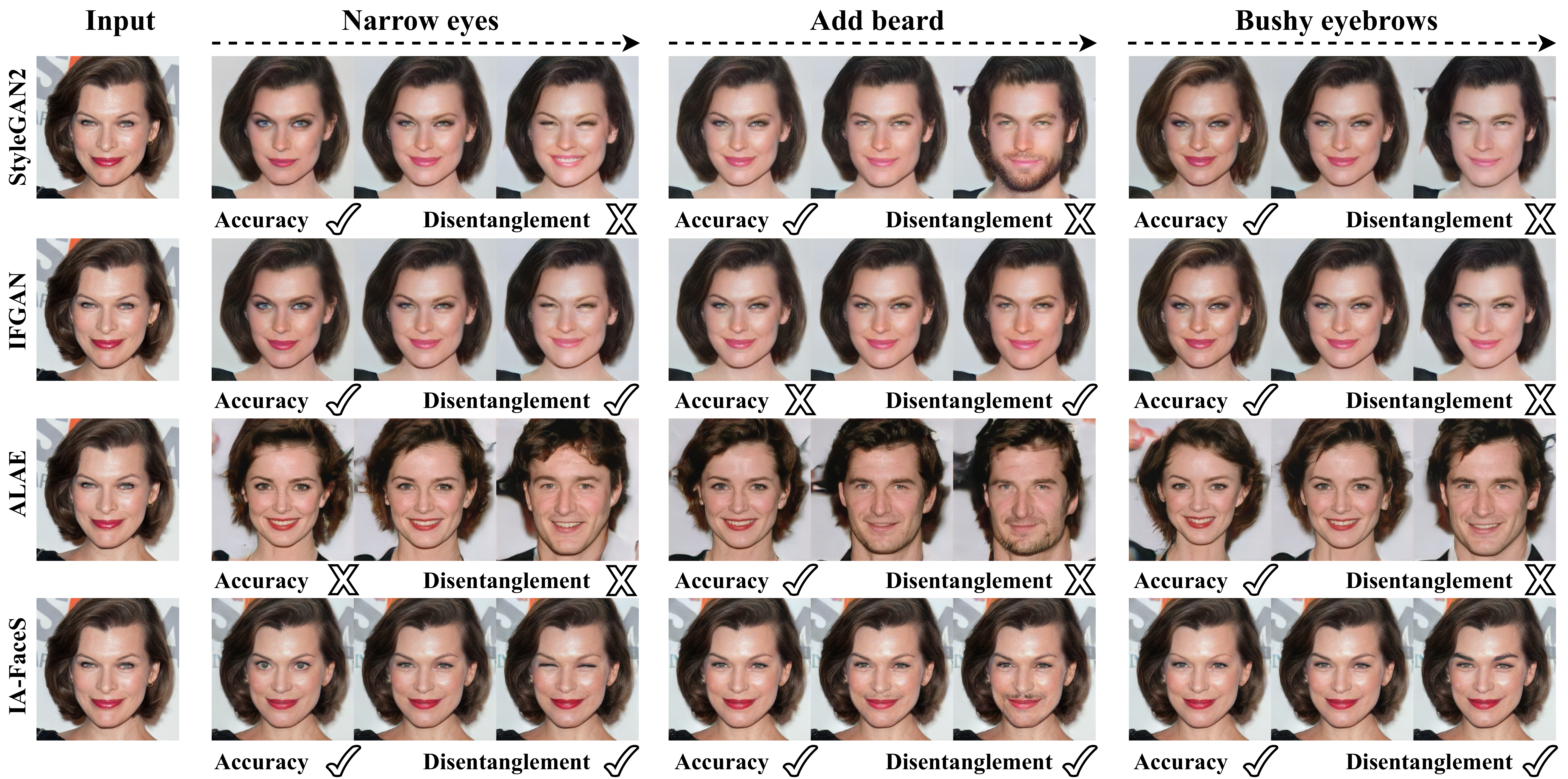}
		\caption{Illustration of attribute manipulation by the proposed method and other state-of-the-art approaches. All test images are wild images used in \cite{alae}. For each attribute, when sliding the code towards attribute direction (from left to right), we use  ``Accuracy"  to mark whether the attribute correctly appears ($\checkmark$) in the synthesized face or not ({\tiny\XSolid}). Meanwhile, we use ``Disentanglement" to evaluate whether the manipulation has side effect ({\tiny\XSolid}) (e.g., changes in gender or hair) or not ($\checkmark$). IFGAN represents the conditional manipulation of InterFaceGAN \cite{interfacegan} with StyleGAN2 as the backbone (\emph{narrow eyes} $|$ \emph{mouth open}, \emph{add beard} $|$ \emph{gender}, \emph{bushy eyebrows} $|$ \emph{gender and hair color}, ``$|$" denotes conditioned on, e.g., \emph{add beard} $|$ \emph{gender} means that add beards by preserving gender). Better viewed on-line in color and zoomed in for details.}
		\label{fig:exp-entanglement}
	\end{figure}
	
	\begin{figure}[!ht]
		\centering
		\includegraphics[width=\columnwidth]{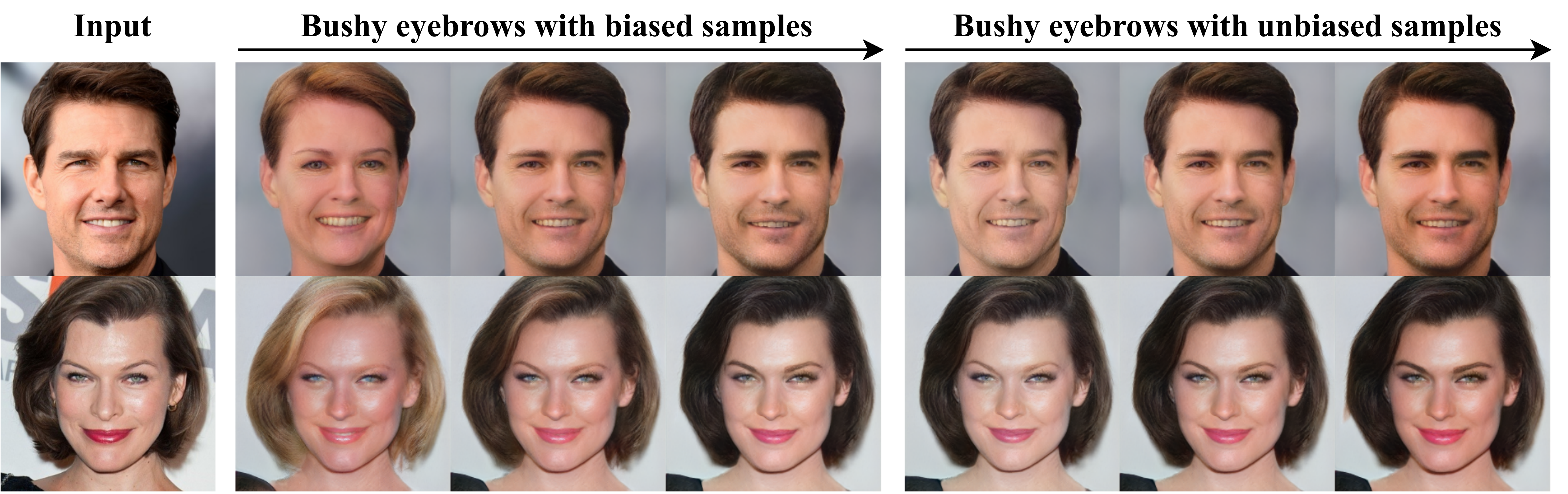}
		\caption{StyleGAN2 \cite{stylegan2} interpolation results along the direction calculated with the biased samples (left) and the unbiased virtual samples (right).}
		\label{fig:balance-sample}
	\end{figure}

	\begin{figure}[ht]
		\centering
		\includegraphics[width=\textwidth]{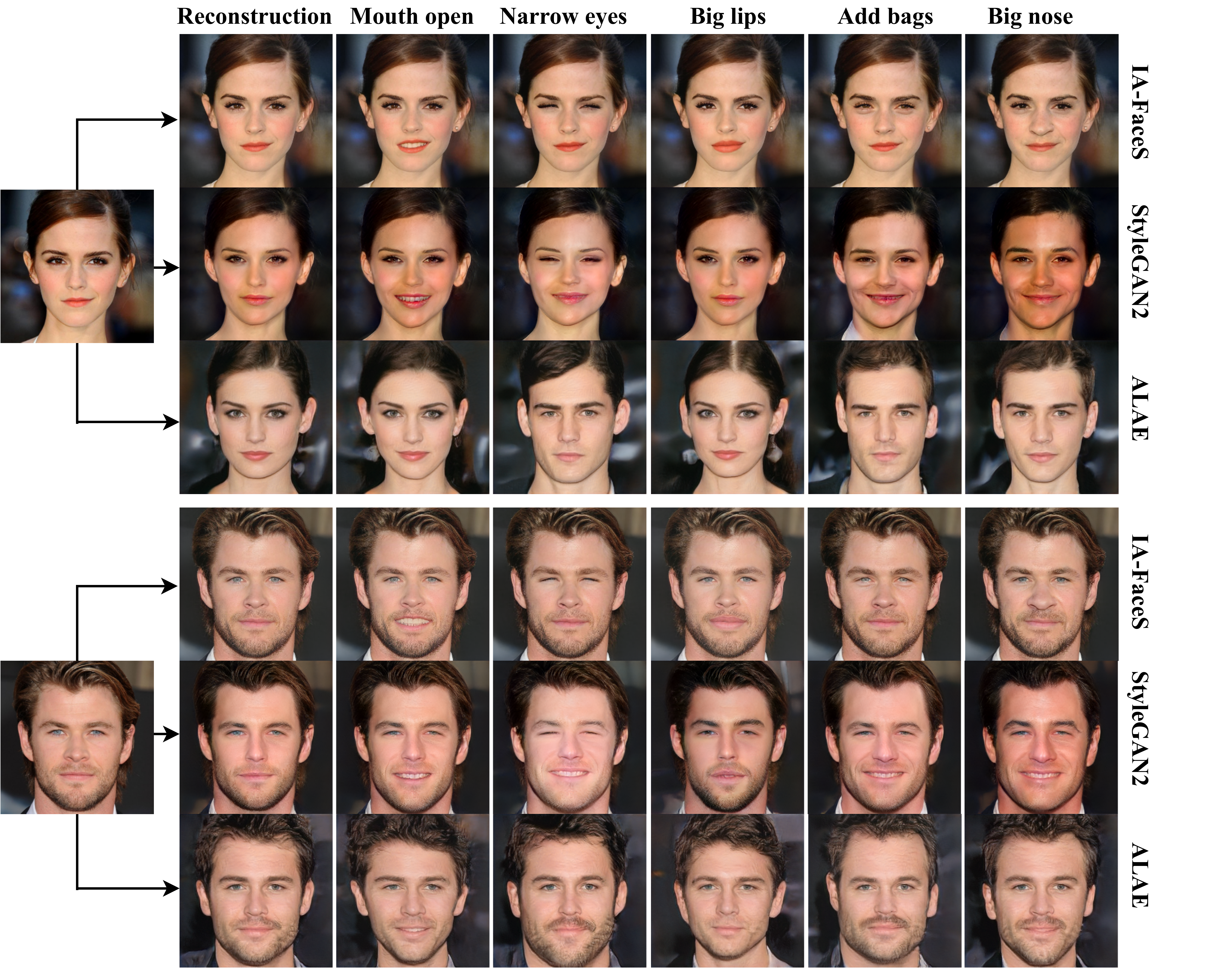}
		\caption{Qualitative results from local attribute editing by  ALAE \cite{alae} (bottom), StyleGAN2 \cite{stylegan2} (middle) and IA-FaceS (top).}
		\label{fig:exp-com-local-attr}
	\end{figure}
	
	\begin{table}[t]
		\centering
		\begin{tabular}{l|ccccccr}
			\toprule
			\multirow{2}{*}{Methods}   &    Bushy     &    Mouth     &    Arched    &     Add      &     Add      & \multirow{2}{*}{Old} & \multirow{2}{*}{Male} \\
			&     E.b.     &     Open     &     E.b.     &     Bags     &    Beard     &                      &                       \\ \midrule
			StyleGAN2 \cite{stylegan2} &     834      & \textbf{984} &     721      &     487      &     657      &     \textbf{971}     &                   887 \\
			IFGAN \cite{interfacegan}  &     801      &     968      &     578      &     487      &     333      &     954     &                   842 \\
			ALAE \cite{alae}           &     303      &     508      &     815      &     312      &     347      &         195          &                   529 \\
			AttGAN \cite{attgan} &     810      &     921      &      -       &      -       &      -       &          -           &                   823 \\ 
			StarGANV2 \cite{starganv2} &      -       &      -       &      -       &      -       &      -       &          -           &                   894 \\ \hline
			IA-FaceS                   & \textbf{979} & 977 & \textbf{845} &     560      & \textbf{925} &         755          &                   \textbf{957}    \\
			IA-FaceS+CAM               &     958      &     976      &     814      & \textbf{628} &     699      &     800     &                   940 \\ \bottomrule
		\end{tabular}
		\caption{The attribute editing accuracy of the proposed methods and other competitors. For a specific attribute, we count the success cases of attribute manipulation on 1,000 samples without the corresponding attribute (e.g., the \emph{mouth open} accuracy is calculated on 1,000 faces with closed mouth). IFGAN represents the conditional manipulation of InterFaceGAN \cite{interfacegan} with StyleGAN2 as the backbone. Note that we use ``$|$" to denote ``conditioned on" (e.g., \emph{add beard} $|$ \emph{gender} means that add beards by preserving gender), and we have \{\emph{bushy eyebrows, arched eyebrows}\} $|$ \{\emph{gender, hair color}\},  \emph{mouth open} $|$ \emph{narrow eyes}, \{\emph{add beard, old}\} $|$ \emph{gender}, \emph{male} $|$ \emph{age}. Note that AttGAN and StarGANV2 are limited to their predefined attributes, thus we only compare the performance on the common attributes. E.b. is short for eyebrows.}
		\label{tab:exp-attr-accuracy}
	\end{table}
	
	\begin{table}[t]
		{\tabcolsep0.055in   
			\begin{tabular}{l|c|cccccc}
				\toprule
				\multirow{2}{*}{Methods} & \multirow{2}{*}{Metrics} &     Narrow      &      Mouth      &      Bushy      &     Arched      &       Big       &       Add       \\ 
				\multicolumn{1}{l|}{}    &                          &      Eyes       &      Open       &    E.b.     &    E.b.     &      Nose       &      bags       \\ \midrule
				\multirow{3}{*}{SG2 \cite{stylegan2}}     &     MSE$_\text{irr}$ $\downarrow$      &      2.275      &      1.679      &      6.500      &      4.384      &      4.663      &      2.186      \\
				&           Arc-dis $\downarrow$            &      0.296      &      0.248      &      0.400      &      0.321      &      0.310      &      0.272      \\
				&           IFG  $\uparrow$           &     -1.028      &     -0.796      &     -2.151      & \textbf{-2.938} &     -5.000      &     -0.620      \\ \midrule
				\multirow{3}{*}{AlAE \cite{alae}}    &     MSE$_\text{irr}$ $\downarrow$      &      6.770      &      7.120      &      9.426      &      7.235      &      7.253      &      6.264      \\
				&           Arc-dis  $\downarrow$           &      0.372      &      0.424      &      0.489      &      0.557      &      0.361      &      0.392      \\
				&           IFG  $\uparrow$           & \textbf{+10.62} &     -8.517      & \textbf{+10.65} &     -10.30      &     -1.704      & \textbf{+7.381} \\ \midrule
				\multirow{3}{*}{IA-FaceS}     &     MSE$_\text{irr}$ $\downarrow$      & \textbf{0.885}  & \textbf{0.977}  & \textbf{0.927}  & \textbf{0.886}  & \textbf{1.018}  & \textbf{0.937}  \\
				&           Arc-dis  $\downarrow$            & \textbf{0.194}  & \textbf{0.194}  & \textbf{0.349}  & \textbf{0.307}  & \textbf{0.245}  & \textbf{0.184}  \\
				&           IFG $\uparrow$            &     +0.344      & \textbf{-0.658} &     -0.383      &     -4.966      & \textbf{+0.826} &     -0.753      \\ \bottomrule
		\end{tabular}}
		\caption{Quantitative evaluation on local facial attribute manipulation. SG2 is short for StyleGAN2 \cite{stylegan2}. The metrics are described in Section \ref{sec:exp-metrics}.}
		\label{tab:exp-attr-quality}
	\end{table}

	One of the major advantages of IA-FaceS is preserving the attribute-irrelevant regions, which is a desirable property for face attribute manipulation. For StyleGAN2 \cite{stylegan2}, AlAE \cite{alae}, and the proposed methods, we calculate the attribute direction following the ways in \cite{interfacegan} (formulated in Eq. (\ref{eq:syn-attmeth-b}), visualized in Figure \ref{fig:me-synthesis-attr}) and interpolate the latent code along the attribute direction to implement attribute control. From Figure \ref{fig:exp-entanglement}, we can conclude that: 
	(i) StyleGAN2 \cite{stylegan2} and ALAE \cite{alae} often produce entangled attribute changes. For instance, when adding the beard or bushy eyebrows to the woman, the gender is also changed. 
	(ii) The proposed method can separate the manipulations of attribute-relevant regions from the rest regions, allowing to manipulate attributes without side effects such as gender change. 
	(iii) InterFaceGAN (IFGAN) can solve the attribute entanglement to a certain extent, but it fails to add a beard and struggles to change the shape of eyebrows when keeping the gender unchanged. 
	
	As suggested in Table \ref{tab:intro-sample-bias}, many attributes have intrinsic correlations in the dataset, which poses challenges for disentangled attribute manipulation. To overcome this problem, InterFaceGAN \cite{interfacegan} proposed to rectify the target attribute direction by subtracting its projection on the conditional attribute direction, and this helped to get disentanglement. However, if the directions of target and conditional attributes are highly correlated, the rectified direction will affect the attribute correctness. For example, in the second row of Figure \ref{fig:exp-entanglement}, InterFaceGAN fails to add beards to the woman with the rectified direction, indicating that the rectified direction has less beard information than the original ones. For the attribute entanglement caused by dataset bias, we propose a simple but effective method to help the existing latent space manipulation models to achieve attribute disentanglement. Utilizing the merit that the proposed model can generate high-quality virtual faces, we can balance the percentage of men and women with arched and bushy eyebrows by attribute manipulation. Then, we use the unbiased virtual samples to calculate the attribute direction for StyleGAN2. Figure \ref{fig:balance-sample} demonstrates that the entanglement between eyebrows and gender is almost eliminated using this method. 
	
	\begin{figure}[!htbp]{	
			\subfigure[The Accuracy-IFG (top) and Accuracy-$\text{MSE}_{\text{irr}}$ (bottom) curves of each method for attribute \emph{Beard} (left) and \emph{Bushy eyebrows} (right). The point highlighted with a magenta circle denotes the point where the accuracy no longer increases or increases slowly with significant decline of IFG\&$\text{MSE}_{\text{irr}}$.]{
				\includegraphics[width=\columnwidth]{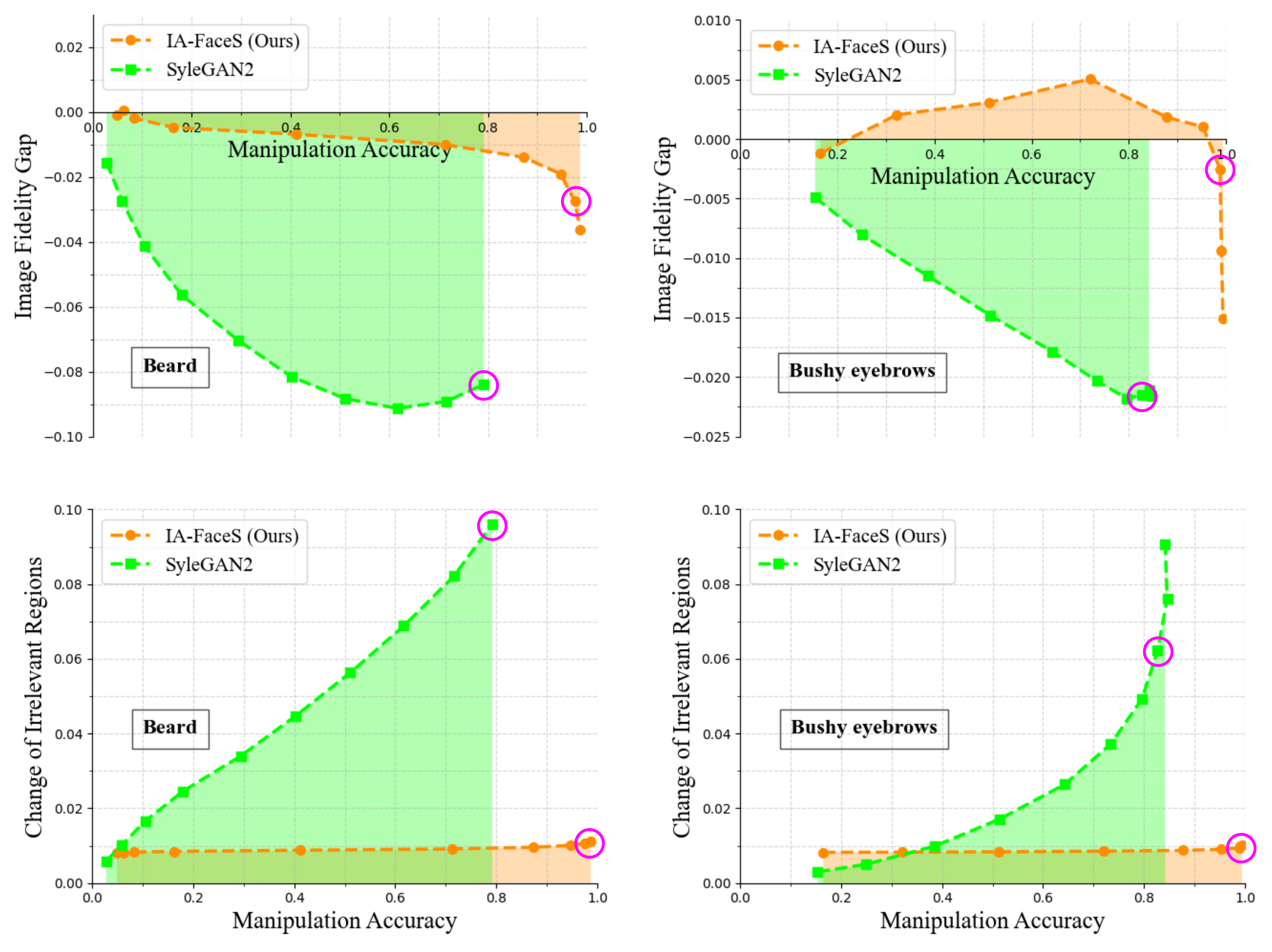}
				\label{fig:exp-roc1}
			}
			\subfigure[Face attribute manipulation results of different methods. Magenta boxes mark the results that use the highlighted hyper-parameters in (a).]{
				\includegraphics[width=\columnwidth]{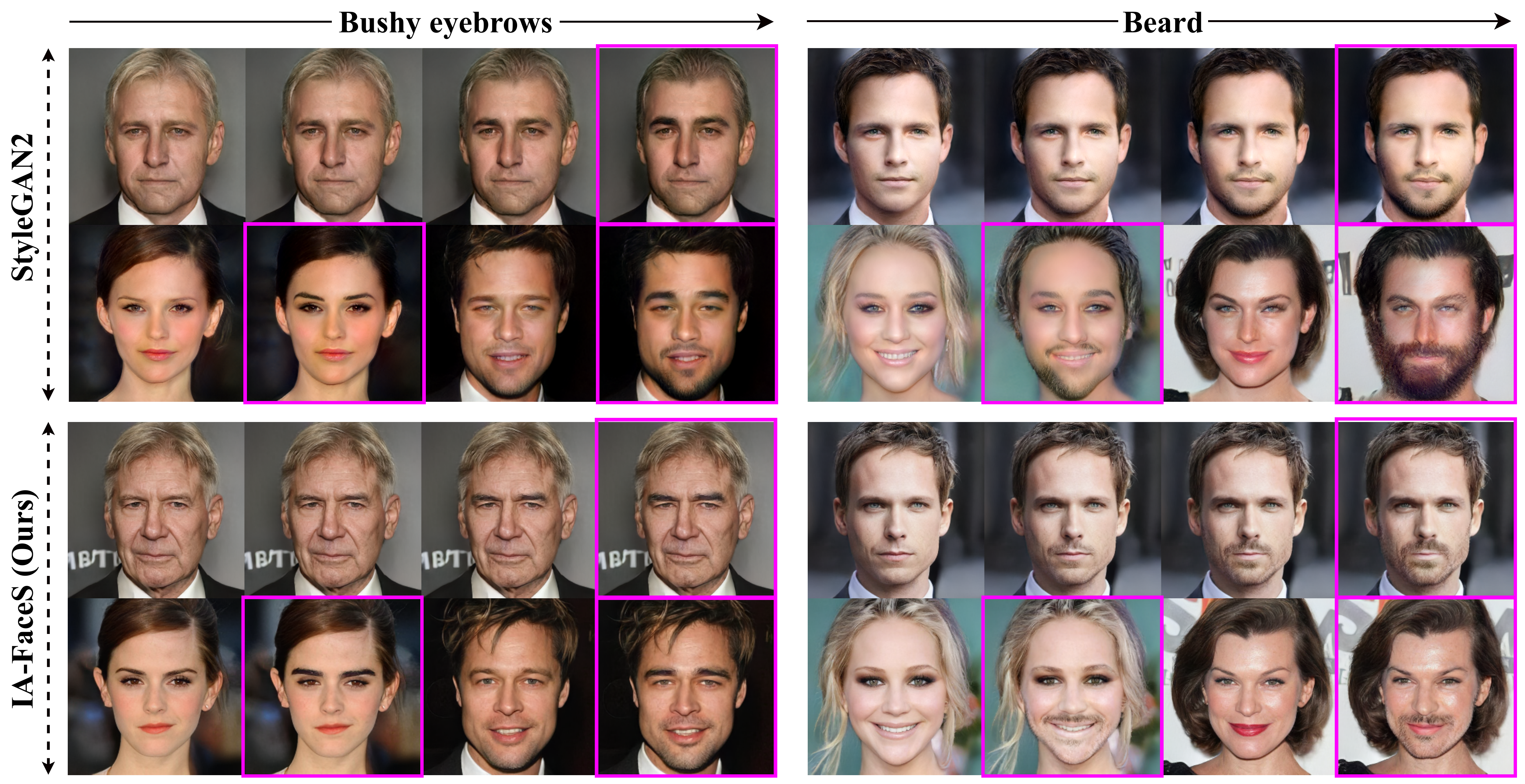}
				\label{fig:exp-roc2}
			}
			\label{fig:exp-roc}
			\caption{Visualization of the trade-off between manipulation accuracy with visual quality \& irrelevance preservation.}
		}
	\end{figure}
	
	\begin{figure}[!htbp]
		\begin{center}
			\includegraphics[width=\columnwidth]{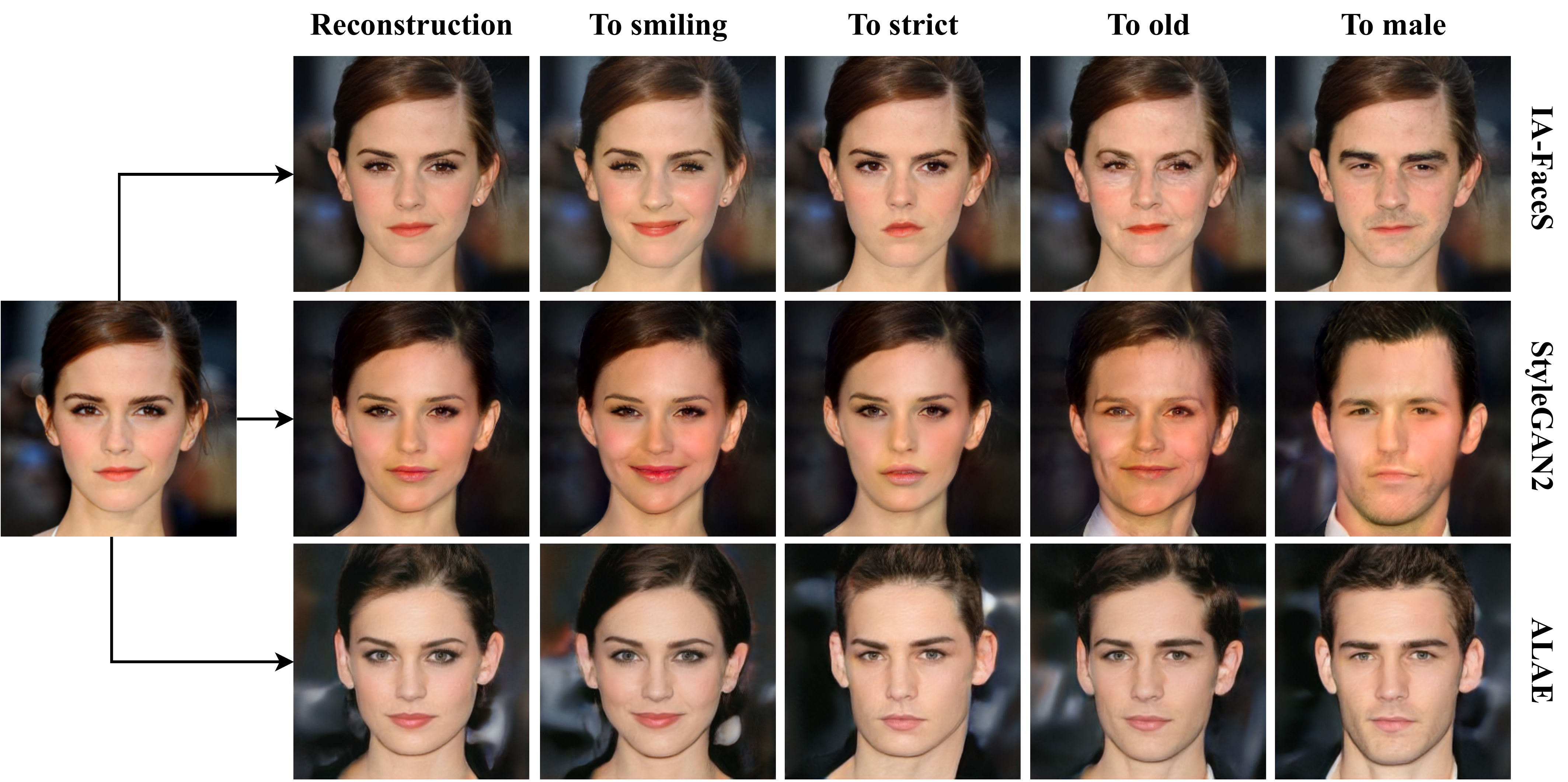}
			\caption{Results of global attribute manipulation by IA-FaceS (ours) and StyleGAN2 \cite{stylegan2}.}
			\label{fig:exp-global-attr}
		\end{center}
	\end{figure}
	
	Figure \ref{fig:exp-com-local-attr} provides more results from local attribute manipulations. Qualitatively, we see clear examples of what we can do, but other competitors cannot: (i) ALAE in general struggles to change attributes, especially for the first woman. (ii) While StyleGAN2 works better, it struggles with certain attributes (e.g., \emph{big lips} and \emph{big nose}) and often leads to serious changes in facial expressions and skin colors. (iii) The proposed method realizes spatial disentanglement on translations of different facial components in various semantic directions. This is because our method embeds the component separately and only changes the embeddings of components relevant to the target attribute. More qualitative results can be found in the supplementary material.
	
	We also compare with three competing image-to-image translation methods in Figure \ref{fig:sup-attr-comprare} of \ref{sec:sup-attr}. Overall, image-to-image translation methods are often restricted to predefined attributes before training and cannot continuously manipulate the attributes. In contrast, the latent space manipulation methods can freely and continuously edit arbitrary attributes, even if the model has been trained.  
	

	For quantitative comparison, we use the metrics described in Section \ref{sec:exp-metrics} and evaluate models on CelebA-HQ \cite{pg-gan} simultaneously considering the attribute editing accuracy (Table \ref{tab:exp-attr-accuracy}), the irrelevance \& identity preservation (Table \ref{tab:exp-attr-quality}), and visual quality (Table \ref{tab:exp-attr-quality}). From Table \ref{tab:exp-attr-accuracy}, we can see that IA-FaceS for most local facial attributes outperforms other competitors by a large margin in terms of editing accuracy. For the attributes \textit{Bushy/Arched Eyebrows}, \textit{Beard} and \textit{Bags under Eyes}, IA-FaceS achieves 10\%-25\% more accurate than competing methods. IA-FaceS+CAM has a similar performance to IA-FaceS except for the significant decline of accuracy on \emph{add beard}. One possible reason is that CAM is not skilled at editing attribute that cannot be constrained to a fixed component region. Although StyleGAN2 \cite{stylegan2} ranks first in controlling the mouth from close to open, it causes undesirable changes on attribute-irrelevant regions and affects image quality (see Table \ref{tab:exp-attr-quality}). Table \ref{tab:exp-attr-quality} shows that IA-FaceS achieves the best results for most attributes in visual quality (lower IFG), identity preservation (lower Arc-dis), and irrelevant region preservation (lower $\text{MSE}_{\text{irr}}$) while maintaining high attribute correctness.

	According to \cite{surrogate}, there is often a trade-off between attribute accuracy and disentanglement. For local attributes, the disentanglement can be measured by the preservation of attribute-irrelevant regions. To analyze the relationship between attribute accuracy with attribute disentanglement \& visual quality in different models, we gradually increase manipulation strength and calculate their attribute accuracy, attribute disentanglement, and visual quality measure at each point. From Figure \ref{fig:exp-roc1}, we observe a sacrifice of visual quality and attribute-irrelevance preservation in StyleGAN2 for higher accuracy.  Similar qualitative observations in Figure \ref{fig:exp-roc2} show that higher manipulation strength in StyleGAN2 will lead to more changes in attribute-irrelevant regions and declines in image fidelity. In contrast, with the increasing manipulation strength, IA-FaceS retains the high image fidelity score and shows excellent performance to keep attribute-irrelevant regions unchanged. Furthermore, Figure \ref{fig:exp-roc2} shows that the same manipulation strength of \emph{beard} exhibits inconsistent influence on different faces in StyleGAN2 \cite{stylegan2}, which is another weakness. Note that the AUCs of the Accuracy-IFG curve and Accuracy-$\text{MSE}_{\text{irr}}$ curve can be used as two strength-agnostic metrics \cite{surrogate}.
	
	\begin{figure}[!ht]
		\includegraphics[width=\columnwidth]{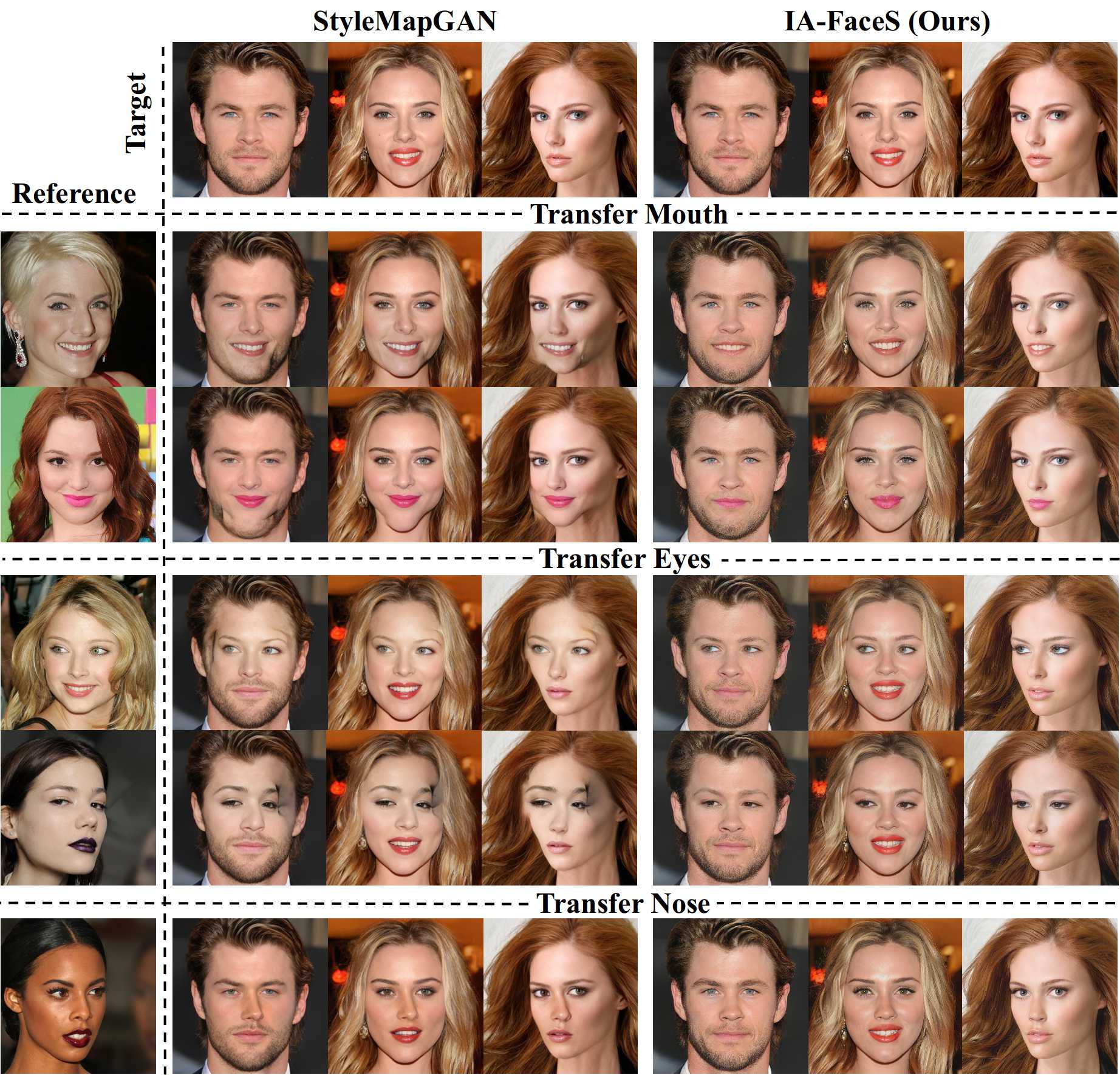}
		\caption{Examples of transferring specific components from the reference image to the target image by StyleMapGAN\cite{stylemapgan} and IA-FaceS. Both the shape and texture of the component can be well transferred to the target images by IA-FaceS. Even though the mouth is in an extreme posture in the reference image (row 2), the make-up and teeth can be successfully transferred by IA-FaceS and naturally fused with the target image. For eyes transfer, the expression of the eyes and the color of the pupil are both successfully inherit from the reference image. In contrast, StyleMapGAN often generates strange results without considering the pose of the target image (row 2, 5, and 6).}
		\label{fig:exp-com-stylemapgan}
	\end{figure}
	The proposed method can also deal with the manipulations of abstract global facial attributes, e.g., smile, age, and gender. Qualitative results in Figure \ref{fig:exp-global-attr} shows that gender is often entangled with expression and age in ALAE \cite{alae}. The entanglement between age and gender also appears in StyleGAN2 \cite{stylegan2}, while IA-FaceS can achieve disentangled attribute manipulation. However, the \emph{older} editing accuracy is lower than StyleGAN2 \cite{stylegan2}, which is a typical limitation of the proposed method. IA-FaceS is skilled in local attribute editing rather than global attribute manipulation, especially when it comes to hair color and face shape. 
	
	\subsection{Component Editing} \label{sec:exp-component}

	\begin{table}[t]
		\centering
		\begin{tabular}{l|ccc|ccr}
			\toprule
			\multirow{2}{*}{Methods}          & \multicolumn{3}{c|}{StylemapGAN \cite{stylemapgan}} &              \multicolumn{3}{c}{IA-FaceS}               \\ \cline{2-7}
			&  Eyes  &      Nose      &          Mouth           &      Eyes      &      Nose       &      Mouth      \\ \midrule
			MSE$_\text{irr}$ (e-2) $\downarrow$   & 1.677  & \textbf{1.067} &      \textbf{1.063}      & \textbf{1.014} &      1.142      &      1.082      \\
			FID $\downarrow$      & 10.28  &     9.435      &          9.560           & \textbf{7.804} & \textbf{7.550}  & \textbf{7.994}  \\
			IFG (e-2) $\uparrow$& -2.440 &     -1.786     &          -1.997          &    \textbf{ -1.906}     & \textbf{-1.141} & \textbf{-0.855} \\ \bottomrule
		\end{tabular}
		\caption{Quantitative comparisons with StyleMapGAN \cite{stylemapgan} on component transfer.}
		\label{tab:exp-com-stylemapgan}
	\end{table}

	Appears difficult for whole-face single-embedding methods \cite{alae, stylegan2} to perform component editing without affecting the rest of the face. Using separated embedding of the components, we can manipulate a specific component $c_i$ by manipulating its corresponding embedding $\boldsymbol{z}_i$, allowing us to flexibly edit the components, including the component transfer or employing unsupervised methods to find interpretable directions. 
	
\begin{figure}[!htb]
		\begin{center}
			\includegraphics[width=\columnwidth]{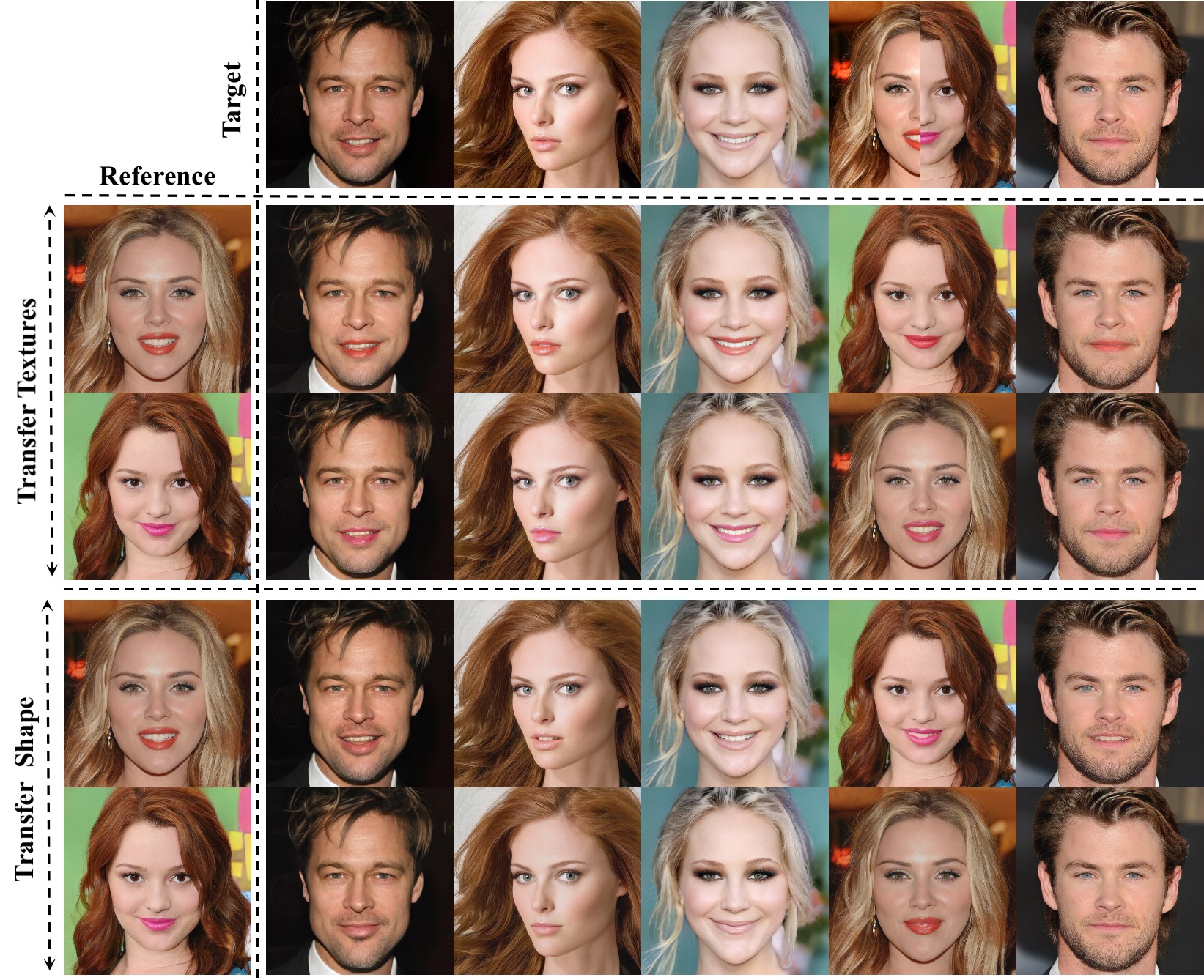}
			\caption{Mouth transfer on different levels. The target images in the first row and reference images in the first column are reconstructions of real images, while the rest are new faces generated by transferring mouth from the reference to target in different levels. (Top) Copying the coarse styles corresponding to resolutions ($8^2$ - $32^2$) brings high-level changes such as geometric characteristics. (Bottom) If we instead copy the styles of resolutions ($64^2$ - $256^2$), we inherit only the textures and colors (e.g., make-up on the mouth) of the component.}
			\label{fig:exp-multi-level}
		\end{center}
	\end{figure}

	We first investigate transferring the facial components from a reference image to the desired image, which corresponds in real life to the idea that we want to try another person's facial component on our face.  We can easily accomplish this by replacing the component embedding $\boldsymbol{z}_i$  of the target face with that of the reference face, as illustrated in Figure \ref{fig:me-synthesis-comtrsf}. According to the results shown in Figure \ref{fig:exp-com-stylemapgan}, IA-FaceS can naturally transfer a particular component in an example image to the target image without color distortions or pose inconsistency.  However, StyleMapGAN often generates strange results without considering the pose of the target image. This is because the proposed method implements component transfer in a semantic way, which can simultaneously locate facial components and consider the consistency with the target face, while StyleMapGAN directly copies the component regions of the reference \emph{stylemap} (size of $8 \times 8 \times 64$) to the target \emph{stylemap}, which is more flexible but lacks understanding of the whole face semantics. Furthermore, our method can freely choose to transfer the overall shape or textures according to the needs of users by replacing the style code in different levels, as suggested in Figure \ref{fig:exp-multi-level}. We can derive a similar observation from Table \ref{tab:exp-com-stylemapgan}  that the proposed method can synthesize more realistic images of high quality (higher IFG, lower FID) and perform better in maintaining the component-irrelevant regions (lower $\text{MSE}_{\text{irr}}$) than StyleMapGAN.
	
	\begin{figure}[!ht]
		\includegraphics[width=\columnwidth]{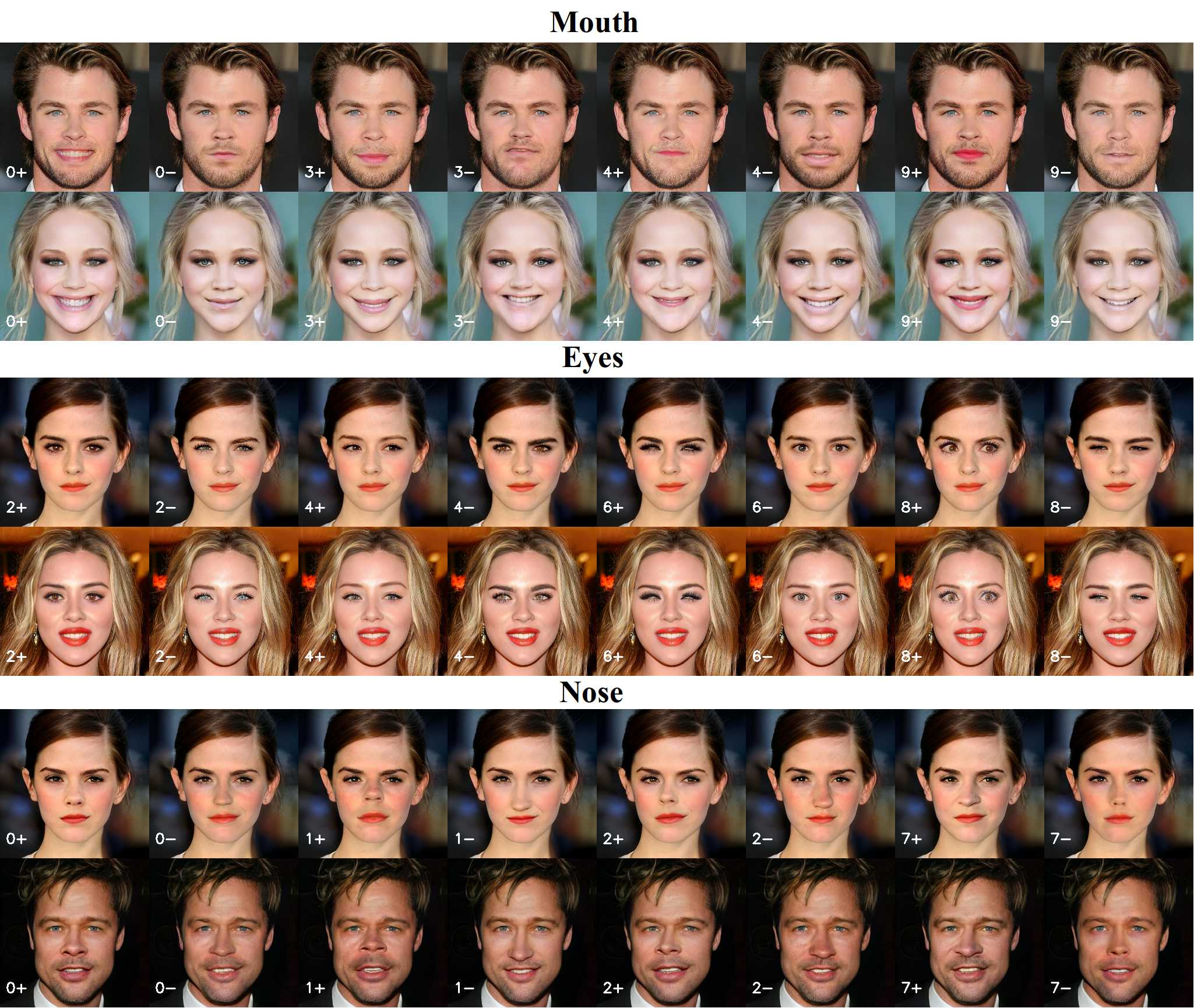}
		\caption{Examples of interpolations on directions found by PCA for mouth (top), eyes (middle), and nose (bottom). Each image shows an editing result of decreasing (-) or increasing (+) the value of a specific principal component (the original image is omitted). The index of principal component is overlaid in the bottom left corner of each image.}
		\label{fig:exp-pca}
	\end{figure}
	We also compare with MGPE \cite{mask-guided} and SEAN \cite{sean} in appendix, the results tell that MGPE \cite{mask-guided} and SEAN \cite{sean} can only transfer the texture and color of the facial component (see \ref{sec:sup-comtrsf}), while our method can freely choose to transfer the shape, textures or both. (See \ref{sec:sup-comtrsf} \& \ref{sec:sup-1024} for more qualitative results of component transfer on images at $256 \times 256$ resolution and $1024 \times 1024$ resolution.)
	
	We further apply unsupervised methods \cite{ganspace} to find interpretable directions in the latent space of each component. As usual, the unsupervised directions in StyleGAN's space tend to find global directions, such as age, illumination and pose, but it is difficult to find meaningful localized manipulations (such as changes in eyes, nose length, and mouth shape, etc.). Here, we can find many interesting fine-grained directions for component editing by applying PCA \cite{ganspace} to the latent space of each component. As illustrated in Figure \ref{fig:exp-pca}, the first few principal components control geometric configuration (0, 4 in mouth) or appearance (9 in mouth) for a specific component. For eyes, the interpolations on directions of principal components bring various changes in the expression of eyes (4, 6, 8) or the color of the pupil (0). For the nose, the length (1), pose (2), and size (7) of the nose can be freely edited by sliding the principal components. These directions can provide users with more fine-grained editing on components and enable the model to produce more facial expressions.

	\subsection{Single-eye Editing} \label{sec:exp-single-eye}
	\begin{figure}[!htb]
		\centering
		\includegraphics[width=\columnwidth]{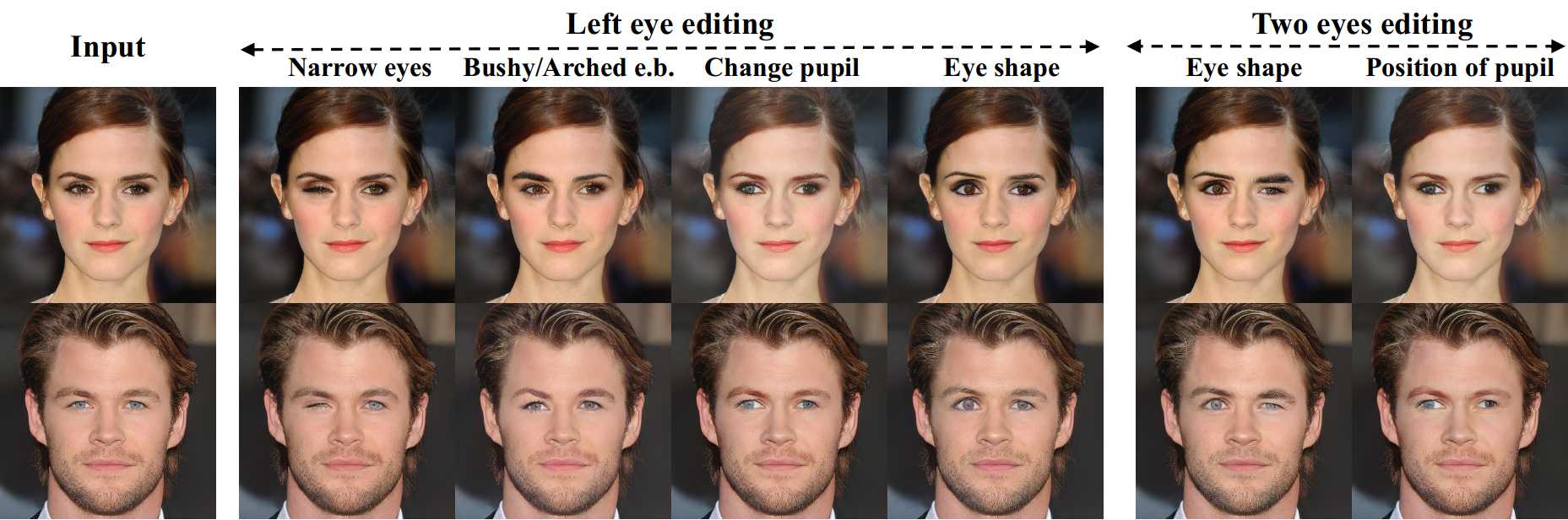}
		\caption{Examples of single-eye editing (left) and different controls over the two eyes (right). e.b. is short for eyebrows.}
		\label{fig:exp-single-eye}
	\end{figure}
	
	\begin{figure}[!htb]
		\centering
		\includegraphics[width=\columnwidth]{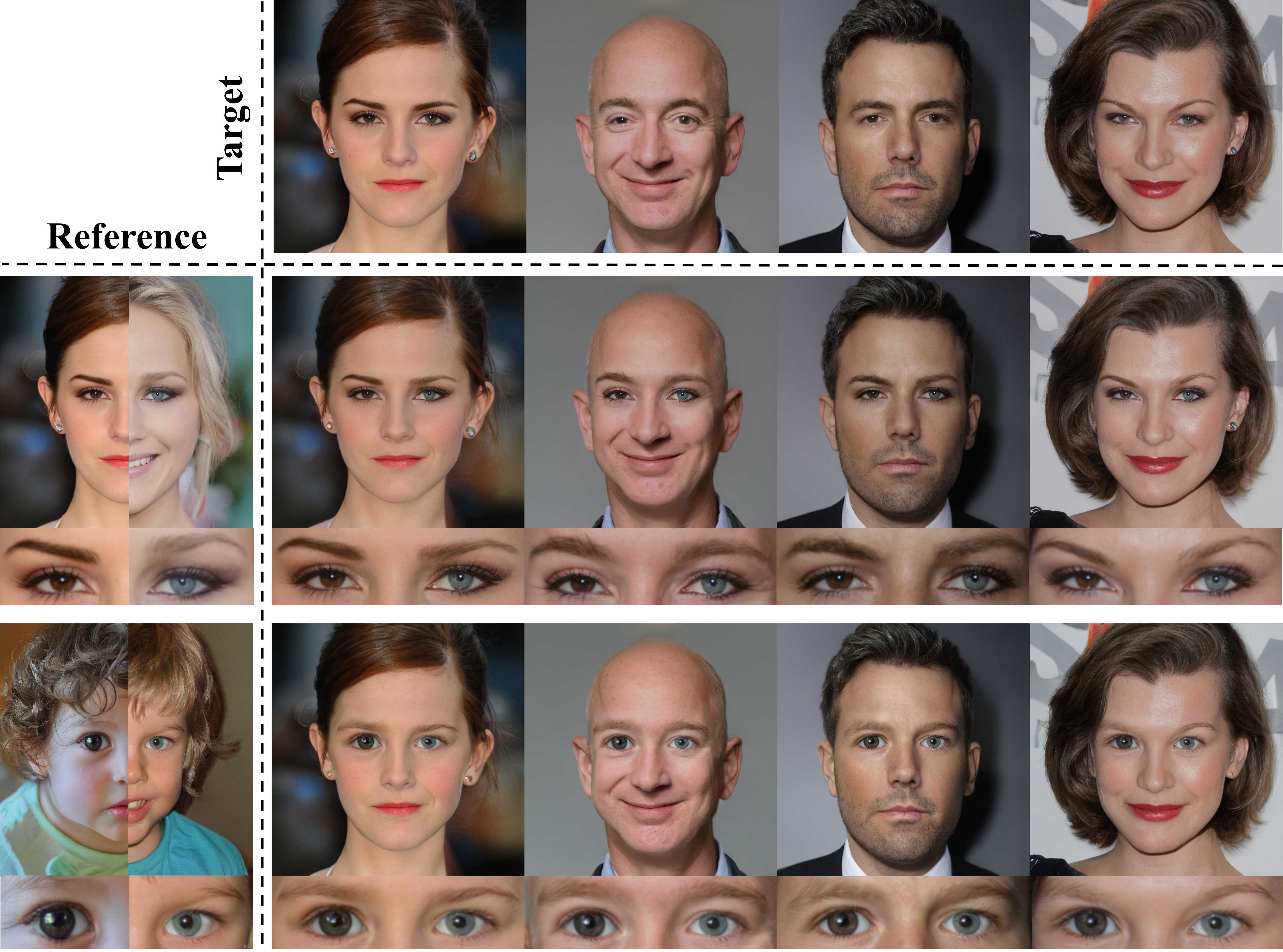}
		\caption{Eyes transfer by IA-FaceS$^*$+CAM on FFHQ. The target images in the first row and reference images in the first column are reconstructions of real images. (Note that the left and right halves of the reference image are cropped from two different faces, where the left face provides the left eye, and the right face provides the right eye.) The rest are new faces generated by transferring the left and right eyes from the left and right halves of the reference image to the target image, respectively. Note that the details of the eyes (such as the shape of the eyes and the color of the pupil) can be correctly transferred and both of the eyes are naturally fused with the target image. The regions of eyes are zoomed in below each image for a better view.}
		\label{fig:exp-ffhq-diffeye}
	\end{figure}
	Armed with component adaptive modulation, IA-FaceS+CAM can implement different control over the left and right eye. As shown in Figure \ref{fig:exp-single-eye}, we can edit the attribute of a single eye without affecting the other one and apply different controls over the two eyes, thus we can produce faces that do not exist in the dataset, e.g., faces with one eye open and the other eye closed. From Figure \ref{fig:exp-ffhq-diffeye}, IA-FaceS$^*$+CAM can generate high-quality faces of high resolution where left and right eyes are transferred from different sources. To the best of our knowledge, IA-FaceS+CAM is the first latent space manipulation method that can edit a single eye without any visual guidance (sketch, mask, landmarks, 3d instructions, etc.).
	
	\subsection{Challenging Cases}   \label{sec:exp-generalization}
\begin{figure}[ht]
		\subfigure[Attribute manipulation under extreme conditions.]{
			\centering
			\includegraphics[width=\columnwidth]{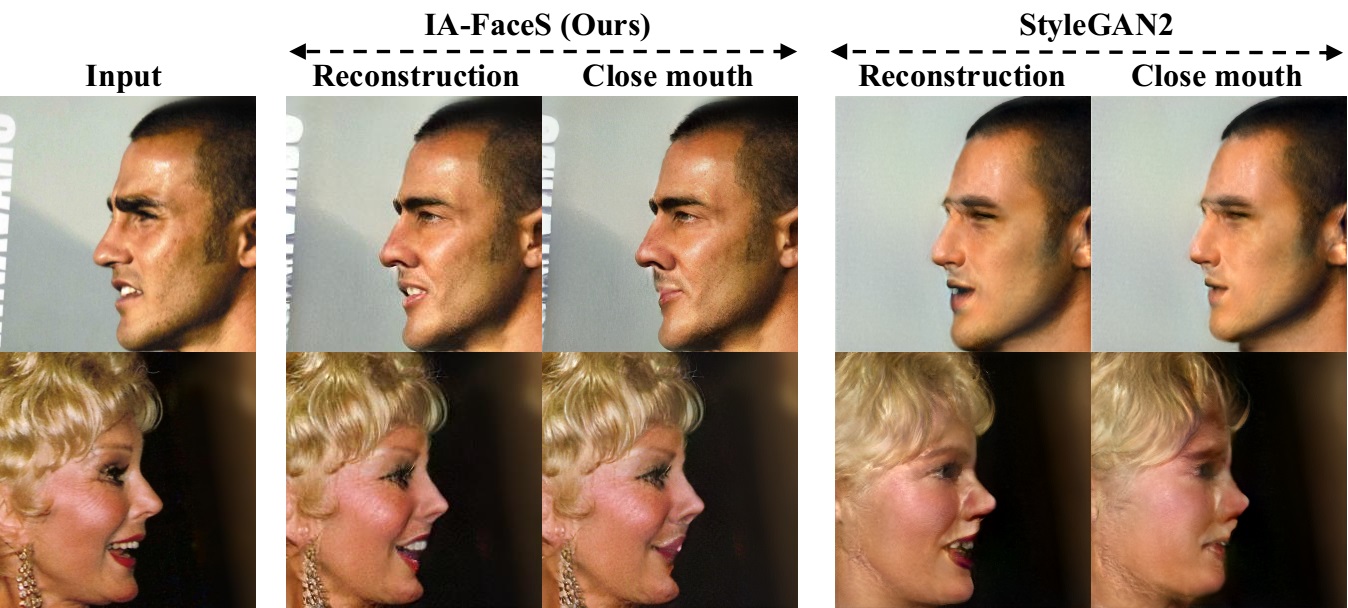}
		}	
		\subfigure[Component transfer under extreme conditions.]{
			\centering
			\includegraphics[width=\columnwidth]{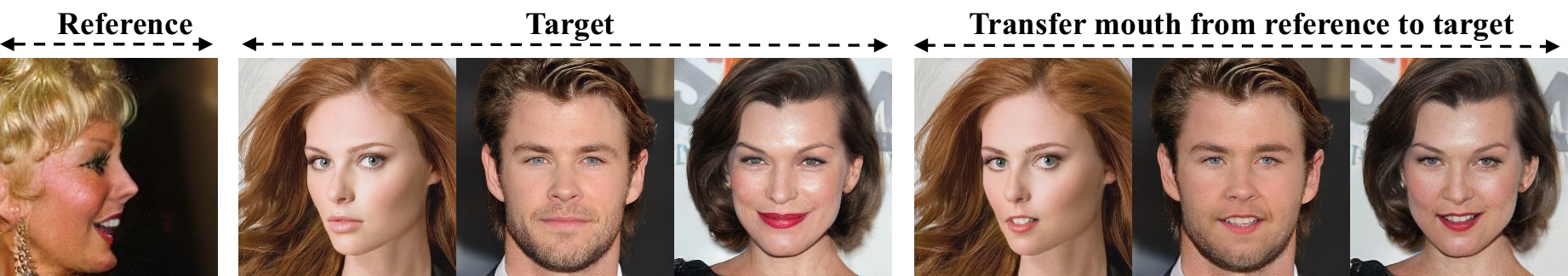}
		}
		\caption{The performance of IA-FaceS on challenging cases.}
		\label{fig:exp-extreme}
	\end{figure}

\begin{figure}[!htb]
		
		\subfigure[Attribute manipulation on cartoon faces. Recon is short for reconstruction.]{
			\centering
			\includegraphics[width=\columnwidth]{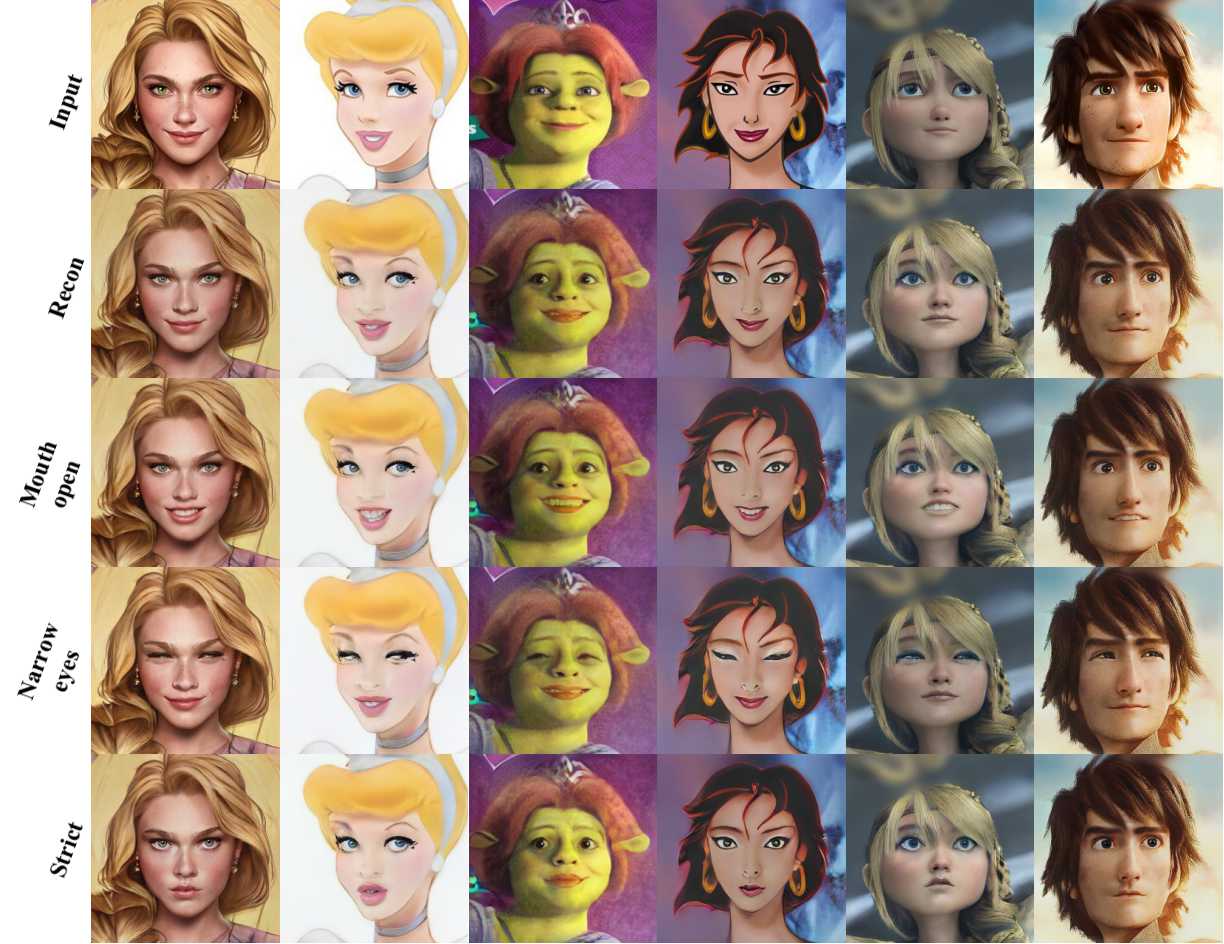}
			\label{fig:exp-attr-cartoon}
		}	
		\subfigure[Component editing on cartoon faces. Each image shows an editing result of decreasing (-) or increasing (+) the value of a specific principal component (the original image is omitted). The index of principal component is overlaid in the bottom left corner of each image.]{
			\centering
			\includegraphics[width=\columnwidth]{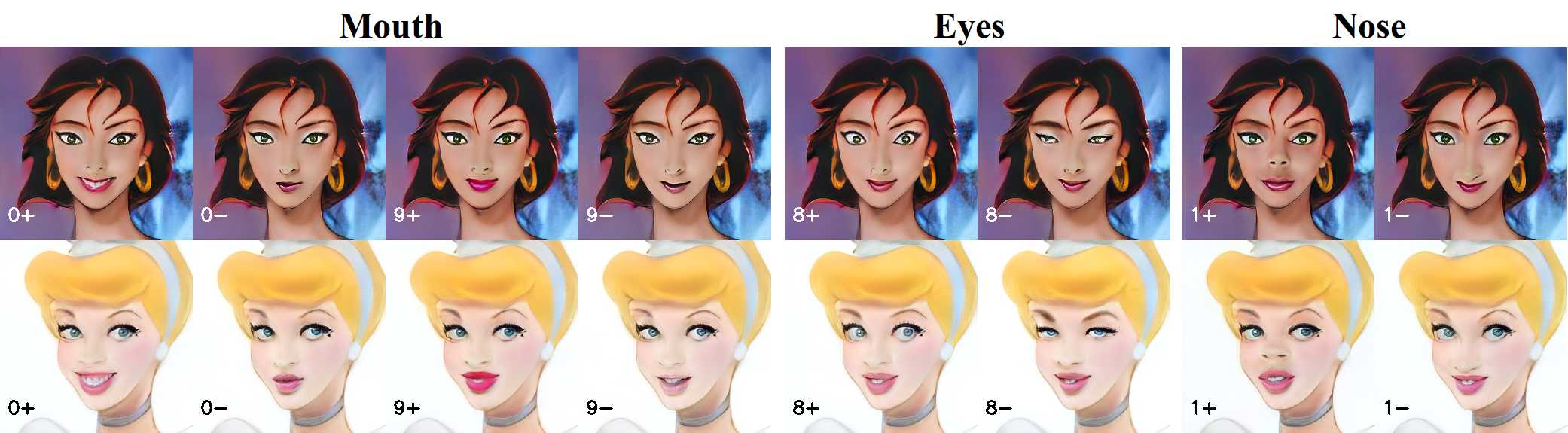}
			\label{fig:exp-pca-cartoon}
		}
		\caption{Examples of cartoon face editing by IA-FaceS trained on CelebA-HQ \cite{pg-gan} at $256\times256$. (a) Face attribute manipulation with the attribute directions calculated on CelebA-HQ. (b) The interpolations on directions found by PCA for mouth (left most), eyes (middle), and nose (right most). }	
		\label{fig:exp-cartoon}
	\end{figure}

	In this section, we demonstrate the strong capability of IA-FaceS by editing wild face images under extreme conditions and cartoon faces  \cite{toonify}.  As illustrated in Figure \ref{fig:exp-extreme}, even under extreme poses, IA-FaceS still has superior visual quality and editing performances. The results in Figure \ref{fig:exp-cartoon} show that IA-FaceS trained on CelebA-HQ \cite{pa-gan} can generalize to non-real faces far from the distribution of the training images, demonstrating the strong generalization ability of the proposed method.
	
	\subsection{Analysis of IA-FaceS} 	\label{sec:exp-ablation}
	In this section, we mainly investigate the importance of loss functions, data augmentation (trained with Bernoulli modified images, see Eq. (\ref{eq:mask})), the roles of face icon \& component embeddings, and the effect of the size of $\boldsymbol{S}$. Table \ref{tab:exp-summary} presents variants of IA-FaceS. 
	
	\begin{table}[t]
		\centering
				\begin{tabular}{lcccr}
					\toprule
					Model                &      Size of   $\boldsymbol{S}$      &        Augmentation        &           CAM           &        LPIPS \\ \toprule
					IA-FaceS             &         $8\times 8\times 512$       &         \Checkmark         &      \XSolidBrush       & \Checkmark \\
					IA-FaceS+CAM        &          $8\times 8\times 512$       &         \Checkmark         &       \Checkmark       & \Checkmark \\ \midrule
					IA-FaceS$^*$-1      &         $1\times 512$          &         \Checkmark         &      \XSolidBrush       & \XSolidBrush \\ 
					IA-FaceS$^*$-4      &    $4\times 4\times 512$    &         \Checkmark         &      \XSolidBrush       & \XSolidBrush \\ 
					IA-FaceS$^*$-8      &        $8\times 8\times 512$         &         \Checkmark         &      \XSolidBrush       & \XSolidBrush \\ 
					IA-FaceS$^*$-16     & $16\times 16\times 512$ &         \Checkmark         &      \XSolidBrush       & \XSolidBrush \\ 
					IA-FaceS$^*$-16\dag &      $16\times 16\times 1$      &         \Checkmark         &      \XSolidBrush       & \XSolidBrush \\ 
					IA-FaceS$^*$-A    &        $8\times 8\times 512$         &  \XSolidBrush &      \XSolidBrush       & \XSolidBrush \\ \bottomrule
			\end{tabular}
		\caption{Summaries of the variants of IA-FaceS. Note that ``augmentation" denotes data augmentation, i.e., the input image has a half chance of being destroyed by a random irregular mask.}
		\label{tab:exp-summary}
	\end{table}
	
	\begin{table}[t]
		\centering
		\begin{tabular}{lccccc}
			\toprule
			& MSE(e-2) $\downarrow$ & LPIPS $\downarrow$ & PSNR$\uparrow$ & SSIM $\uparrow$ & FID$\downarrow$ \\ \midrule
			IA-FaceS             &         2.315         &       0.224        &     22.34      &      0.642      &      3.131      \\
			IA-FaceS$^*$(w/o lp) &         2.710         &       0.311        &     21.64      &      0.603      &      6.275      \\ \bottomrule
		\end{tabular}
		\caption{Reconstruction performance of IA-FaceS (with LPIPS loss) and IA-FaceS$^*$ (w/o LPIPS loss). lp denote the LPIPS loss. Both models are trained according to the settings described in Section \ref{sec:exp-settings}.}
		\label{tab:exp-aba-recon}
	\end{table}
	\begin{figure}[!htb]
		\subfigure[The results of reconstructions for corrupted images by IA-FaceS*-A (without data augmentation) and IA-FaceS* (with data augmentation). Recon. is short for reconstruction.]{
			\centering
			\includegraphics[width=\columnwidth]{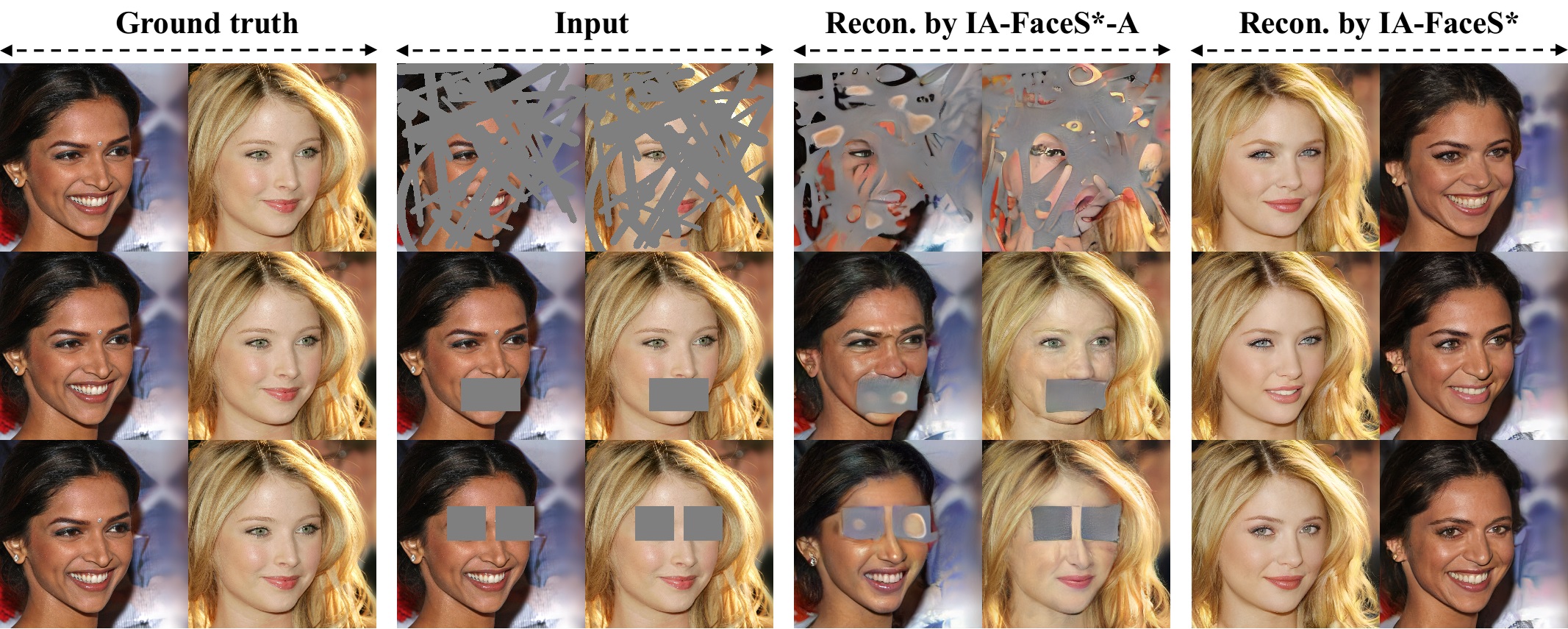}
			\label{fig:exp-aba-inpainting}
		}	
		\subfigure[IA-FaceS* can produce various inpainting results for the masked components.]{
			\centering
			\includegraphics[width=\columnwidth]{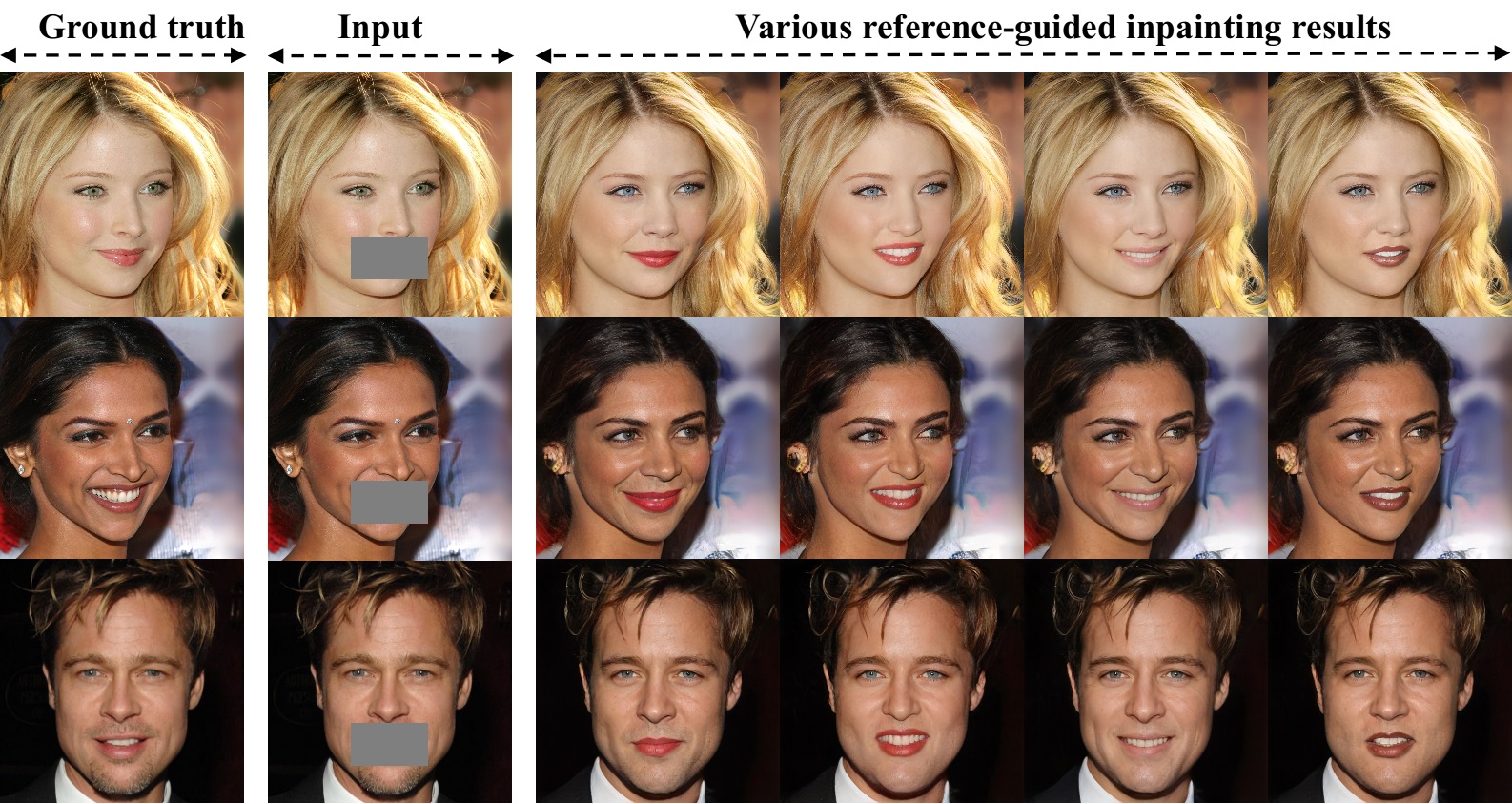}
			\label{fig:exp-aba-inpainting2}
		}
		\caption{Results of image completion for corrupted input images.}
		\label{fig:exp-aba-inpainting_all}
	\end{figure}
	
	\begin{table}[t]
		\centering
		\begin{tabular}{lcccccc}
			\toprule
			IA-FacsS$^*$-      &   1   &   4   &   8   &  16   & 16 $\dag$ &   A   \\ \toprule
			MSE $\downarrow$   & 0.096 & 0.060 & 0.029 & 0.013 &   0.073   & 0.026 \\ \midrule
			LPIPS $\downarrow$ & 0.449 & 0.388 & 0.297 & 0.195 &   0.416   & 0.308 \\ \midrule
			PSNR$\uparrow$     & 16.20 & 18.21 & 21.43 & 24.97 &   17.37   & 21.91 \\ \midrule
			SSIM $\uparrow$    & 0.473 & 0.526 & 0.612 & 0.729 &   0.510   & 0.630 \\ \midrule
			FID$\downarrow$    & 9.685 & 7.035 & 4.861 & 2.918 &   10.21   & 8.300 \\ \bottomrule
		\end{tabular}
		\caption{Reconstruction performance of the variants of the proposed framework. All variants are presented in Table \ref{tab:exp-summary}, and are trained according to the settings described in Section \ref{sec:exp-settings}, except that we set epoch=120 for time efficiency.}
		\label{tab:exp-aba-recon2}
	\end{table}
	
	\begin{figure}[ht]
		\subfigure[The results of reconstruction by the variants in different scenarios. For each pair, from left to right: normal reconstruction, reconstruction with component embeddings $\boldsymbol{z}=0$.]{
			\centering
			\includegraphics[width=\columnwidth]{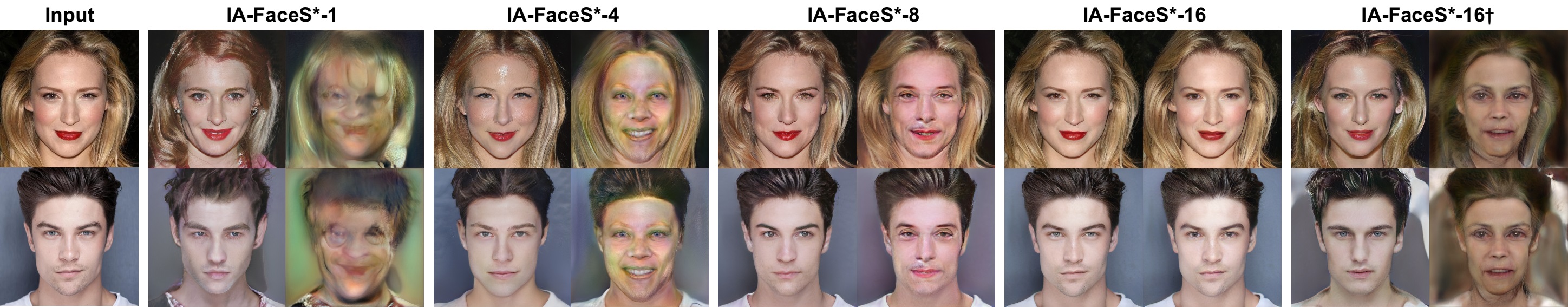}
			\label{fig:exp-aba-intervention}
		}	
		\subfigure[The results of manipulating the mouth from close to open by the variants. For each pair, from left to right: reconstruction, editing result of \emph{mouth open}. ]{
			\centering
			\includegraphics[width=\columnwidth]{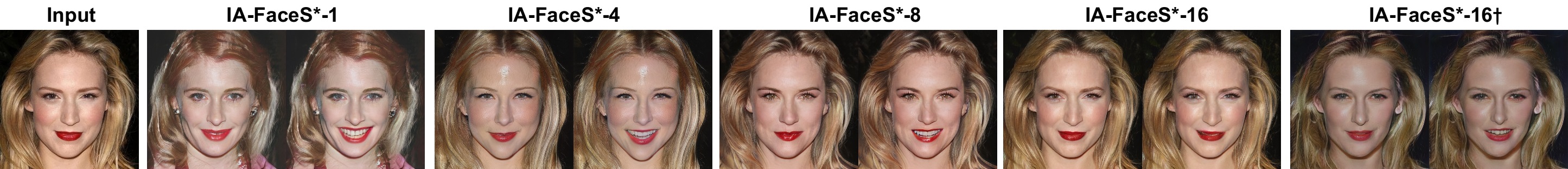}
			\label{fig:exp-aba-control}
		}
		\subfigure[The results of transferring mouth from reference to target by the variants. ]{
			\centering
			\includegraphics[width=\columnwidth]{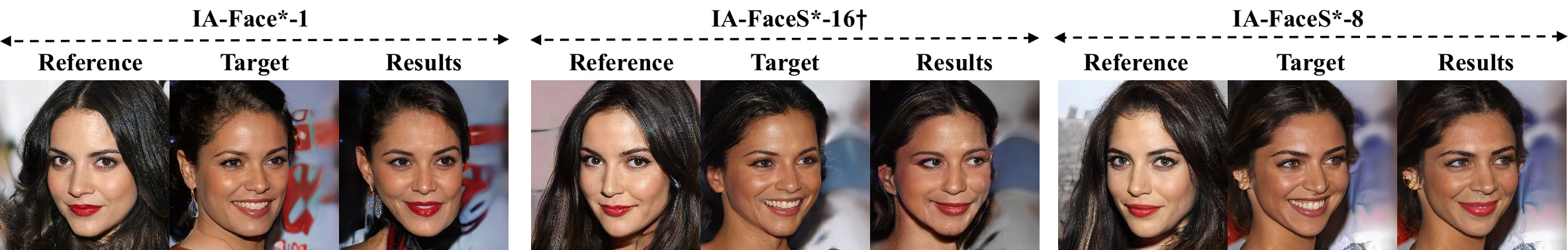}
			\label{fig:exp-aba-comtrsf}
		}
		
		\caption{Visualized effects of different resolutions of $\boldsymbol{S}$ on image reconstruction and component editing. From top to bottom: the results of (a)  intervention experiments, (b) face attribute manipulation, and (c) facial component transfer. The smaller the size of $\boldsymbol{S}$, the easier the model manipulates the face, but the worse the reconstruction accuracy. All variants are presented in Table \ref{tab:exp-summary}.}
	\end{figure}

	\paragraph{Loss Functions}
	The ablation study in Table \ref{tab:exp-aba-recon} shows that LPIPS loss can improve reconstruction accuracy and lead to higher FID scores. By comparing IA-FaceS$^*$-8 (i.e., IA-FaceS$^*$) and IA-FaceS$^*$-A in Table \ref{tab:exp-aba-recon2}, we can get that although the data augmentation (the input image has a half chance of being destroyed by a random irregular mask, see Eq. (\ref{eq:mask}) in Section \ref{sec:me-objective}) have a little negative influence on reconstruction accuracy, it enables our model to deal with corrupted images and perform image inpainting, as suggested in Figure \ref{fig:exp-aba-inpainting_all}. Moreover, it can also improve the qualities of the generated images (lower FID).

	\paragraph{Roles of Face Icon and Component Embeddings}
	According to the results shown in Figure \ref{fig:exp-aba-intervention}, the component areas in facial icon $\boldsymbol{S}$ are invariant to different faces, while the other parts vary with input faces. See the fourth column in Figure \ref{fig:exp-aba-intervention} for example, the decoder can recover the basic outline of a face in IA-FaceS$^*$-8 with only $\boldsymbol{S}$ (the right column in each pair), but no personal characteristics appear in facial components. When the component embeddings are activated, the component synthesis becomes personalized (the left column). The observations indicate that the component features are encoded into $\boldsymbol{z}$ rather than $\boldsymbol{S}$, allowing $\boldsymbol{z}$ to control components freely.  
	\paragraph{Effects of Resolutions of $\boldsymbol{S}$}
	We further investigate the effect of the size and dimensions of the face icon $\boldsymbol{S}$. By comparing the results of IA-FaceS$^*$-(1,4,8,16) in Table  \ref{tab:exp-aba-recon2} and Figure \ref{fig:exp-aba-intervention}, we find that as the spatial resolution increases, the reconstruction accuracy improves. This is expected because a larger $\boldsymbol{S}$ provides a stronger bottleneck layer of the encoder-decoder, encoding more details of the face. However, there exists a trade-off between reconstruction accuracy and component editing. As discussed in Section \ref{sec:me-synthesis}, we manipulate each component $c_i$ by modifying its embedding $\boldsymbol{z}_i$ while keeping $\boldsymbol{S}$ unchanged. If the size of $\boldsymbol{S}$ is too large, the face icon branch would compete with the component branch for capturing component-specific features. Then, the component embeddings would become relatively weak and be ignored in face synthesis.  As shown in the fifth column of Figure \ref{fig:exp-aba-intervention}, the component features are embedded in $\boldsymbol{S}$ of IA-FaceS$^*$-16 model, thus it can hardly manipulate the mouth in Figure \ref{fig:exp-aba-control}. However, it is easy for IA-FaceS$^*$-(1,4,8) to manipulate components since the features are embedded in the component embedding  $\boldsymbol{z}$ instead of $\boldsymbol{S}$. The face icon is designed to preserve image topology while not influencing the control of component embeddings. A larger $\boldsymbol{S}$ can reconstruct more details, but it would constrain the flexibility of component editing. 
	
	On the contrary, only a little information is embedded in the one-dimensional and two-dimensional $\boldsymbol{S}$, while most of the face information is embedded in $\boldsymbol{z}$, including the pose and outlines of the face. This brings flexible control over the components, but it will also change the face's outline when editing a component, leading to unexpected inconsistency with the original face. For example, in Figure \ref{fig:exp-aba-comtrsf}, when transferring the mouth from reference to target by IA-FaceS$^*$-(1,16\dag), the pose of the target image is also changed, which violates the purpose of independent component editing, i.e., keeping component-irrelevant regions unchanged. 
	
	In summary, there exists a trade-off between reconstruction accuracy and component editing. The resolution of $\boldsymbol{S}$ can seriously influence the performance of IA-FaceS. $8 \times 8 \times 512$ balances well between reconstruction accuracy and flexible component editing.

	\subsection{Limitations and Failure Cases}   \label{sec:exp-limitation}
	
	\begin{figure}[t]
		\centering
		\includegraphics[width=\columnwidth]{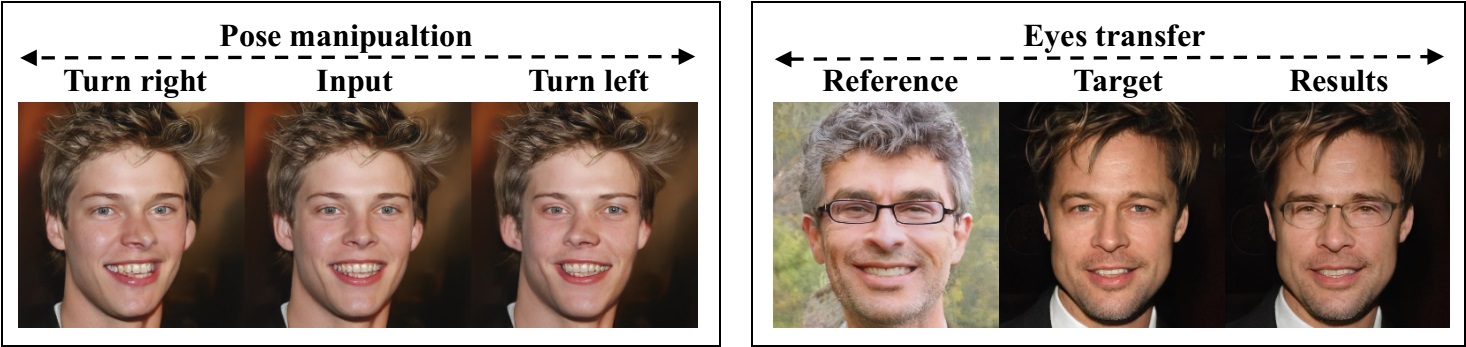}
		\caption{Failure cases of IA-FaceS. Left: attribute manipulation results of \emph{pose}. The facial components successfully turn from left to right, but the face outline has no change, leading to a strange synthesis output. Right: the results of transferring eyes from reference to target. Although the eyes are successfully transferred,  the glasses are not successfully transferred to the target face (it fails to keep the glasses frames).}
		\label{fig:exp-failure}
	\end{figure}
	Despite the convincing experimental results, there are some limitations to IA-FaceS. Figure \ref{fig:exp-failure} shows typical failure cases of IA-FaceS.  The first limitation is the inability to change the face outline. IA-FaceS focuses on component editing, while the manipulation of the overall face can be limited. As discussed in Section \ref{sec:me-decoder}, we edit the face by component embeddings while keeping $\boldsymbol{S}$ unchanged. Thus, IA-FaceS struggles with attributes related to the face outline. Take Figure \ref{fig:exp-failure} as an example, when editing the orientation of a face, IA-FaceS can only edit the orientation of components, while the orientation of the face outline cannot be changed. 
	
	Another limitation is that the proposed method cannot generalize to an arbitrary dataset. The computing of component embeddings in IA-FaceS heavily depends on their fixed positions, meaning that our framework is only fit for the data with constant component locations, or the components can be aligned to fixed positions. It is our future direction to generalize IA-FaceS to unaligned datasets. 
	
	Furthermore, while this paper aims to achieve component-level disentanglement and relieve users from painting skills, we cannot deny that the disentanglement is restricted without segmentation masks.
	\section{Conclusion} \label{sec:conclusion}
	
	Most existing face manipulation methods embed a face into a single low-dimensional vector in the latent space, making it difficult for them to reconstruct image details and conduct flexible component editing. Other mask-guided or sketch-guided methods require users to edit the semantic segmentation mask or sketches to provide visual guidance before changing the component. In this paper, we presented a bidirectional method for flexibly editing facial components without editing the masks or sketches in visual space. We embedded a face into a high-dimensional face icon and four low-dimensional component embeddings, separately. The face icon enhances the consistency between the output and the original face image, while the component embedding captures the facial component's features and allows the model to control or transfer the components. Moreover, we introduced a CAM (component adaptive modulation) module into the decoder to disentangle highly-correlated components, i.e., the left eye and right eye. Experimental results demonstrated the effectiveness of the proposed method in independent component editing and disentangled attribute manipulation. To the best of our knowledge, IA-FaceS+CAM is the first latent space manipulation method to edit a single eye without any additional visual guidance.
	
    \section*{Acknowledgment}
	This work was supported by The National Key Research and Development Program of China (2018AAA0100700) of the Ministry of Science and Technology of China, and Shanghai Municipal Science and Technology Major Project (2021SHZDZX0102). Shikui Tu and Lei Xu are corresponding authors.

	\bibliography{main}
	\newpage
	\appendix
	
	\section{Overview}
	This supplementary material is organized as follows: 
	
	\begin{enumerate}
		\item \ref{sec:sup-implementation} supplements the implementation details of the compared baselines.
		\item \ref{sec:sup-implementation-our} describes the implementation details of our approach, including network architectures (\ref{sec:sup-network}), training details (\ref{sec:sup-training}), and the visualizations of the perturbed input in training phase (\ref{sec:sup-perturb-input}), etc.
		\item \ref{sec:sup-experiments} shows additional experimental results.
	\end{enumerate}
	For better observation, all the images reported in the manuscript and this supplementary material are recommended to zoom in and view in color.
	
	\section{Implementation Details of the Compared Baselines}   \label{sec:sup-implementation}
	For StyleGAN2 \cite{stylegan2}, we adopt a popular PyTorch implementation\footnote{\url{https://github.com/rosinality/stylegan2-pytorch}} (1.2k stars), which is able to reproduce the results in the original paper. We train StyleGAN2 on CelebA-HQ \cite{pg-gan} 256px from scratch for 250k iterations with batch size 16. When projecting an image back into the latent space, we use the method described in the original work \cite{stylegan2} with 1000 iterations. (We also used the official PyTorch implementation\footnote{\url{https://github.com/NVlabs/stylegan2-ada-pytorch}} of StyleGAN2-ADA \cite{karras2020training} without ADA and train on CelebA-HQ \cite{pg-gan} following the default settings for 256px. The reconstruction performance of the official implementation seems not good as the unofficial one on CelebA-HQ 256px. Thus, we take the unofficial implementation as the baseline.)
	
	For ALAE \cite{alae}\footnote{\url{https://github.com/podgorskiy/ALAE}} , StyleMapGAN\cite{stylemapgan}\footnote{ \url{https://github.com/naver-ai/StyleMapGAN}}, StarGANV2 \cite{starganv2}\footnote{\url{https://github.com/clovaai/stargan-v2}}, SEAN \cite{sean}\footnote{\url{https://github.com/ZPdesu/SEAN}} and PSP \cite{psp}\footnote{\url{https://github.com/eladrich/pixel2style2pixel}}, we adopt their official implementations and use their released models on CelebA-HQ \cite{pg-gan} 256px or FFHQ \cite{stylegan} 1024px.
	
	For AttGAN \cite{attgan}, we adopt the PyTorch implementation\footnote{\url{https://github.com/elvisyjlin/AttGAN-PyTorch.git}} recommended in their official TensorFlow implementation \footnote{\url{https://github.com/LynnHo/AttGAN-Tensorflow}} and use the released model. For computing the attribute editing accuracy in Table \ref{tab:exp-attr-accuracy} in the manuscript, we also retrained AttGAN on the compared local attributes for a fair comparison. However, the re-trained AttGAN fails to edit any attribute. Finally, we use the released model and compute the editing accuracy only on several common attributes.
	
	For InterFaceGAN \cite{interfacegan}, we adopt its official implementation\footnote{\url{https://github.com/genforce/interfacegan}} and apply their single attribute manipulation method to StyleGAN2 \cite{stylegan2}, ALAE \cite{alae}, and our framework. 
	
	For MGPE \cite{mask-guided} (mask-guided portrait editing), we adopt the official implementation\footnote{\url{https://github.com/cientgu/Mask_Guided_Portrait_Editing}}. The released model is trained on a dataset combined by Helen Dataset \cite{smith2013exemplar} and VGGFace2 \cite{cao2018vggface2}. Thus, we evaluate MGPE on the images in Helen dataset.
	
	\section{Implementation Details of Our Method} \label{sec:sup-implementation-our}
	In this section, we provide more details of the implementations of our methods. 
	\subsection{Network Architectures} \label{sec:sup-network}
\begin{figure}[ht]
		\centering
		\includegraphics[width=\textwidth]{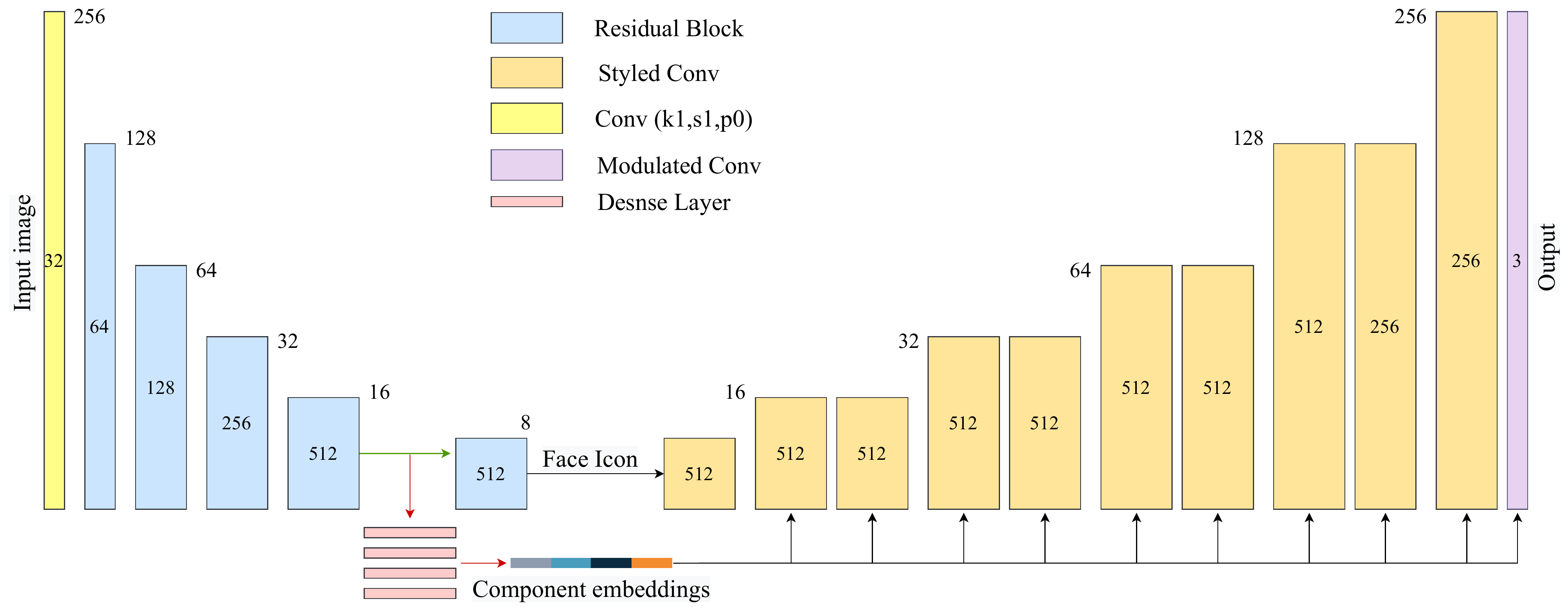}
		\caption{The architecture of IA-FaceS. The encoder network first applies 4 down-sampling residual blocks \cite{he2016deep} to produce an intermediate feature tensor, which is then passed to two separate branches, producing the facial component embedding (red line) and face icon (green line). The decoder $G$ is composed of 10 styled convolution layers and the details of $G$ follow the settings in StyleGAN2 \cite{stylegan2}, including weight demodulation, bi-linear up-sampling,  noise injection, and leaky ReLU with slope 0.2.}
		\label{fig:sup-network-arch}
	\end{figure}

	The network structures are described in detail in Figure \ref{fig:sup-network-arch}. As shown, the backbone of $F_1$ comprises 4 down-sampling residual blocks \cite{he2016deep} for images at $256 \times 256$ (6 down-sampling residual blocks for images at $1024 \times 1024$). $F_2$ is a down-sampling residual block. The decoder $G$ follows the settings in StyleGAN2 \cite{stylegan2} except for the number of channels. The discriminator $D$ is the same as that in StyleGAN2.
	
	\subsection{Additional Training Details} \label{sec:sup-training}
	
	For training IA-FaceS on FFHQ 1024px, we use a machine with $4 \times$ Titan V GPUs with batch size 4 and train for 58 epochs (about 967k iterations). For training IA-FaceS$^*$+CAM on FFHQ 1024px, we use $4 \times$ Titan V GPUs with batch size 8 and train for 80 epochs (about 662k iterations). 
	
	The loss for the image discriminator is computed on the real and reconstructed images, using the adversarial loss $L_{adv}(G,D) = \mathbb{E}[log(D(\boldsymbol{x}))]+\mathbb{{E}}[log(1-D(\boldsymbol{\hat{x}}))]$. For the details of GAN loss, we follow the settings of StyleGAN2 \cite{stylegan2}, including the non-saturating GAN loss \cite{gan} and lazy R1 regularization \cite{mescheder2018training,stylegan2}. 
	
	\subsection{Visualizations of Perturbed Input} \label{sec:sup-perturb-input}
\begin{figure}[ht]
		\centering
		\includegraphics[width=\columnwidth]{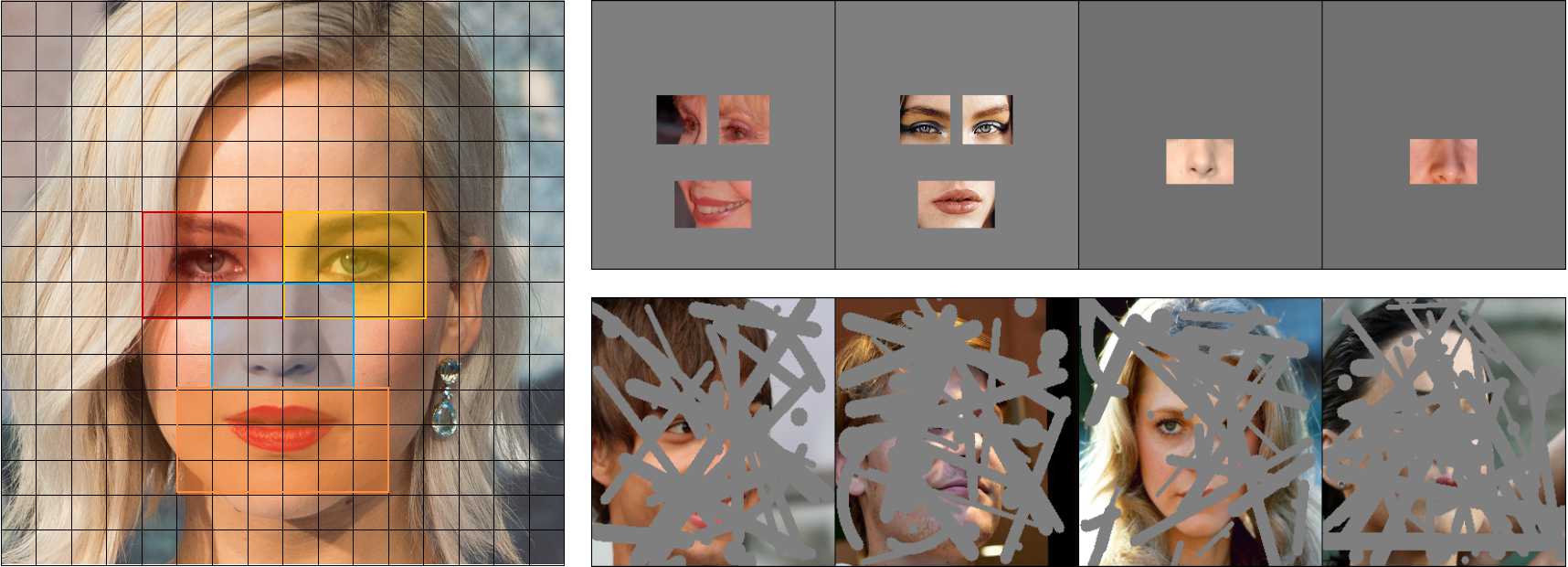}
		\caption{Left: The spatial distribution of the re-scaled boxes $b_i$ in the intermediate feature tensor $\boldsymbol{\phi}$ of size $16 \times 16\times 512$. Top Right: Examples of the facial components segmented by their corresponding boxes $B_i$ in the visual space. Since the area of the nose overlaps that of other components, we separately show the regions of the nose and other components. Bottom right: examples of images perturbed by random, irregular masks.}
		\label{fig:sup-inputs}
	\end{figure}

	As mentioned in Section \ref{sec:me-objective} of the manuscript, the input image has a half chance of being perturbed by a random irregular mask. The random irregular masks are generated by randomly drawing some circles and lines, following the method introduced in a Keras implementation \footnote{\url{https://github.com/MathiasGruber/PConv-Keras}} of Partial Convolution \cite{pconv}. Examples of the perturbed input are shown in the bottom right corner of Figure \ref{fig:sup-inputs}.
	
	\subsection{Visualizations of Fixed Latent Mask}
	In Section \ref{sec:me-encoder} of the manuscript, we apply a fixed, simple latent mask to segment the component regions on the intermediate feature tensor $\boldsymbol{\phi}$, where $b_i$ is resized from the predefined component regions in the visual space, namely, $B_i$. Here, we visualize the distribution of $b_i$ on $\boldsymbol{\phi}$ ($16\times16\times512$) in the left most column of Figure \ref{fig:sup-inputs} and show the distribution of $B_i$ in the original image on the top right of Figure \ref{fig:sup-inputs}.
	
	\section{Additional Experimental Results}   \label{sec:sup-experiments}
	
	\subsection{Attribute-irrelevant Regions} \label{sec:sup-exp-metrics}
	
\begin{figure}[ht]
		\centering
		\includegraphics[width=\textwidth]{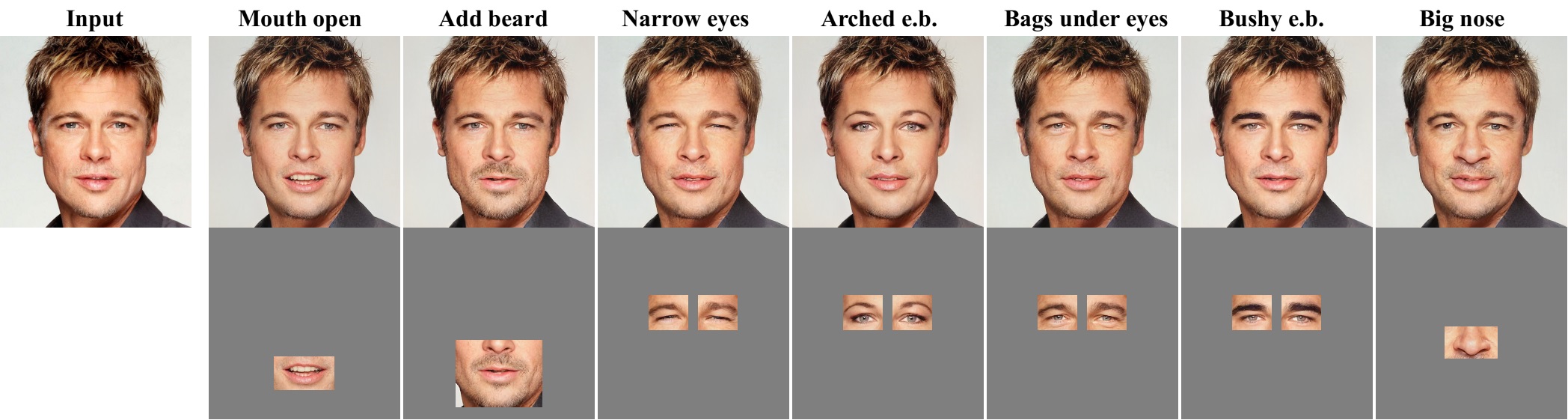}
		\caption{Predefined attribute-irrelevant regions. For each attribute, the gray region is defined as the attribute-irrelevant region,
			which should not be changed when editing the corresponding attribute. The irrelevance preservation error is calculated as the $L_2$ difference
			of the gray region between the editing result and the original reconstruction of the input image.}
		\label{fig:sup-exp-attr-region}
	\end{figure}

	In Section \ref{sec:exp-metrics} of the manuscript, we describe the irrelevance preservation metric, i.e., $\text{MSE}_\text{irr}$, which is calculated as the $L_2$ difference of a predefined attribute-irrelevant region between the editing result and the original reconstruction. Figure \ref{fig:sup-exp-attr-region} shows the predefined attribute-irrelevant regions for all the appeared local attributes. As shown in Figure \ref{fig:sup-exp-attr-region}, the attribute-irrelevant region is defined as the area outside the attribute-related components in the image.
	
	\subsection{Reconstruction} \label{sec:sup-exp-recon}
	We present quantitative comparisons with more existing methods in Table \ref{tab:sup-recon}, where the results of the competitors are copied from Table 3 in \cite{stylemapgan}. All the metrics are calculated on 5,00 images provided in the official implementation \footnote{\url{https://github.com/naver-ai/StyleMapGAN}} of StyleMapGAN \cite{stylemapgan}. Thus, we only compute the metrics of IA-FaceS on the provided 5,00 images for comparison.  According to Table \ref{tab:sup-recon}, our method ranks second in reconstruction accuracy. Although Image2StyleGAN \cite{stylemapgan} shows the best reconstruction accuracy, it suffers from runtime and has bad support for interpolations \cite{indomain,stylemapgan}, which limits its application in practical applications. Note that we do not report the runtime of IA-FaceS in Table \ref{tab:sup-recon} because the runtime is related to computer and GPU. Instead, we calculate the average per-image runtime of IA-FaceS and StyleMapGAN on 5,00 images in our machine with one GeForce GTX 1080Ti. The results are 0.0869 (StyleMapGAN) vs \textbf{0.0358} (Ours). Thus, we can conclude that our method outperforms other competitors on the speed of inference.
	
\begin{figure}[ht]
		\centering
		\includegraphics[width=\textwidth]{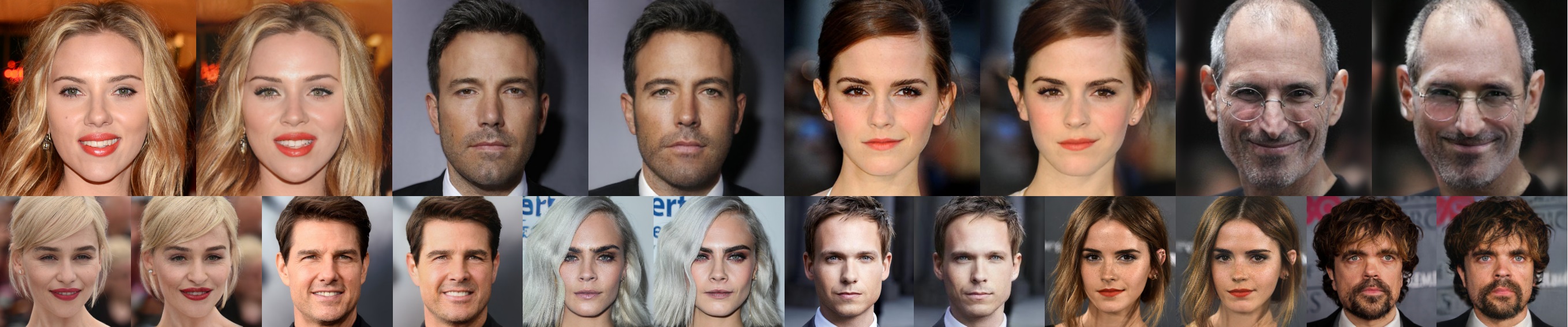}
		\caption{Real image reconstructions. Reconstructions of real-world images with IA-FaceS trained on CelebA-HQ \cite{pg-gan} at $256\times256$. For each pair, the left column is ground truth, the right column is the reconstruction by IA-FaceS.}
		\label{fig:sup-exp-recon}
	\end{figure}

	Additional qualitative results of image reconstructions by IA-FaceS are shown in Figure \ref{fig:sup-exp-recon}.
	
	\begin{table}[ht]
		\centering
		\begin{tabular}{lccr}
			\toprule
			Model                                              & Runtime $\downarrow$ & MSE $\downarrow$ & LPIPS $\downarrow$ \\ \toprule
			StyleGAN2 \cite{stylegan2}                         &       80.04        &      0.079       &              0.247 \\
			Image2StyleGAN \cite{image2stylegan}               &       192.5        &  \textbf{0.009}  &     \textbf{0.203} \\
			In-DomainGAN \cite{indomain}                       &        6.8         &      0.052       &              0.340 \\
			Structured Noise \cite{structured-noise-injection} &        64.4        &      0.097       &              0.256 \\ \hline
			SEAN  \cite{sean}                                  &       0.146        &      0.064       &              0.334 \\
			StyleMapGAN \cite{stylemapgan}                     &   \textbf{0.082}   &      0.024       &              0.242 \\
			IA-FaceS                                           &         -          &  \textbf{0.023}  &     \textbf{0.221} \\ \bottomrule
		\end{tabular}
		\caption{Comparisons of real image projection on 5,00 CelebA-HQ \cite{pg-gan} images at $256\times 256$. The results of the competitors are copied from Table 3 in \cite{stylemapgan}. Runtime covers the end-to-end interval of projection and generation in seconds. The horizontal line between method separates optimization-based and encoder-based methods.}
		\label{tab:sup-recon}
	\end{table}
	
	\begin{figure}[!htb]
		\centering
		\includegraphics[width=\columnwidth]{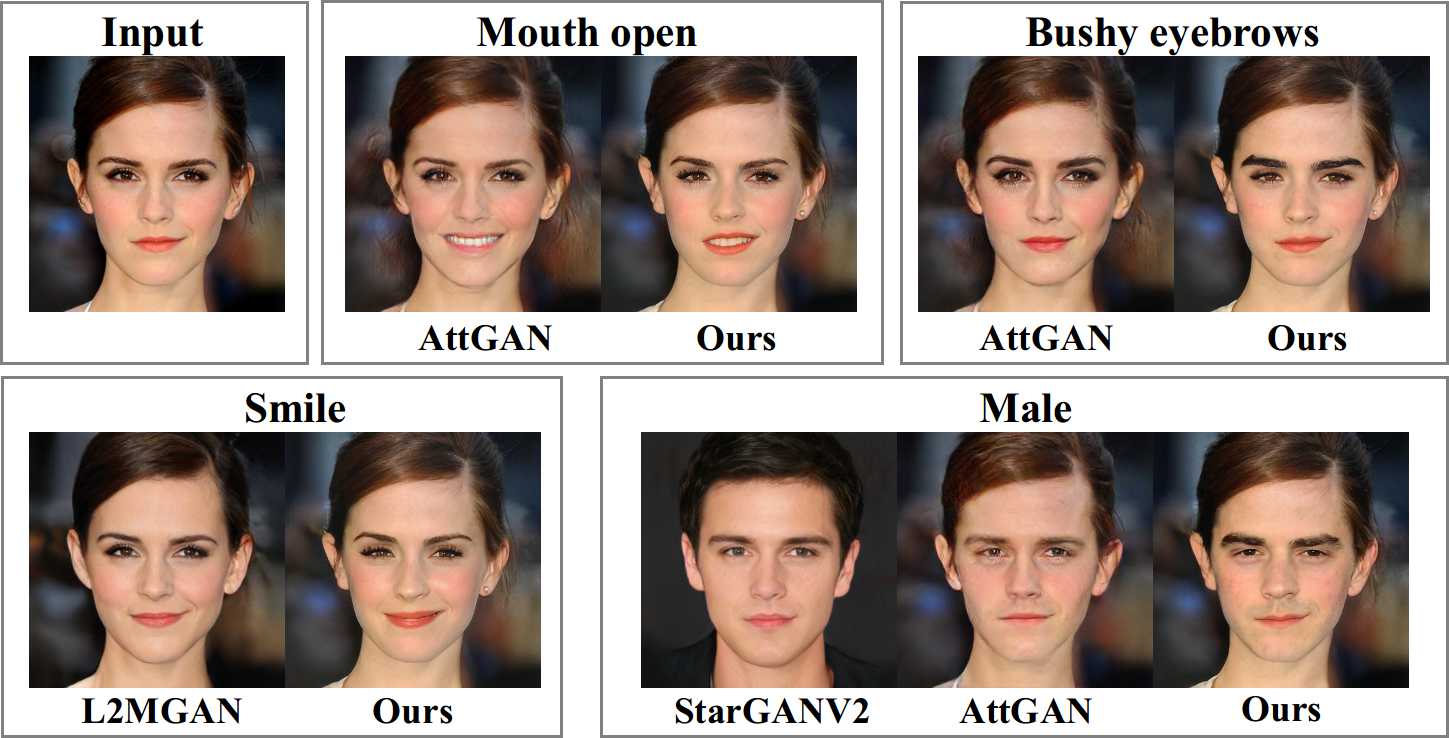}
		\caption{Comparisons with three face attribute editing models. Since the three image-to-image translation methods are limited to their predefined attributes (smile for L2MGAN \cite{l2mgan}, male for StarGANV2 \cite{starganv2}, 13 attributes for AttGAN \cite{attgan}), we only compare the performance on several common attributes.}
		\label{fig:sup-attr-comprare}
	\end{figure}
	
	\subsection{Attribute manipulation} \label{sec:sup-attr}
	
	Firstly, we present additional qualitative comparisons with the image-to-image translation models omitted in the manuscript, i.e., AttGAN \cite{attgan}, StarGANV2 \cite{starganv2}, and L2MGAN \cite{l2mgan}. As can be seen, image-to-image translation models are often limited to the predefined attributes and cannot continuously manipulate the attribute. In contrast, the latent space manipulation methods are free to continuously edit arbitrary attributes even if the model has been trained.
	
\begin{figure}[ht]
		\centering
		\includegraphics[width=\columnwidth]{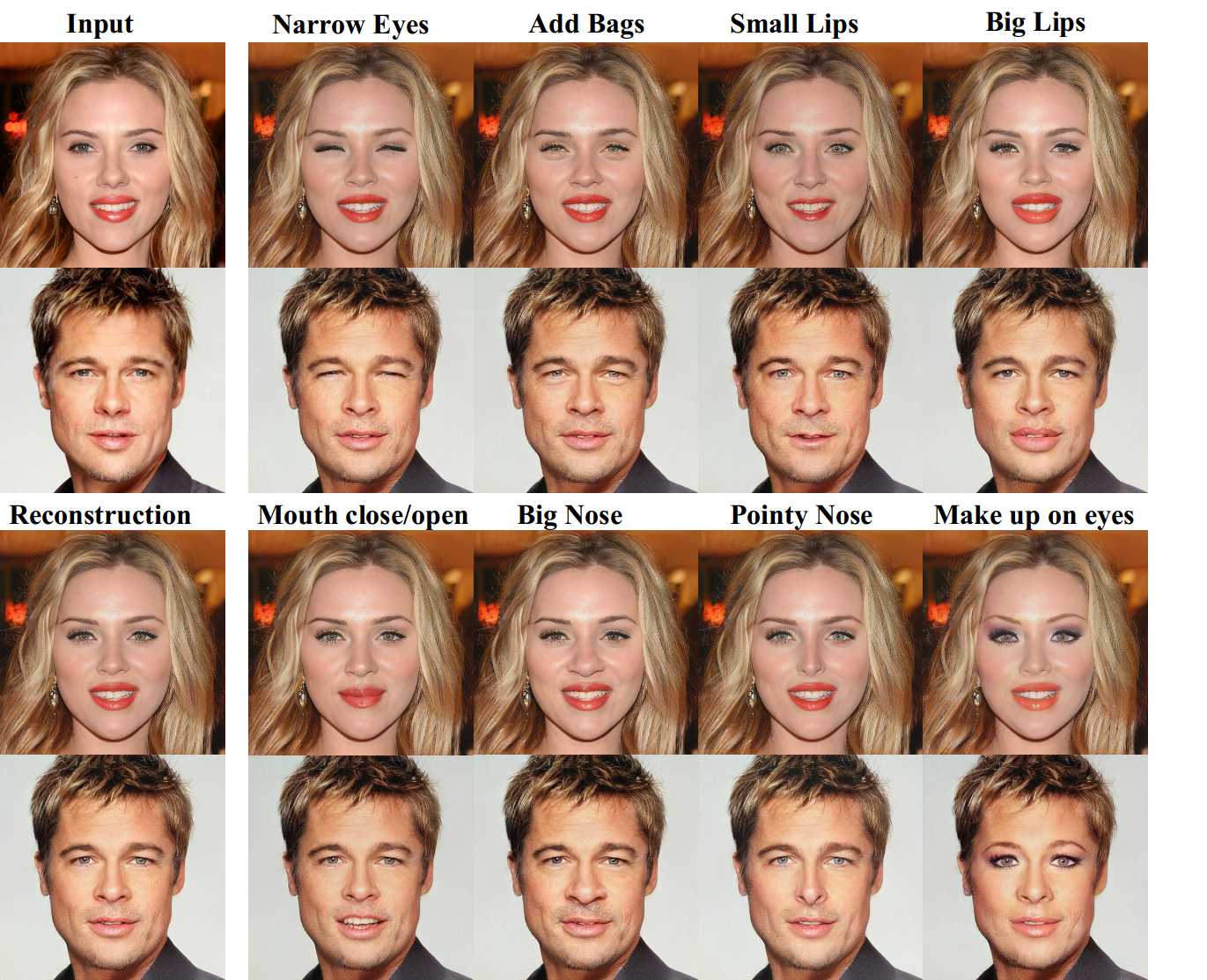}
		\caption{Qualitative results of attribute manipulation by IA-FaceS.}
		\label{fig:sup-attr-iafaces}
	\end{figure}

\begin{figure}[ht]
		\centering
		\includegraphics[width=\columnwidth]{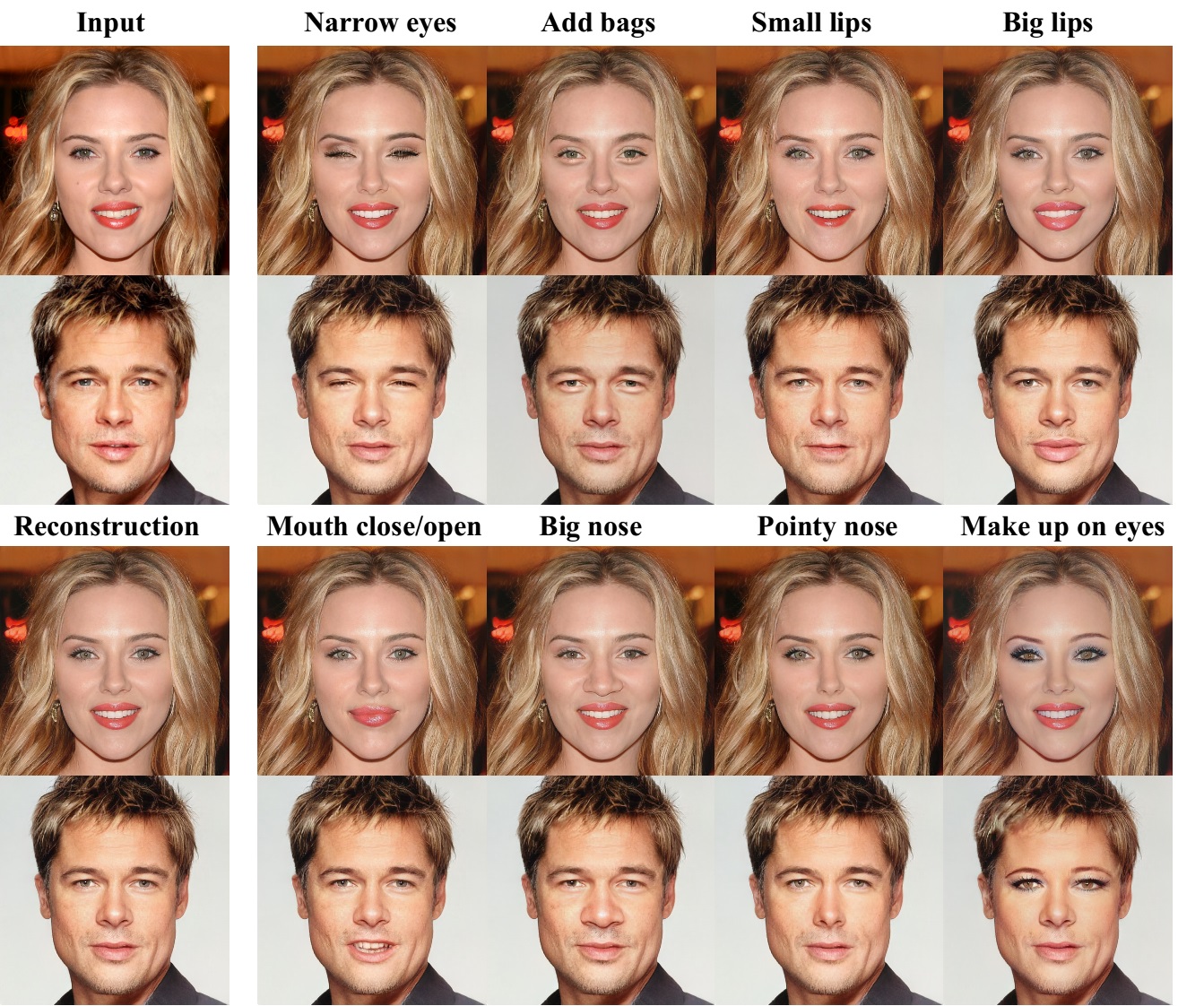}
		\caption{Qualitative results of attribute manipulation by IA-FaceS+CAM.}
		\label{fig:sup-attr-cam}
	\end{figure}

	Secondly, we provide more qualitative examples of attribute manipulation by IA-FaceS and IA-FaceS+CAM in Figure \ref{fig:sup-attr-iafaces} and Figure \ref{fig:sup-attr-cam}. Overall, both of the two architectures can generate correct attributes with high image fidelity and well preserve the irrelevant characteristics.

	\subsection{Component Transfer} \label{sec:sup-comtrsf}
	
\begin{figure}[!ht]
		\includegraphics[width=\columnwidth]{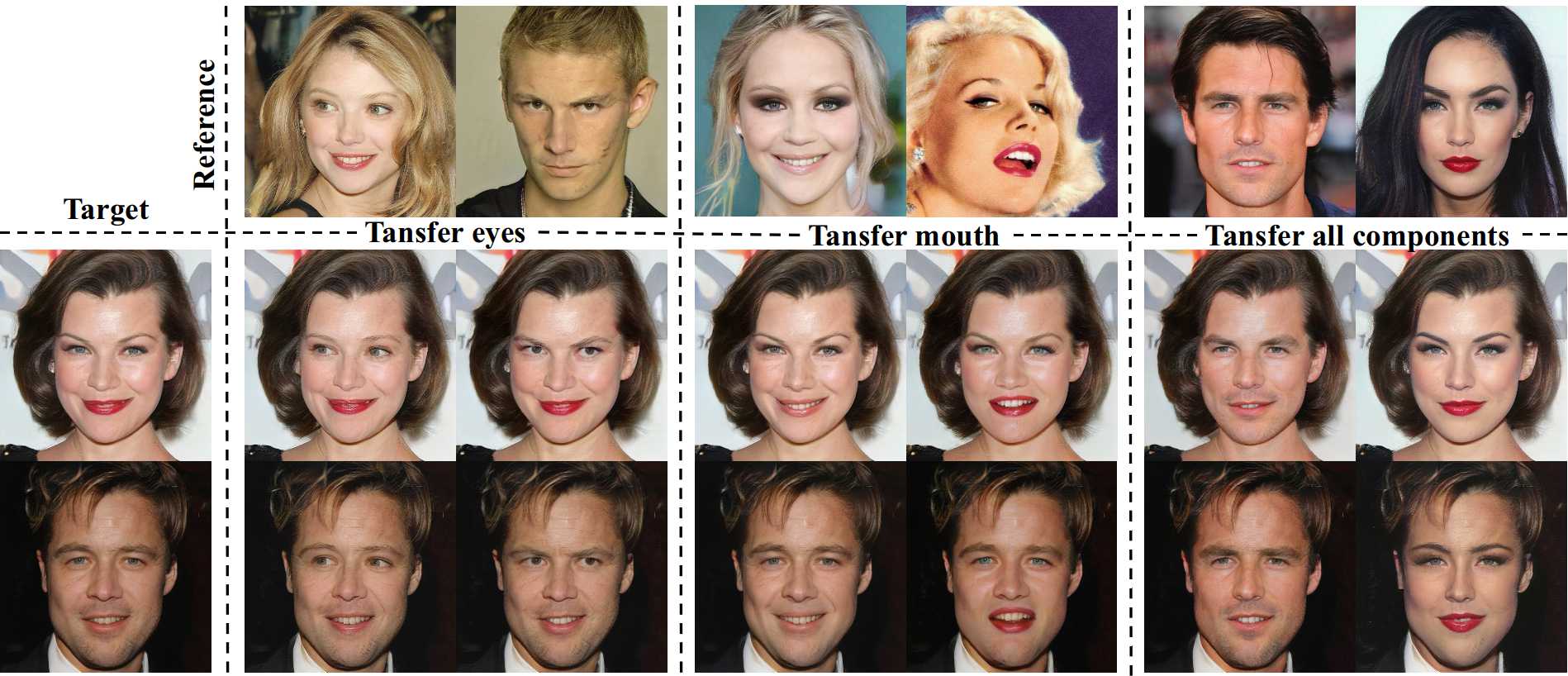}
		\caption{More examples of transferring specific components from the reference image to the target image by IA-FaceS. Both the shape and texture of the component can be well transferred to the target images.}
		\label{fig:sup-our-com-trsf}
	\end{figure}

	Firstly, additional examples of component transfer by the proposed methods are provided in Figure \ref{fig:sup-our-com-trsf}.
	
\begin{figure}[ht]
		\centering
		\includegraphics[width=\columnwidth]{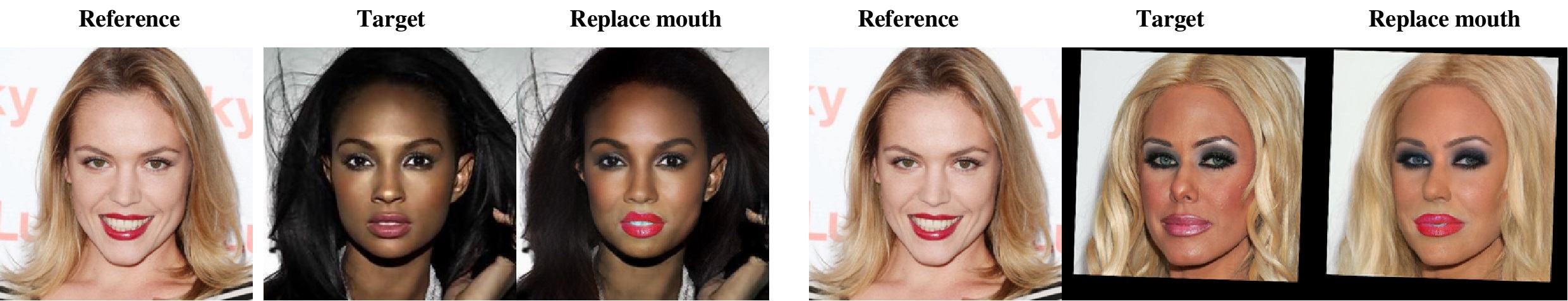}
		\caption{Examples of mouth transfer by MGPE \cite{mask-guided} on Helen dataset \cite{smith2013exemplar}. Only the textures of mouth can be transferred to the target image.}
		\label{fig:sup-mgpe-replace}
	\end{figure}

\begin{figure}[ht]
		\centering
		\includegraphics[width=\columnwidth]{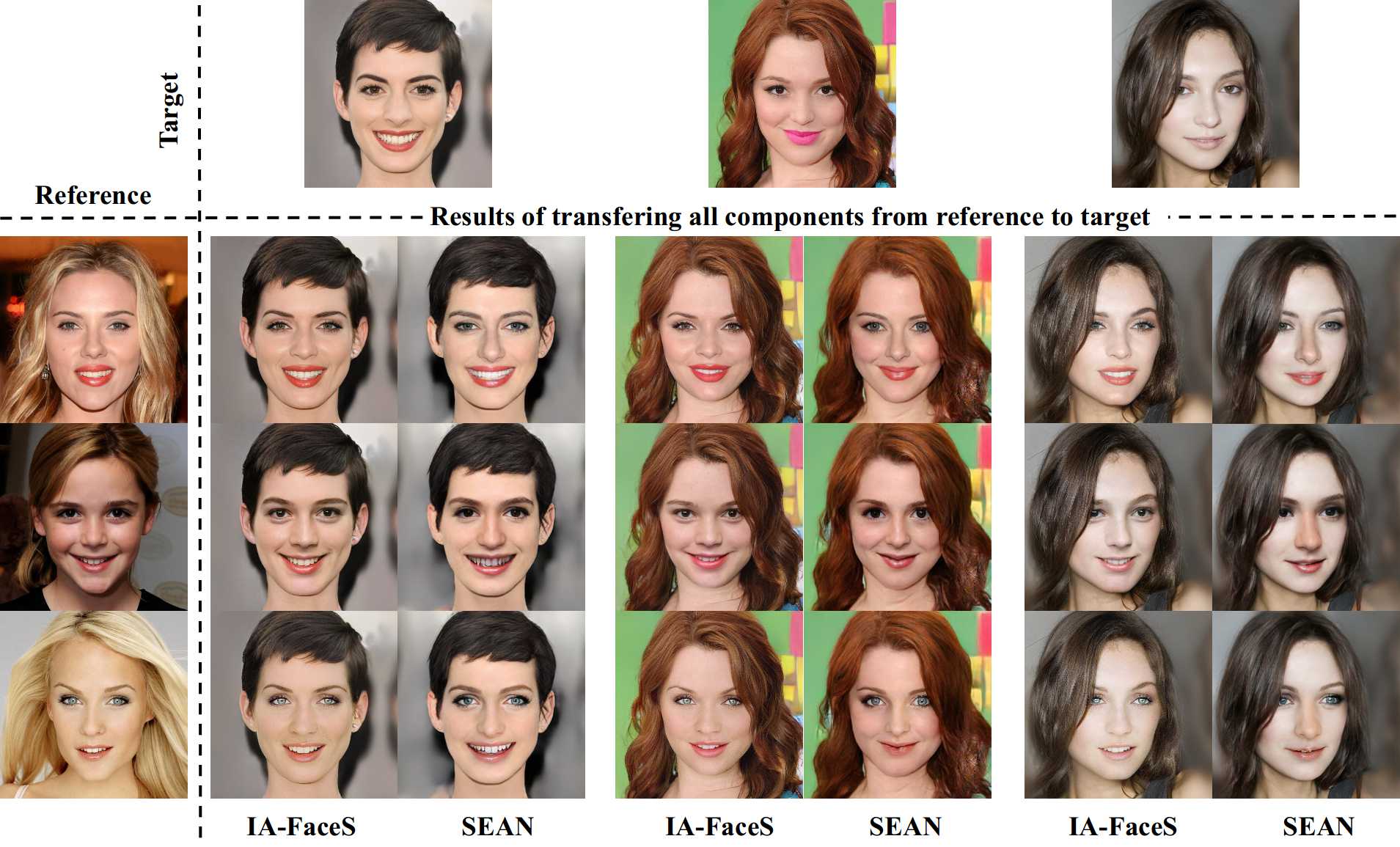}
		\caption{Examples of transferring all the components (left eye, right eye, nose, mouth) from the reference image (the first column) to the target image (the first row). For each pair, the left column is the result of IA-FaceS, the right column is the result of SEAN. By IA-Faces, both the shape and textures of the components can be well transferred to the target images. By SEAN \cite{sean}, only the textures can be transferred, and it can produce artifacts sometimes.}
		\label{fig:sup-com-sean}
	\end{figure}

	Moreover, we qualitatively compare with two segmentation-based methods, MGPE \cite{mask-guided} and SEAN \cite{sean}, on component transfer in Figure \ref{fig:sup-mgpe-replace} and Figure \ref{fig:sup-com-sean}, separately. As shown, they can only transfer the textures (lipsticks of mouth) of the components. Besides, SEAN \cite{sean} can produce color distortions (in the last column of Figure \ref{fig:sup-com-sean}, the nose color is inconsistent with the skin color) and artifacts (mouth of the first target, hair of the second target). In contrast, our method can naturally transfer both the textures and shapes of the components from reference to target while maintaining the fidelity and consistency of the generated image, as suggested in Figure \ref{fig:sup-our-com-trsf} \& Figure \ref{fig:sup-com-sean}.
	
	\subsection{Performance on High-resolution Images} \label{sec:sup-1024}
	
\begin{figure}[ht]
		\centering
		\includegraphics[width=\columnwidth]{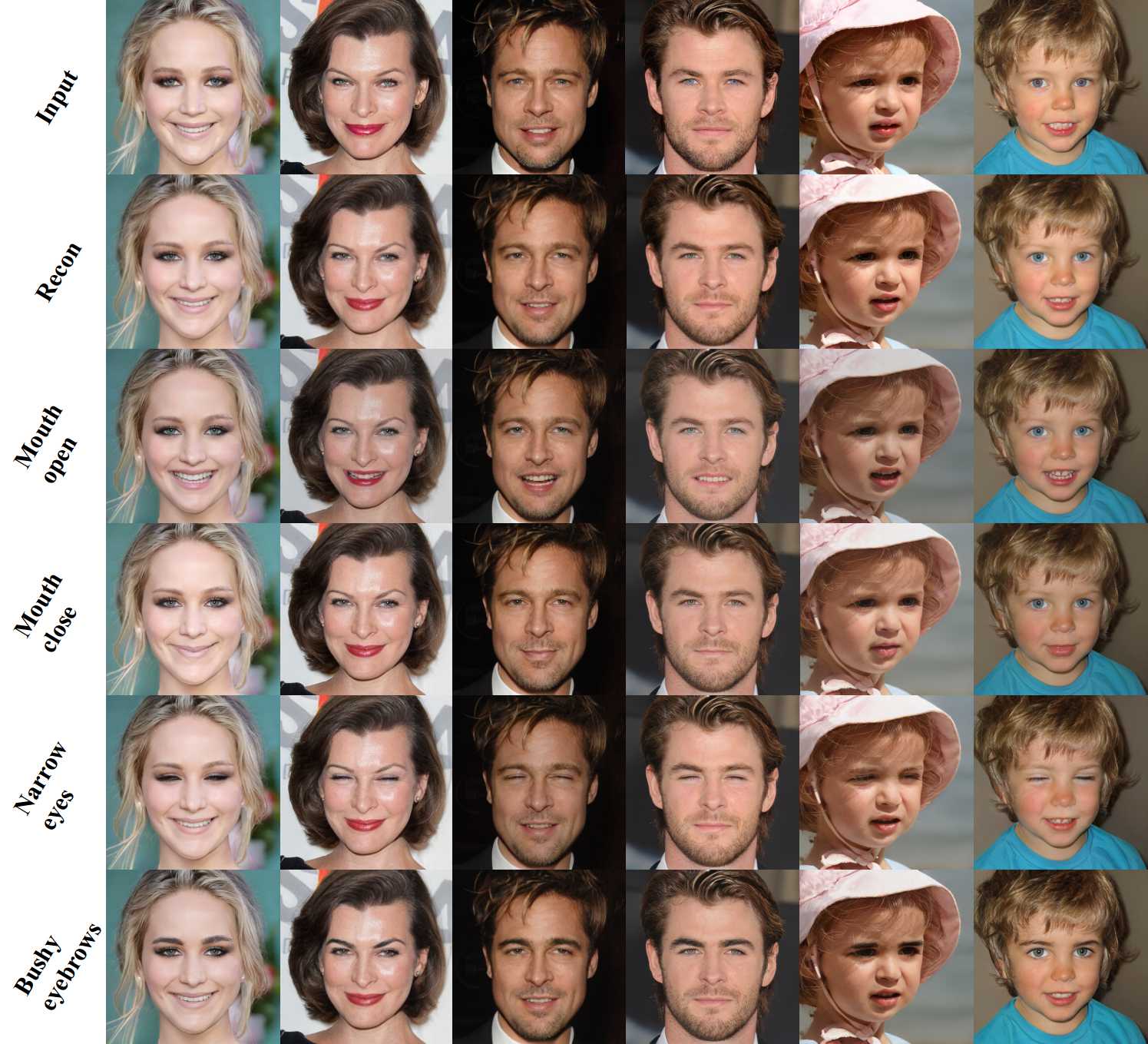}
		\caption{Attribute manipulation by IA-FaceS on real world images at $1024 \times 1024$. Note that the attribute directions are calculated on CelebA-HQ \cite{pg-gan}, which provides attribute annotations. Recon is short for reconstruction.}
		\label{fig:sup-ffhq-attr}
	\end{figure}

\begin{figure}[ht]
		\centering
		\includegraphics[width=\columnwidth]{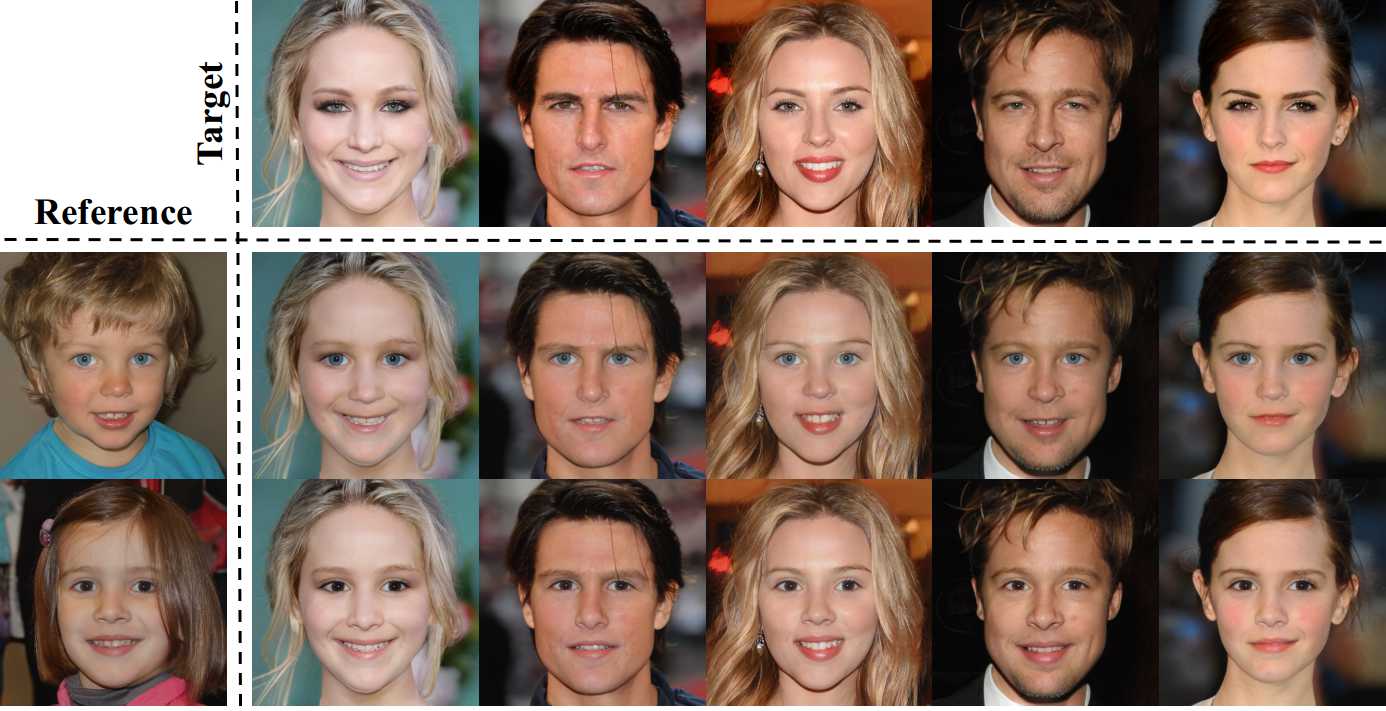}
		\caption{Component transfer by IA-FaceS on real world images at $1024 \times 1024$. All the components (i.e., eyes, nose, and mouth) are successful transferred from the reference image to the target image.}
		\label{fig:sup-ffhq-comptrsf}
	\end{figure}

\begin{figure}[ht]
		\centering
		\includegraphics[width=\columnwidth]{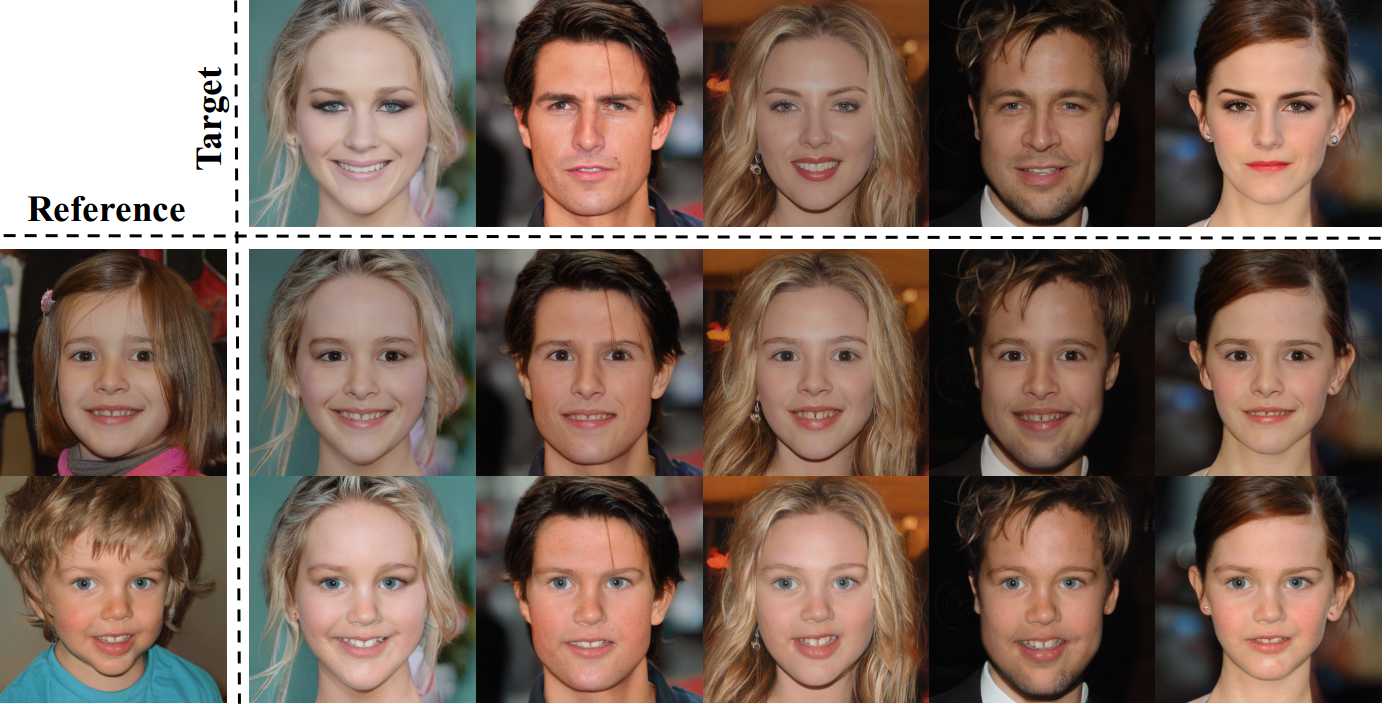}
		\caption{Component transfer by IA-FaceS$^*$+CAM on real world images at $1024 \times 1024$. All the components (i.e., eyes, nose, and mouth) are successful transferred from the reference image to the target image.}
		\label{fig:sup-ffhq-comptrsf-cam}
	\end{figure}

	This section provides evaluations of the proposed method trained on FFHQ \cite{stylegan} images at  $1024 \times 1024$ for various applications, i.e., attribute manipulation (Figure \ref{fig:sup-ffhq-attr}), and component transfer (Figure \ref{fig:sup-ffhq-comptrsf} \& Figure \ref{fig:sup-ffhq-comptrsf-cam}), to evident the scalability of our method on high-resolution images.
	
	\section{Additional Materials for Synthesis Reasoning} \label{sec:sup-svm}
	
	For a binary classification problem, $\{x^i \text{,} y^i\}|_{i=1}^N$ are the collected positive and negative samples. Here, a linear SVM is trained to obtain the hyperplane that separates the two classes. To calculate the hyperplane, we need to solve the following constrained optimization problem,
	\begin{align}
		\min_{\eta} & \quad \frac{1}{2} \sum_{i=1}^{N} \sum_{j=1}^{N} \eta_{i} \eta_{j} y_{i} y_{j}\left(x_{i} \cdot x_{j}\right)-\sum_{i=1}^{N} \eta_{i} \\ \nonumber
		\text { s.t. } & \quad \sum_{i=1}^{N} \eta_{i} y_{i}=0 \\ \nonumber
		& \quad \eta_{i} \geqslant 0, \quad i=1,2, \cdots, N \nonumber
	\end{align}
	
	The optimal solution of the above constrained optimization problem is denoted as $\eta^*=(\eta_1^*,\eta_2^*,\cdots,\eta_N^*)$. Then, $w^*$ can be calculate by:
	\begin{align}
		w^{*}&=\sum_{i=1}^{N} \eta_{i}^{*} y_{i} x_{i}
	\end{align}
	Next, choose a  positive component  $\eta_j^*$ to calculate $b^*$ as below,
	\begin{align}
		b^{*}&=y_{j}-\sum_{i=1}^{N} \eta_{i}^{*} y_{i}\left(x_{i} \cdot x_{j}\right)
	\end{align}
	
	In SVM, the hyperplane that separates the two classes is formulated as:
	\begin{equation}
		\label{eq:svm-sol}
		w^{*} \cdot x+b^{*}=0
	\end{equation}
	
	From Eq (\ref{eq:svm-sol}), $w^{*}$ is the normal vector of the hyperplane.

\end{document}